\documentclass[MAL,biber]{nowfnt}

\usepackage[utf8]{inputenc}
\usepackage{outlines}
\usepackage{amsmath,amssymb}
\usepackage{xcolor}
\usepackage{graphicx}
\usepackage{textcomp} 
\usepackage{caption}
\usepackage{subcaption}
\usepackage{multicol}
\usepackage{multirow}
\usepackage{xspace}
\usepackage{array}
\usepackage{longtable}
\usepackage{algorithm}
\usepackage{algorithmic}

\newcolumntype{x}[1]{>{\centering\let\newline\\\arraybackslash\hspace{0pt}}p{#1}}
\newcolumntype{C}[1]{ >{\centering\arraybackslash} m{#1} }

\usepackage{amsmath,amsfonts,bm}

\def\eqref#1{equation~\ref{#1}}

\def\1{\bm{1}}

\DeclareMathAlphabet{\mathsfit}{\encodingdefault}{\sfdefault}{m}{sl}
\SetMathAlphabet{\mathsfit}{bold}{\encodingdefault}{\sfdefault}{bx}{n}

\newcommand{\E}{\mathbb{E}}

\newcommand{\D}{\mathcal{D}}
\newcommand{\rlsquared}{RL$^2$\xspace}

\definecolor{luisas_color}{HTML}{3e9137}

\DeclareSourcemap{
\maps[datatype=bibtex]{
\map[overwrite]{
\step[fieldset=isbn, null]
\step[fieldset=issn, null]
\step[fieldsource=doi, final]
\step[fieldset=url, null]
}}}

\usepackage[font=footnotesize,labelfont=bf]{caption}

\title{A Tutorial on Meta-Reinforcement Learning}

\booltrue{authortwocolumn}
\maintitleauthorlist{
    Jacob Beck \\
    University of Oxford \\
    jacob.beck@cs.ox.ac.uk
    \and
    Risto Vuorio \\
    University of Oxford \\
    risto.vuorio@cs.ox.ac.uk
    \and
    Evan Zheran Liu \\
    Stanford University \\
    evanliu@cs.stanford.edu
    \and
    Zheng Xiong \\
    University of Oxford \\
    zheng.xiong@cs.ox.ac.uk
    \and
    Luisa Zintgraf \\
    University of Oxford \\
    zintgraf@deepmind.com
    \and
    Chelsea Finn \\
    Stanford University \\
    cbfinn@cs.stanford.edu
    \and
    Shimon Whiteson \\
    University of Oxford \\
    shimon.whiteson@cs.ox.ac.uk
}

\author[1*$\dagger$]{Beck,Jacob}
\author[1*$\ddagger$]{Vuorio,Risto}
\author[2$\Diamond$]{Liu,Evan Zheran}
\author[1]{Xiong,Zheng}
\author[1\S]{Zintgraf,Luisa}
\author[2]{Finn,Chelsea}
\author[1]{Whiteson,Shimon}

\affil[1]{University of Oxford, UK; jacob.beck@cs.ox.ac.uk, risto.vuorio@cs.ox.ac.uk, zheng.xiong@cs.ox.ac.uk, zintgraf@deepmind.com, shimon.whiteson@cs.ox.ac.uk}
\affil[2]{Stanford University, USA; evanliu@cs.stanford.edu, cbfinn@cs.stanford.edu}

\articledatabox{\textsuperscript{*}Contributed equally.\\ $^\dagger$Now at Oracle.\\ $^\ddagger$Now at Inflection.\\ $^\Diamond$Now at Imbue.\\ \textsuperscript{\S}Now at DeepMind.\\ \nowfntstandardcitation}

\issuesetup
{%
 copyrightowner={J. Beck \textit{et al.}},
 volume        = 18,
 issue         = 2-3,
 pubyear       = 2025,
 isbn          = 978-1-63828-540-3,
 eisbn         = 978-1-63828-541-0,
 doi           = 10.1561/2200000080,
 firstpage     = 224,
 lastpage      = 384
 }

\begin{document}

\makeabstracttitle

\begin{abstract}
While deep reinforcement learning (RL) has fueled multiple high-profile successes in machine learning, it is held back from more widespread adoption by its often poor data efficiency and the limited generality of the policies it produces.
A promising approach for alleviating these limitations is to cast the development of better RL algorithms as a machine learning problem itself in a process called meta-RL.
Meta-RL is most commonly studied in a problem setting where, given a distribution of tasks, the goal is to learn a policy that is capable of adapting to any new task from the task distribution with as little data as possible.
In this survey, we describe the meta-RL problem setting in detail as well as its major variations.
We discuss how, at a high level, meta-RL research can be clustered based on the presence of a task distribution and the learning budget available for each individual task.
Using these clusters, we then survey meta-RL algorithms and applications.
We conclude by presenting the open problems on the path to making meta-RL part of the standard toolbox for a deep RL practitioner.
\end{abstract}

\addtocontents{toc}{\protect\enlargethispage{-2\baselineskip}}
\chapter{Introduction}
\label{section:introduction}
Meta-reinforcement learning (meta-RL) considers a family of machine learning (ML) methods that \emph{learn to reinforcement learn}.
That is, meta-RL methods use sample-inefficient ML to learn sample-efficient RL algorithms, or components thereof.
As such, meta-RL is a special case of meta-learning \citep{vanschoren2018meta,hospedales2020meta,huisman2021survey}, with the property that the learned algorithm is an RL algorithm.
Meta-RL has been investigated as a machine learning problem for a significant period of time \citep{schmidhuber1987evolutionary,schmidhuber1997shifting,thrun1998learning,schmidhuber2007godel}.
Intriguingly, research has also shown an analogue of meta-RL in the brain~\citep{wang2018prefrontal}.

Meta-RL has the potential to overcome some limitations of existing human-designed RL algorithms.
While there has been significant progress in deep RL over the last several years, with success stories such as mastering the game of Go \citep{silver2016mastering}, stratospheric balloon navigation \citep{bellemare2020autonomous}, or robot locomotion in challenging terrain~\citep{miki2022learning}.
RL remains highly sample inefficient, which limits its real-world applications.
Meta-RL can produce (components of) RL algorithms that are much more sample efficient than existing RL methods, or even provide solutions to previously intractable problems.

At the same time, the promise of improved sample efficiency comes with two costs.
First, meta-learning requires significantly more data than standard learning, as it trains an entire learning algorithm (often across multiple tasks).
Second, meta-learning fits a learning algorithm to meta-training data, which may reduce its ability to generalize to other data.
The trade-off that meta-learning offers is thus improved sample efficiency at test time, at the expense of sample efficiency during training and generality at test time.

\paragraph{Example application}
Consider, as a conceptual example, the task of automated cooking with a robot chef.
When such a robot is deployed in somebody's kitchen, it must learn a kitchen-specific policy, since each kitchen has a different layout and appliances.
This challenge is compounded by the fact that not all items needed for cooking are in plain sight; pots might be tucked away in cabinets, spices could be stored on high shelves, and utensils might be hidden in drawers. Therefore, the robot needs not only to understand the general layout but also remember where specific items are once discovered.
Training the robot directly in a new kitchen from scratch is too time consuming and potentially dangerous due to random behavior early in training.
One alternative is to pre-train the robot in a \textit{single} training kitchen and then fine-tune it in the new kitchen.
However, this approach does not take into account the subsequent fine-tuning procedure.
In contrast, meta-RL would train the robot on a \textit{distribution} of training kitchens such that it can adapt to any new kitchen in that distribution.
This may entail learning some parameters to enable better fine-tuning, or learning the entire RL algorithm that will be deployed in the new kitchen.
A robot trained this way can both make better use of the data collected and also collect better data, e.g., by focusing on the unusual or challenging features of the new kitchen.
This meta-learning procedure requires more samples than the simple fine-tuning approach, but it only needs to occur once, and the resulting adaptation procedure can be significantly more sample efficient when deployed in the new test kitchen.

This example illustrates how, in general, meta-RL may be particularly useful when the need for efficient adaptation is frequent, so the cost of meta-training is relatively small.
This includes, but is not limited to, safety-critical RL domains, where efficient data collection is necessary and exploration of novel behaviors is prohibitively costly or dangerous.
In many cases, a large investment of sample-inefficient learning upfront (either with oversight, in a laboratory, or in simulation) is worthwhile to enable subsequent improved adaptation behavior.
This example represents an aspirational application for meta-RL.
In practice, meta-RL is applied to more limited robotics tasks such as robotic manipulation~\citep{akkaya2019solving,zhao2022offline} and locomotion \citep{song2020rapidly}.

\paragraph{Survey scope}
This is a survey of the meta-RL topic in machine learning and leaves out research on meta-RL in other fields such as neuroscience.
Research on closely related machine learning topics is discussed in Section~\ref{subsec:related}.
To capture the breadth and depth of machine learning research on meta-RL, we surveyed the proceedings of several major machine learning conferences, as well as specialized workshops from the time period between 2017 and 2022.
We found that a major portion of the meta-RL literature emerged post-2016, with the lion's share of contributions being concentrated in three conferences: NeurIPS, ICML, and ICLR.
For a full list of conferences and workshops covered, see Appendix~\ref{app:conferences}.
While our survey primarily emphasizes these conferences and the specified timeframe, we also discuss a selection of relevant papers from outside this scope.
From the proceedings of these conferences and workshops, we searched for papers that explicitly mention meta-RL as well those that do not make an explicit reference but that we judged to nevertheless fit the topic.
Finally, we do not claim exhaustive coverage of meta-RL research included in our survey scope but rather a holistic overview of the most salient ideas and general directions.

\paragraph{Survey overview}
The aim of this survey is to provide an entry point to meta-RL, as well as reflect on the current state of the field and open areas of research.
In Section~\ref{section:background}, we define meta-RL and the different problem settings it can be applied to, together with two example algorithms.

\begin{figure}[ht!]
    \centering
    \begin{subfigure}[b]{0.35\textwidth}
         \centering
         \includegraphics[width=\textwidth,alt={The agent is meta-trained on tasks to navigate to a goal located on a unit circle around its starting position.}]{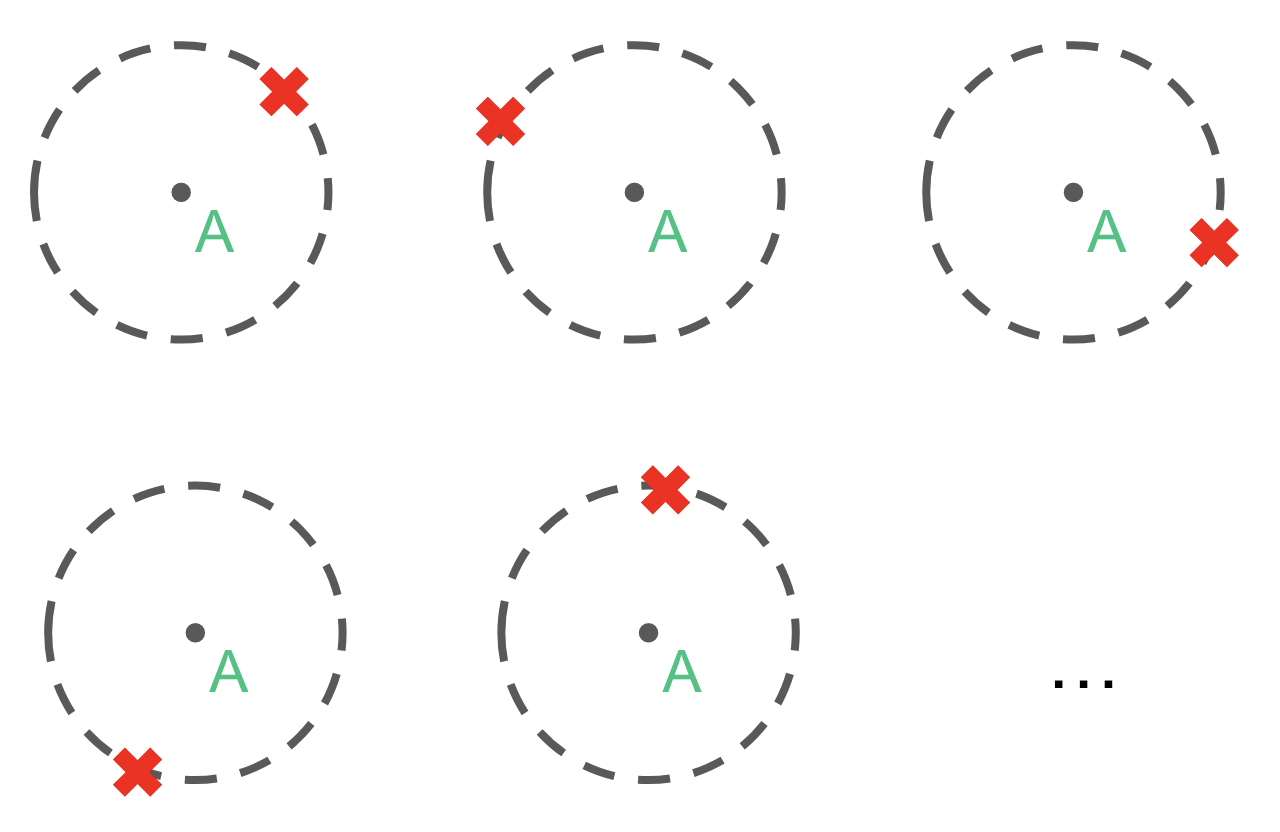}
         \caption{Meta-Training Tasks}
         \label{subfig:intro_example_short:train}
     \end{subfigure}
     \begin{subfigure}[b]{0.6\textwidth}
         \centering
         \includegraphics[width=\textwidth,alt={At meta-test time, the agent quickly navigates to new tasks with initially unknown goal positions.}]{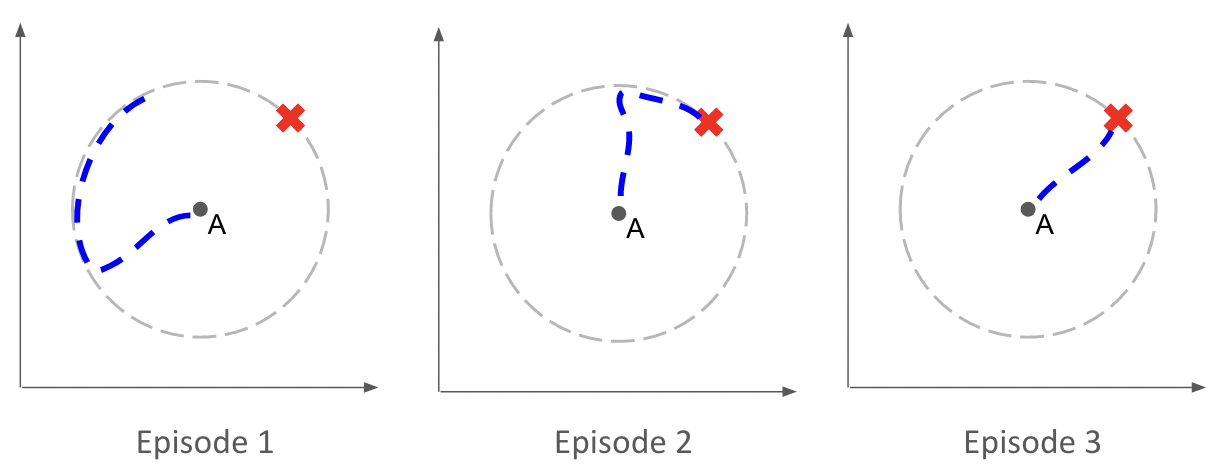}
         \caption{Rollout at Meta-Test Time}
         \label{subfig:intro_example_short:test}
     \end{subfigure}
    \caption{
    Example of the \emph{fast adaptation} meta-RL problem setting discussed in Section~\ref{sec:fast_adaptation}.
    The agent (A) is meta-trained on a distribution of meta-training tasks to \emph{learn} to go to goal position (X) located on a unit circle around its starting position \textbf{(a)}.
    At meta-test time, the agent can adapt quickly (within a handful of episodes) to new tasks with initially unknown goal positions \textbf{(b)}.
    In contrast, a standard RL algorithm may need hundreds of thousands of environment interactions when trained from scratch on one such task.
    }
    \label{fig:intro_example_short}
\end{figure}

In Section~\ref{sec:fast_adaptation}, we consider the most prevalent problem setting in meta-RL: few-shot meta-RL.
Here, the goal is to learn an RL algorithm capable of \emph{fast adaptation}, i.e., learning a task within just a handful of episodes.
These algorithms are often trained on a given task distribution, and meta-learn how to efficiently adapt to any task from that distribution.
A toy example to illustrate this setting is shown in Figure~\ref{fig:intro_example_short}.
Here, an agent is meta trained to learn how to navigate to different (initially unknown) goal positions on a $2$D plane.
At meta-test time, this agent can adapt efficiently to new tasks with unknown goal positions.

In Section~\ref{sec:long_task_horizon_meta_rl}, we consider many-shot settings.
The goal here is to learn general-purpose RL algorithms not specific to a narrow task distribution, similar to those currently used in practice.
There are two flavors of this: training on a distribution of tasks as above, or training on a single task but meta-learning alongside standard RL training.

Next, Section~\ref{sec:application} presents some applications of meta-RL such as robotics.
To conclude the survey, we discuss open problems in Section~\ref{sec:open_problems}.
These include generalization to broader task distributions for few-shot meta-RL, optimization challenges in many-shot meta-RL, and reduction of meta-training costs.

To provide high-level summaries of meta-RL research cited in this survey, we collect representative papers discussed in each section in a summary table presented within the section.

\chapter{Background}
\label{section:background}
Meta-RL can be broadly described as learning a part or all of an RL algorithm. In this section, we define and formalize meta-RL. In order to do so, we start by defining RL.

\section{Reinforcement Learning}
An RL algorithm learns a policy to take actions in a Markov decision process (MDP), also called the agent's \emph{environment}.
An MDP is defined by a tuple $\mathcal{M} = \langle\mathcal{S}, \mathcal{A}, P, P_0, R, \gamma, N\rangle$, where $\mathcal{S}$ is the set of states, $\mathcal{A}$ the set of actions, $P(s_{t+1}|s_t, a_t) : \mathcal{S} \times \mathcal{A} \times \mathcal{S} \rightarrow \mathbb{R}_+$ the transition function, $P_0(s_0): \mathcal{S} \rightarrow \mathbb{R}_+$ is a distribution over initial states, $R(s_t, a_t): \mathcal{S} \times \mathcal{A} \rightarrow \mathbb{R}$ is a reward function, $\gamma \in [0, 1]$ is a discount factor, and $N$ is the horizon.
We give the definitions assuming a finite horizon for simplicity, though many of the algorithms we consider also work in the variable and infinite horizon setting.
A policy is a function $\pi(a | s) : \mathcal{S} \times \mathcal{A} \rightarrow \mathbb{R}_+$ that maps states to action probabilities.
The interaction of the policy with the MDP takes place in \textit{episodes}, which start from initial states sampled from $P_0$ followed by $N$ transitions between states where the actions are sampled from the policy $\pi$ and states from the dynamics $P$.
After the $N$ transitions, a new episode begins starting from a freshly sampled initial state.
The reward for each transition is defined by the reward function $r_t = R(s_t, a_t)$.
This defines a distribution over episodes
\begin{equation}
    P(\tau) = P_0(s_0) \prod_{t=0}^{N-1} \pi(a_t|s_t) P(s_{t+1} |s_t, a_t).
\end{equation}
We refer to the data $\tau = (s_0, a_0, r_0, \dots, s_{N-1}, a_{N-1}, r_{N-1}, s_{N})$ collected during an episode as a \textit{trajectory}.
Note that the trajectory consists of one fewer action and reward than state, since the agent does not take an action in the final state.

The objective of RL is to learn a policy that maximizes the expected discounted return within an episode,
\begin{equation}
    J(\pi)=\mathbb{E}_{\tau \sim P(\tau)}\left[\sum_{t=0}^{N-1} \gamma^t r_t\right], \label{eq:rl-objective}
\end{equation}
where $r_t$ are the rewards along the trajectory $\tau$.
In the process of optimizing this objective, multiple episodes are gathered.
We denote the trajectories collected so far, across multiple episodes, including the current (incomplete) trajectory, as $\mathcal{D}$.
When learning is done and $\mathcal{D}$ is at its largest size, we write this as $\mathcal{D} = (\tau_0, \tau_1, \dots, \tau_H)$, where $H$ is the maximum number of episodes used for learning.

An RL algorithm is a function that maps the data to a policy.
This includes choosing the policies used for data collection during training.
These intermediate policies are not necessarily greedy with respect to the RL objective but may take exploratory actions instead.
In this survey, we mostly consider parameterized policies with parameters $\phi \in \Phi$.
Therefore, we define an RL algorithm as the function $\phi = f(\mathcal{D})$.

\section{Meta-RL Definition}

RL algorithms are traditionally designed, engineered, and tested by humans.
The idea of meta-RL is instead to learn (parts of) an algorithm $f$ using machine learning.
Where RL learns a policy, meta-RL learns the RL algorithm $f$ that outputs the policy.
This does not remove all of the human effort from the process, but shifts it from directly designing and implementing the RL algorithms into developing the training environments and parameterizations required for learning parts of them in a data-driven way.

\begin{figure}[t!]
    \centering
    \includegraphics[width=0.5\textwidth,alt={Diagram of a meta-RL algorithm showing the relationship between the inner-loop and outer-loop.}]{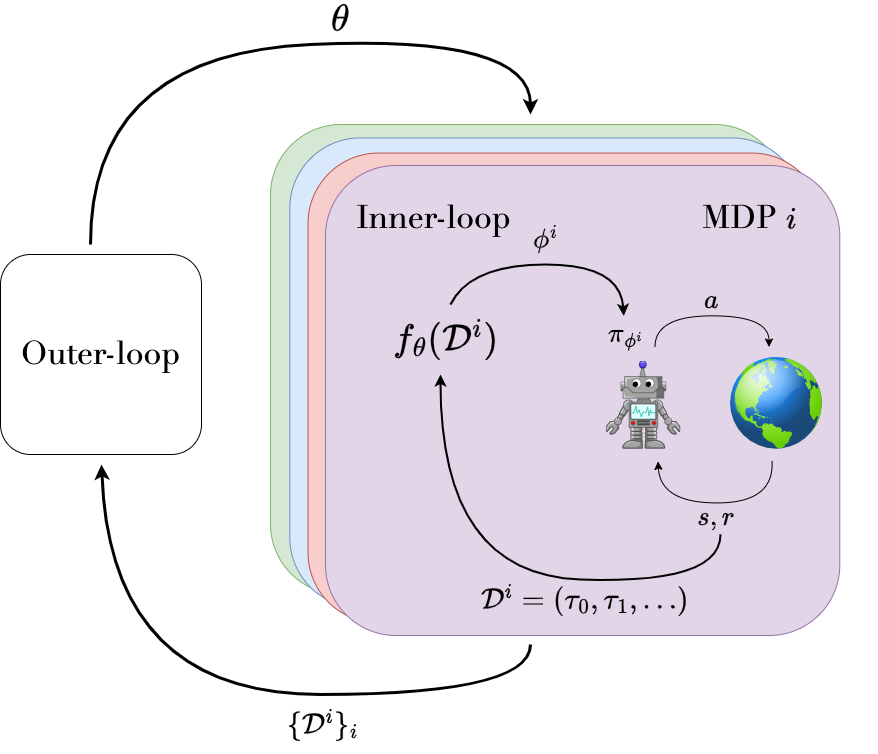}
    \caption{The relationship between the inner-loop and outer-loop in a meta-RL algorithm. The policy for MDP $i$, parameterized by $\phi^i$, produces the meta-trajectory data $\mathcal{D}^i$.
    The inner-loop $f_\theta$ computes adapted policy parameters for each MDP based on the meta-trajectory during the policy's interaction with the MDP.
    To compute the adapted parameters, the inner-loop can use all data collected in the MDP so far.
    The outer-loop computes updated meta-parameters using all of the meta-trajectories collected in all of the MDPs.
    }
    \label{fig:inner-outer-loop}
\end{figure}

\begin{figure}[ht!]
\centering
\includegraphics[width=\textwidth,alt={Meta-training consists of trials each made up of multiple episodes.}]{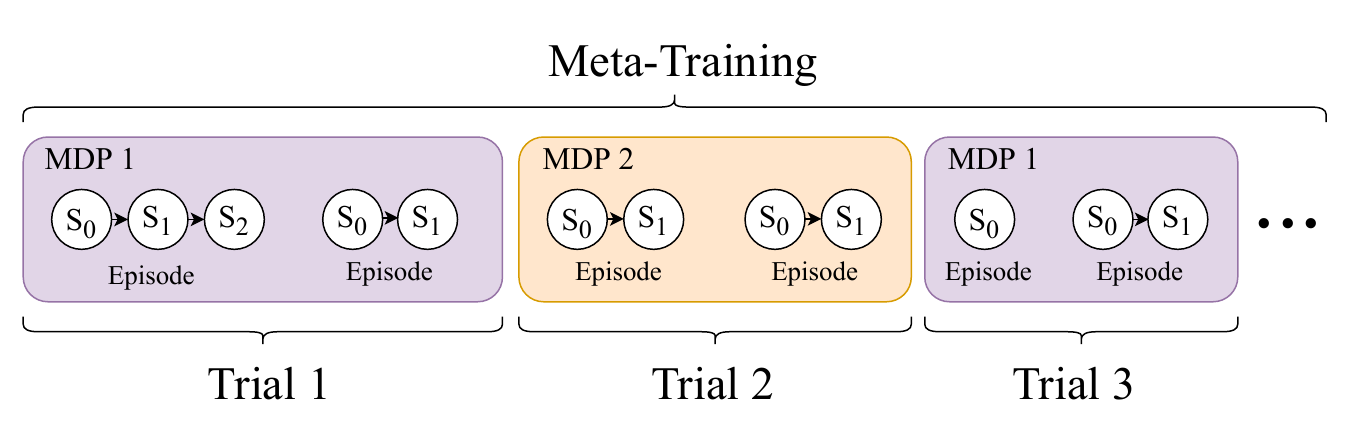}
\caption{Meta-training consists of trials (or lifetimes), each broken up into multiple episodes from a single task (MDP).
In this example, each trial consists of two episodes ($H=2$).
}
\label{fig:meta_rl_training}
\end{figure}

Due to this bi-level structure, the algorithm for learning $f$ is often referred to as the \emph{outer-loop}, while the learned $f$ is called the \emph{inner-loop}.
The relationship between the inner-loop and the outer-loop is illustrated in Figure~\ref{fig:inner-outer-loop}.
Since the inner-loop and outer-loop both perform learning, we refer to the inner-loop as performing \textit{adaptation} and the outer-loop as performing \textit{meta-training}, for the sake of clarity.
The learned inner-loop, that is, the function $f$, is assessed during \textit{meta-testing}.
We want to meta-learn an RL algorithm, or an inner-loop, that can adapt quickly to a new MDP.
This meta-training requires access to a set of training MDPs.
These MDPs, also known as \textit{tasks}, come from a distribution denoted $p(\mathcal{M})$.
In principle, the task distribution can be supported by any set of tasks.
However, in practice, it is common for $\mathcal{S}$ and $\mathcal{A}$ to be shared between all of the tasks and the tasks to only differ in the reward $R(s,a)$ function, the dynamics $P(s'|s, a)$, and initial state distributions $P_0(s_0)$.
Meta-training proceeds by sampling a task from the task distribution, running the inner-loop on it, and optimizing the algorithm to improve the policies it produces.
The interaction of the inner-loop with the task, during which the adaptation happens, is called a \textit{lifetime} or a \textit{trial} and is illustrated in Figure~\ref{fig:meta_rl_training}.
A trial can consist of multiple episodes.
After an episode ends, a new one begins with the initial state of the new episode sampled from the initial state distribution $P_0(s_0)$.
The concatenation of all the data during a single trial, $\mathcal{D}$, which may contain multiple episodes, is called a
\textit{meta-trajectory}.

\paragraph{Parameterization} We generally parameterize
the inner-loop $f_\theta$ with \emph{meta-parameters} $\theta$, and learn these parameters to maximize a meta-RL objective.
Hence, $f_\theta$ outputs the parameters of $\pi_{\phi}$ directly: $\phi=f_\theta(\mathcal{D})$.
We refer to the policy $\pi_\phi$ as the \textit{base policy}.
Frequently, the inner-loop will only adapt a subset of $\phi$, which we refer to as the \textit{adapted policy parameters}, or the \textit{task parameters}, since they are adapted to each task.
Some meta-RL algorithms use $f_\theta$ to map $\mathcal{D}$ to $\phi$ at every MDP step, while others do so less frequently.
In either case case, we use $\phi_j$ to denote the $j$th set of parameters.
Additionally, if the parameters are bound to a specific task, and there are a finite number of tasks, we use $\phi_j^i$ to indicate the $j$th set of parameters for the $i$th task.
Likewise, we use $\mathcal{D}_j^i$ to indicate the data for the $j$th update for the $i$th task.

\paragraph{Meta-RL objective}
In standard RL, a Markov policy suffices to maximize the RL objective given by Equation~\ref{eq:rl-objective}.
In contrast, in meta-RL, the inner-loop is an entire RL algorithm that outputs parameters of policies as a function of the meta-trajectory.
The performance of a meta-RL algorithm is measured in terms of the returns achieved by the policies $\pi_\phi$ produced by the inner-loop during a trial on tasks $\mathcal{M}$ drawn from the task distribution.
Depending on the application, slightly different objectives are considered.
For some applications, we can afford a free exploration or adaptation period, during which the performance of the policies produced by the inner-loop is not important as long as the final policy found by the inner-loop solves the task.
The episodes during this phase can be used by the inner-loop for freely exploring the task.
For other applications, a free exploration period is not possible and correspondingly the agent must maximize the expected return from the first timestep it interacts with the environment.
Maximizing these different objectives leads to different learned exploration strategies: a free exploration period enables more risk-taking at the cost of wasted training resources when the risks are realized.
Formally, the meta-RL objective for both settings is:
\begin{align}
    \mathcal{J}(\theta) = \E_{\mathcal{M} \sim p(\mathcal{M})} \bigg[ \E_{\mathcal{D}} \bigg[ G(\mathcal{D}) \bigg| \pi_{f_\theta(\mathcal{D})}, \mathcal{M} \bigg]\bigg], \label{eq:meta-rl-objective}
\end{align}
where $G(\D) = \sum_{i=K}^{H} \sum_{j=0}^{N} \gamma^{i * N + j} r_{i, j}$ is the discounted return in the MDP $\mathcal{M}$ starting at the $K$th episode of the meta-trajectory $\mathcal{D}$, $r_{i, j}$ is the $j$th reward in episode $i$, and $H$ is the length of the trial, or the \textit{task-horizon}.
This objective discounts rewards across episodes in $\mathcal{D}$.
The exploration period is captured by $K$, also called the \textit{shot}, which is the index of the first episode during the trial in which return counts towards the objective.
$K=0$ corresponds to no free exploration episodes.

The meta-RL objective defined by Equation \ref{eq:meta-rl-objective} is evaluated in expectation over samples from the task distribution, $p(\mathcal{M})$.
During meta-training, the task distribution  $p(\mathcal{M})$ is usually accesses only via sampling.
This results in a generalization problem in meta-testing, as testing tasks may not match the training tasks.
This problem is made worse when the meta-testing task distribution does not share support with the training task distribution.
Such out-of-distribution cases are often studied in meta-RL.

\section{POMDP Formalization}
\label{subsec:pomdp-formalization}
The problem setting defining meta-RL, posed by Equation \ref{eq:meta-rl-objective}, can additionally be viewed as a special case of a partially observable Markov decision process (POMDP) \citep{duan2016rl, humplik2019meta}.
We specify the POMDP as a tuple of $\mathcal{P} = \langle\mathcal{S}, \mathcal{A}, \Omega, P, P_0, R, O, \gamma, H \rangle$.
Here, $\Omega$ is the set of observations, $O$ is the observation function, $H$ is the task-horizon, and the rest are defined as in an MDP.
To resolve ambiguity, when a function or variable depends on a particular MDP, we use a superscript to indicate the dependence, e.g., $R^\mathcal{M}$, for a reward function that depends on the MDP, $\mathcal{M}$.
The POMDP states, actions, and transitions function similarly to an MDP.
However, unlike in an MDP, the agent does not observe the current state, $s_t$, which is hidden from the agent.
Instead, the agent observes an observation, $o_t \in \Omega$, with its probability distribution defined by $O(o_t|s_t,a_t): \Omega \times \mathcal{S} \times \mathcal{A} \rightarrow \mathbb{R}_+$.
The POMDP policy is then a function that maps a history of observations to a distribution over actions: $\pi(a | \tau_{:t}): \mathcal{A} \times \Omega^t \rightarrow \mathbb{R}_+$.

Given a distribution of MDPs, $p(\mathcal{M})$, the POMDP corresponding to the meta-RL problem has a specific structure.
In particular, it maintains the identity of an MDP sampled from this distribution, i.e., the reward function and the dynamics function, in its hidden state.
Therefore, the state of the POMDP is the identity and state of the sampled MDP: $s_t = (\mathcal{M},s^\mathcal{M}_t)$.
The initial state distribution of the POMDP is responsible for sampling an MDP, $\mathcal{M}$, and an initial MDP state, $s^\mathcal{M}_0$, given the MDP:
$P_0(s_0) = p(\mathcal{M})P^{\mathcal{M}}_0(s^\mathcal{M}_0)$.
The reward function of the POMDP is the same as the reward function of the underlying MDP: $R(s_t, a_t) = R^{\mathcal{M}}(s^\mathcal{M}_t, a_t)$.
The observation function deterministically returns the state of the underlying MDP, along with the previous action, and the reward: $o_t = (s^\mathcal{M}_t, a_{t-1}, r_{t-1})$.
In meta-RL, the unknown MDP identity includes an unknown reward function.
In order for the inner-loop to learn an optimal policy for the MDP, it requires information about the unknown reward function.
Therefore, the observation includes the sampled reward, $r_{t-1}$, which is not always the case for observation functions for POMDPs considered outside of meta-RL.
The dynamics function of the POMDP is that of the sampled MDP stored in the hidden state.
Additionally, after $N$ timesteps of the MDP, the dynamics function of the POMDP must be constructed so that it transitions to initial states sampled from $P^\mathcal{M}_0$ of the underlying MDP.
Formally,
\[
s_{t+1} =
\begin{cases}
      (\mathcal{M},s_0^\mathcal{M} \sim P_0^{\mathcal{M}}) & \text{if } t \equiv 0 \mod N,\\
      (\mathcal{M},s^\mathcal{M}_{t+1}) & \text{otherwise}.
   \end{cases}
\]
A single episode in this particular type of POMDP thus consists of an entire trial, itself comprised of $H$ episodes in the sampled MDP.
When the task-horizon, $H$, is reached, the POMDP episode terminates.
In the beginning of the next POMDP episode, a new POMDP state, including a new MDP, is sampled from the initial state distribution $P_0$.
Note that the transition function depends on the timestep, $t$, which can be stored in $s$, but is omitted from the state for brevity.
Also, note that the history of observations in this POMDP, $\tau$, is the concatenation of observations, $(s^\mathcal{M}_t, a_{t-1}, r_{t-1})$, across the $H$ MDP episodes.
Together, the sequence of these observations
is equivalent to the dataset $\mathcal{D}$ from Section \ref{section:background}.
(This formalism can also be cast as a Contextual MDP, or CMDP, \cite{hallak2015contextual}, where the context is not shown to the agent.)
This perspective leads to the following insights into the meta-RL problem.

First, from the theory of POMDPs, we know that the optimal policy is either history-dependent or dependent on a sufficient statistic, or the information state for the POMDP~\citep{subramanian2022approximate}.
For the POMDP representing the meta-RL problem, it is therefore possible to use either representation.
In the case that the inner-loop, $f_\theta$, approximates a general function of history, then the inner-loop and the policy it produces, $\pi^\mathcal{M}_{\phi}$, can be viewed together as a single object, forming a history-dependent policy, $\pi_\theta(a|\tau)$.
While such history-dependent policies do not take into account the specific structure of the meta-RL problem beyond that of a more general POMDP, they are sufficient for solving the meta-RL problem.
Alternatively, the optimal policy may condition on a sufficient statistic that summarizes the history.
In a POMDP, one such statistic is the posterior distribution over POMDP states, $b(s): \mathcal{S} \rightarrow \mathbb{R}_+$~\citep{subramanian2022approximate}, often referred to as the belief state.
In the case that the inner-loop approximates such a belief state, the combined object can be seen as a policy dependent on approximations of a sufficient statistic, $\pi_\theta(a|\hat{b})$, where the inner-loop computes $\hat{b}=f_\theta(\tau)$.
The form of these sufficient statistics can be more specific than in the general POMDP, since the hidden state, $s$, has a specific form in meta-RL.
In particular, one sufficient statistic for meta-RL, given the current observable MDP state, is the posterior distribution over tasks, $b = p(\mathcal{M} \mid \tau) = p(R^\mathcal{M}, P^\mathcal{M} \mid \tau)$, which we discuss further in Section \ref{sec:optimal_exploration}.
This dichotomy in POMDP methods, between history-dependent and belief-dependent, leads to two different meta-RL methods: black box methods, discussed in Section \ref{sec:black_box}, and task inference methods, discussed in Section \ref{sec:task_inf}, respectively.

\looseness=+1
Second, this perspective enables us to draw connections to Bayesian RL \citep{duff2002optimal, ghadirzadeh2021bayesian}.
In particular, Bayesian RL solves this POMDP by explicitly maintaining a posterior over tasks and updating it using Bayesian inference.
The Bayesian framework provides a convenient method for incorporating prior knowledge and explicitly maintaining uncertainty.
Using this framework a new MDP can be constructed, where the state includes the posterior over POMDP hidden states, or equivalently, the MDP state and a posterior over tasks.
In this case, the resulting MDP is called a Bayes-adaptive Markov decision process (BAMDP).
This construction allows learning a Markovian policy to solve the meta-RL problem and therefore explores optimally.
BAMDPs and Bayes-optimal policies are discussed in Section~\ref{sec:optimal_exploration}.
However, since Bayesian RL methods must explicitly model and update a distribution over MDPs, they are tractable only in simple domains without strong approximations.
For example, even scalable Bayesian RL methods may be limited to discrete-space MDPs \citep{guez2013scalable}.
Instead of engineering approximations for the Bayesian posterior, meta-RL methods may learn to model these components as needed.
For example, most meta-RL methods only require sample access to the prior instead of the prior being explicitly known.
The meta-RL agent may learn to implicitly model the prior over tasks from samples.

Finally, while Meta-RL generally considers a distribution over MDPs, it is also possible to consider a distribution over POMDPs.
In this case, each task is itself partially observable.
This forms yet another type of POMDP, called a meta-POMDP~\citep{akuzawa2021estimating}.
The meta-POMDP can be written as a POMDP where only a part of the hidden state remains constant throughout a trial, and it is possible to adapt existing methods to accommodate this structure~\citep{akuzawa2021estimating}.

\section{Example Algorithms}
\label{sec:example_algos}

We now describe two canonical meta-RL algorithms that optimize the objective given by Equation~\ref{eq:meta-rl-objective}: Model-Agnostic Meta-Learning (MAML), which uses meta-gradients~\citep{finn2017model}, and Fast RL via Slow RL (RL$^2$), which uses a history dependent policy~\citep{duan2016rl, wang2016learning}.
Many meta-RL algorithms are based on concepts and techniques similar to those used in MAML and RL$^2$, making them excellent entry points to meta-RL.

\paragraph{MAML}
Many designs of the inner-loop algorithm $f_\theta$ build on existing RL algorithms and use meta-learning to improve them. MAML
\citep{finn2017model} is an influential design following this pattern.
Its inner-loop is a policy gradient algorithm whose initial parameters are the meta-parameters $\phi_0 = \theta$.
The key insight is that such an inner-loop is a differentiable function of the initial parameters, and therefore the initialization can be optimized with gradient descent to be a good starting point for learning on tasks from the task distribution.
When adapting to a new task, the inner-loop MAML collects data using the initial policy and computes an updated set of parameters by applying a policy gradient step for a task $\mathcal{M}^i \sim p(\mathcal{M})$:
\begin{equation*}
\phi_{1}^i = f_{\theta}(\D_0^i) = \phi_0 + \alpha \nabla_{\phi_0} \hat{J}(\D_0^i, \pi_{\phi_0}),
\end{equation*}
where $\hat{J}(\D_0^i, \pi_{\phi_0})$ is an estimate of the returns of $\pi_{\phi_0}$ in task $\mathcal{M}^i$ computed on data $\D_0^i$ collected using $\pi_{\phi_0}$.
Note that $\phi_0 = \theta$ is a learnable meta-parameter, which is not specific to any task.
In contrast, $\phi_{1}^i$ are the adapted policy parameters that are specific to the task $\mathcal{M}^i$.
Since $\phi_0$ affects the initial policy, it not only determines initial rewards, but also determines the data collected to produce~$\phi_{1}^i$.

\begin{figure}[ht!]
    \centering
    \includegraphics[height=5cm,alt={Diagram of MAML, left shows the problem setting, right shows the concept.}]{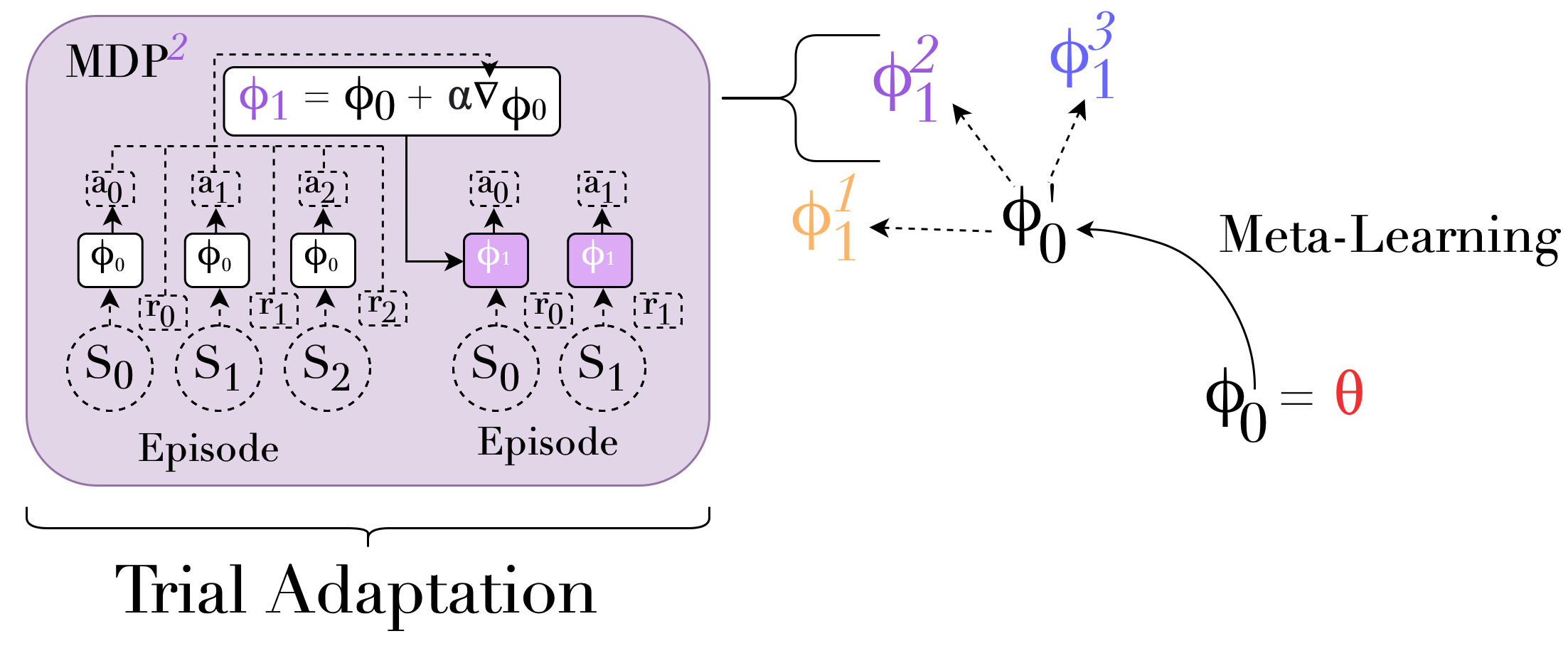}
    \caption{MAML in the problem setting (left) and conceptually (right). The meta-parameters $\theta$ are the initial parameters of the inner-loop policies $\phi_0$. The inner-loop computes new parameters $\phi_1^i$ adapted to task $i$ using one step of a policy gradient algorithm. The outer-loop updates the meta-parameters, from $\phi_0$ to $\phi_0'$, to optimize the performance of the policies after adaptation.
    \label{fig:maml}
    }
\end{figure}

To learn $\theta$ in the outer-loop, it must be updated based on its effect on the initial policy and its effect on $\phi_{1}^i$.
To differentiate through the adaptation process, MAML collects a second batch of data $\mathcal{D}^i_1$ using the policy $\pi_{\phi_1^i}$ and computes the gradient of the returns of the updated policy with respect to the initial parameters across the tasks given by
\begin{equation*}
    \frac{1}{M} \sum_{i=0}^M \nabla_{\phi_0} \hat{J}(\D_1^i, \pi_{\phi_1^i}),
\end{equation*}
where $M$ is the number of sampled tasks, $\pi_{\phi_1^i}$ is the policy for task $\mathcal{M}^i$ updated once by the inner-loop, and $\nabla_{\phi_0} \hat{J}(\D_1^i, \pi_{\phi_1}^i)$ is the gradient of the returns of the updated policy computed w.r.t.\ the initial policy parameters.
Such a gradient through an RL inner-loop is often referred to as a \textit{meta-gradient}.
Descending the meta-gradient corresponds to optimizing the outer-loop objective given by Equation~\ref{eq:meta-rl-objective}.
The version of MAML described here considers only a single update in the inner-loop but for harder problems, the inner-loop can compute multiple updates.
The additional updates do not change the meta-gradient computation.
The algorithm is illustrated in Algorithm~\ref{alg:maml} and Figure~\ref{fig:maml}.

\begin{algorithm}[ht!]
\caption{MAML for Reinforcement Learning}
\label{alg:maml}
\begin{algorithmic}[1]
\STATE Initialize meta-parameters, $\theta$, which here are the initial policy parameters, $\phi_0$.
\WHILE{not done}
    \STATE Sample M tasks, $\mathcal{M} \sim p(\mathcal{M})$
    \FOR{each task index, $i$}
        \STATE Collect data $\D_0^i$ using the initial policy $\pi_{\phi_0}$.
        \STATE Adapt policy parameters using a policy gradient step: $\phi_1^i \gets \phi_0 + \alpha \nabla_{\phi_0} \hat{J}(\D_0^i, \pi_{\phi_0})$
        \STATE Collect data $\D_1^i$ using the updated policy $\pi_{\phi_1^i}$.
    \ENDFOR
    \STATE Update $\phi_0$ using the meta-gradient: $\phi_0 \gets \phi_0 + \beta \nabla_{\phi_0} \frac{1}{M} \sum_{i} \hat{J}(\D_1^i, \pi_{\phi_1^i}).$
\ENDWHILE
\end{algorithmic}
\end{algorithm}

Typically in MAML, the shot $K$ and task-horizon $H$ are chosen such that the outer-loop objective only considers the returns of the last policy produced by the inner-loop.
In the simplest case, this would mean setting $K=1$ and $H=2$, i.e., using one episode for computing the inner-loop update and another for computing the outer-loop objective.
To accommodate multiple updates in the inner-loop and sampling multiple episodes with each policy for variance reduction, higher values of $K$ and $H$ are often used.

\paragraph{RL$^2$}
Another popular approach to meta-RL is to represent the inner-loop as a policy that is dependent on \textit{the entire history} of its interaction with the environment and train it on tasks from the task distribution end-to-end with RL.
As discussed in Section~\ref{subsec:pomdp-formalization}, when $f_\theta$ is represented as a sequence model, $f_\theta$ and $\pi_{f_\theta(\mathcal{D})}(a|s)$ can be viewed together as a single history-dependent policy, $\pi_\theta(a|s,\mathcal{D})$.
This approach is the same as applying a history-dependent policy, as used in the more general POMDP setting, to meta-RL as a special case.
This history includes all of the states, actions, and rewards the policy encounters during a trial.
This results in learning an adaptive policy that changes its behavior as it gathers more information about the environment.
A similar idea has been explored for supervised learning, for example by \citet{hochreiter2001learning}.
For meta-RL specifically \citet{duan2016rl} and \citet{wang2016learning} propose a method called RL$^2$, where the history dependent policy is represented as a recurrent neural network (RNN).
Using $\phi$ to represent the RNN hidden state, we can write the recurrent policy as $\pi_\theta(a|s,\phi)$.
While in this example we focus on RNNs, any history dependent policy could be used.

\begin{figure}[ht!]
\centering
\includegraphics[width=.85\textwidth,alt={Diagram of RL2, left shows the problem setting, right shows the concept.}]{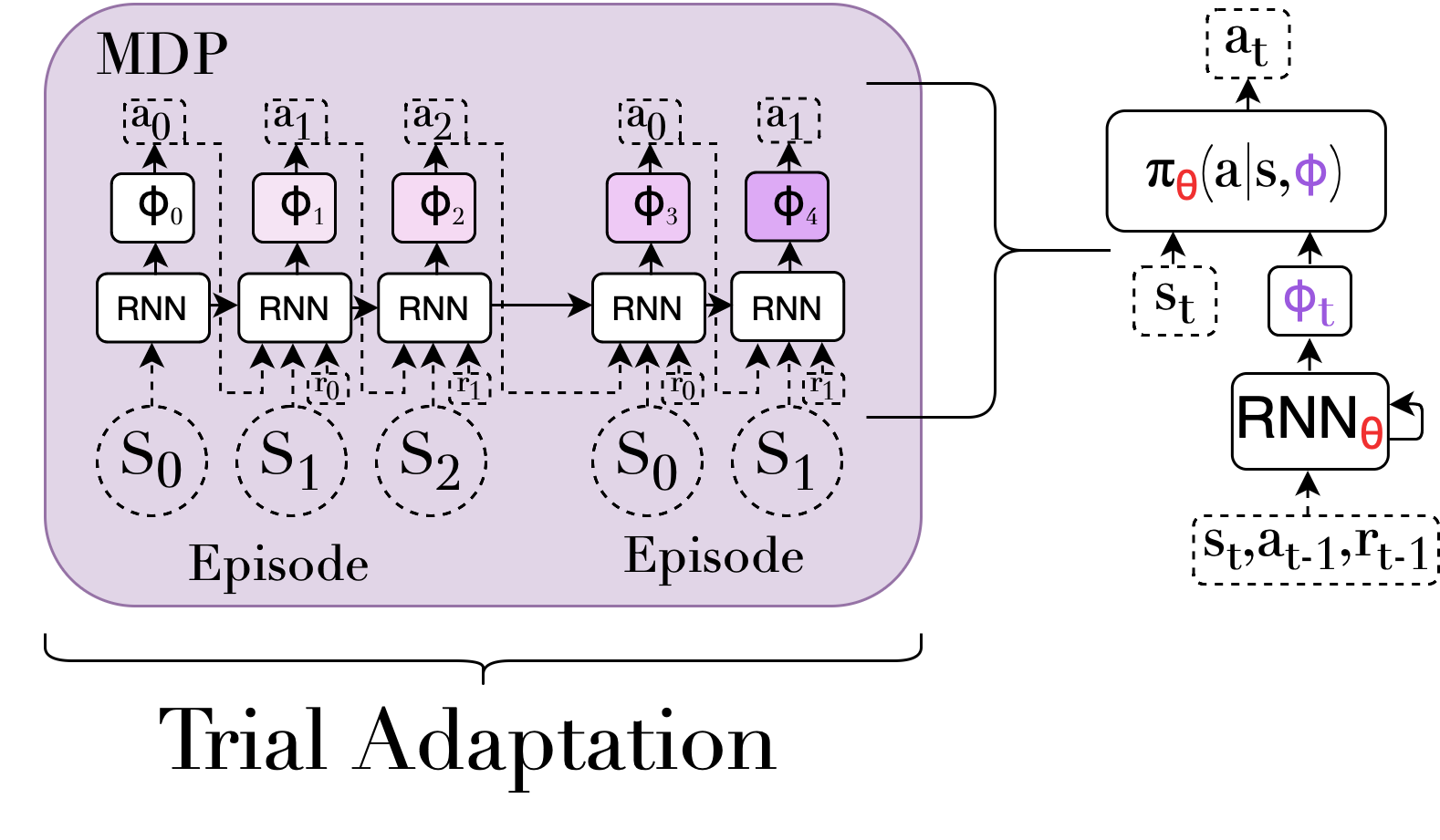}
\caption{RL$^2$ in the problem setting (left) and conceptually (right). The inner-loop algorithm is implemented by an RNN parameterized by the meta-parameters $\theta$. The RNN takes as input the states, actions, and rewards from the environment. The RNN hidden state $\phi_t$ defines the task parameters at each timestep, which are passed as input to the MLP policy. The hidden state is not reset during a trial and instead carries over across episode boundaries. The outer-loop is a standard RL algorithm.
}
\label{fig:rl2}
\end{figure}

\enlargethispage{-2\baselineskip}
To optimize the meta-RL objective from Equation \ref{eq:meta-rl-objective}, RL$^2$ treats the trial as a single continuous sequence, during which the RNN hidden state is not reset, even if the trial spans multiple episodes in the underlying MDP.
The meta-parameters $\theta$ are the parameters of the RNN and other neural networks used in processing the inputs and outputs of $f_\theta$.
The task parameters $\phi$ are the ephemeral hidden states of the RNN, which may change after every timestep.
The operation of the algorithm is illustrated in Algorithm~\ref{alg:rl2} and Figure~\ref{fig:rl2}.

\begin{algorithm}[ht!]
\caption{RL$^2$ for Meta-Reinforcement Learning}
\label{alg:rl2}
\begin{algorithmic}[1]
\STATE Initialize meta-parameters $\theta$ (RNN and other neural network parameters)
\WHILE{not done}
    \STATE Sample M tasks, $\mathcal{M} \sim p(\mathcal{M})$
    \FOR{each task index, $i$}
        \STATE Initialize RNN hidden state $\phi_0^i$
        \STATE Run a continuous trial consisting of multiple episodes
        \FOR{each timestep $t$ in the trial}
            \STATE Observe current state $s_t$, previous action $a_{t-1}$, and previous reward $r_{t-1}$
            \STATE Update RNN hidden state with input $[s_t, a_{t-1}, r_{t-1}]$:
            \[
            \phi_t^i \leftarrow f_\theta(s_t, a_{t-1}, r_{t-1}, \phi_{t-1}^i)
            \]
            \STATE Sample action $a_t \sim \pi_\theta (\cdot | s_t,  \phi_t^i)$
            \STATE Execute action $a_t$ and receive reward $r_t$
        \ENDFOR
    \ENDFOR
    \STATE Update meta-parameters by optimizing Equation \ref{eq:meta-rl-objective}: $\theta \leftarrow \theta + \beta \nabla_{\theta} \frac{1}{M} \sum_{i} G(\D^i).$
\ENDWHILE
\end{algorithmic}
\end{algorithm}

These two distinct approaches to meta-RL, MAML and RL$^2$, each bring their unique advantages and disadvantages.
On one hand, the generality of MAML's policy gradient algorithm in its inner-loop allows it, under specific conditions, to learn a policy starting from its initial state for \textit{any} given task, including those outside the task distribution.
On the other hand, RL$^2$ directly approximates the optimal policy of the meta-RL objective, given by Equation~\ref{eq:meta-rl-objective}.
This policy, known as Bayes-optimal, is the best policy for the task distribution and is further discussed in Section~\ref{sec:optimal_exploration}.
Bayes-optimal policies always choose actions that maximize the expected return under uncertainty about the MDP identity, whereas the policy-gradient dependent MAML can only update when full episodes have been collected in the inner-loop.
However, a drawback of RL$^2$ is that it faces a challenging generalization problem when tested on tasks outside the task distribution.
The end-to-end RL training on a narrow task distribution may not result in a policy that generalizes well outside the task distribution.

\section{Problem Categories}

\begin{figure}[p!]
    \centering
    \includegraphics[width=\linewidth,alt={Illustration of different meta-RL settings, summarizing their goals, task distributions, and related algorithms.}]{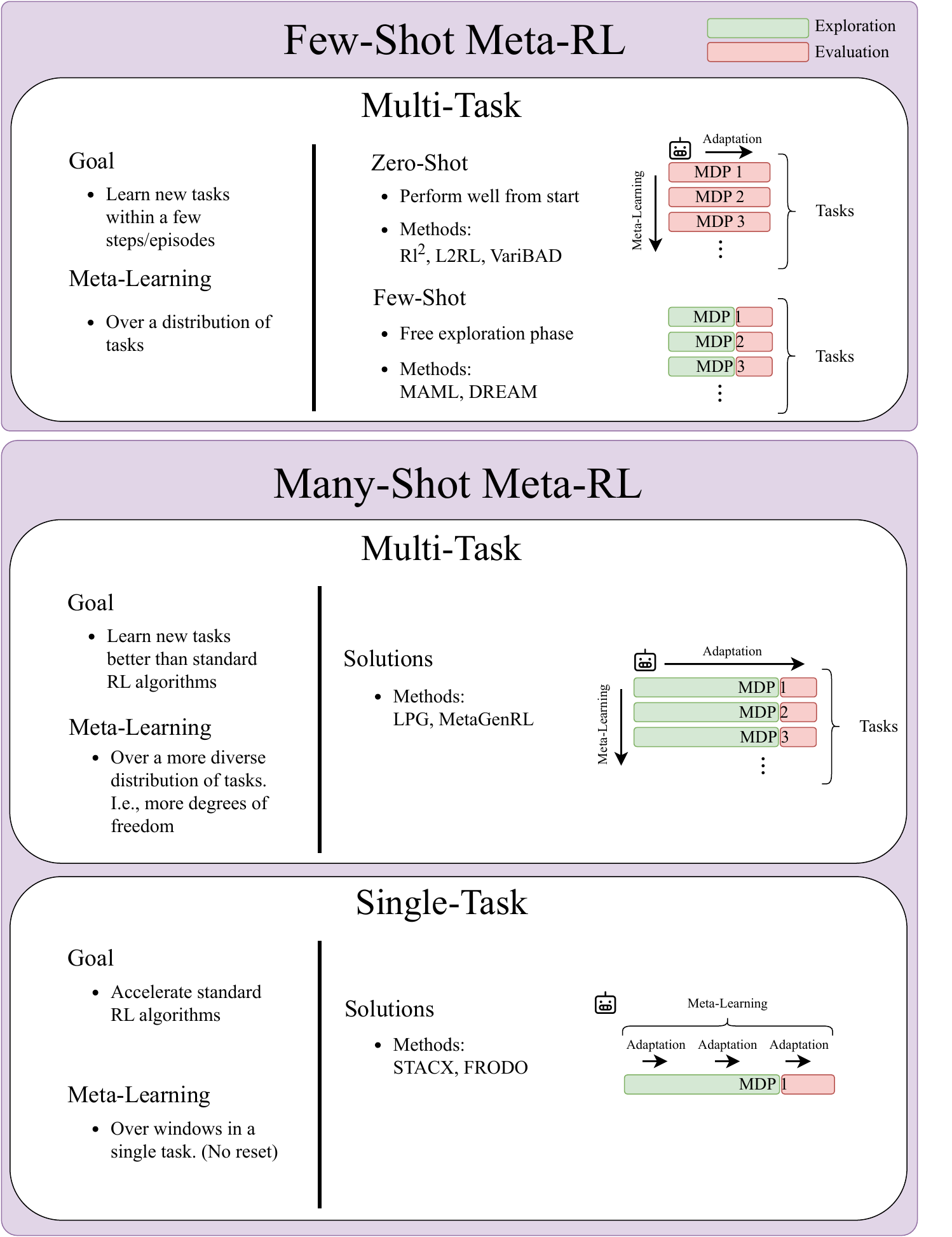}
    \caption{
        Illustration of the different meta-RL settings considered in this survey.
        Each setting is summarized in terms of its motivating goal, task distribution, and the defining features of its algorithms. The small embedded figures on the right illustrate the adaptation over task distributions in each setting. The tasks are shown as bars depicting trials with the shot and task-horizon depicted by the lengths of the green and red bars.
    }
    \label{fig:meta_rl_settings}
\end{figure}

While the given problem setting applies to all of meta-RL, distinct clusters in the literature have emerged based on two dimensions: whether the task-horizon $H$ is short (a few episodes) or long (hundreds of episodes or more), and whether the task distribution $p(\mathcal{M})$ contains multiple tasks or just one.
This creates four clusters of problems, of which three yield practical algorithms, as shown in Table~\ref{tab:categories_of_meta_rl}.
We illustrate how these categories differ in Figure~\ref{fig:meta_rl_settings}.

\begin{table}[ht!]
    \footnotesize
    \centering
    \caption{
    Example methods for the three categories of meta-RL problems. The categorization is based on whether they consider multi-task task distributions or a single task and few or many shots.}
    \label{tab:categories_of_meta_rl}
    \begin{tabular}{C{0.15\textwidth}C{0.35\textwidth}C{0.35\textwidth}}
          & Multi-task & Single-task  \\
          \hline \hline
         Few-Shot
            &
            RL$^2$~\citep{duan2016rl,wang2016learning}, MAML~\citep{finn2017model}
            &
            -
            \\ \hline
         Many-Shot
            &
            LPG~\citep{oh2020discovering}, MetaGenRL~\citep{kirsch2019improving}
            &
            STAC~\citep{zahavy2020self}, FRODO~\citep{xu2020meta} \\ \hline
    \end{tabular}
\end{table}

The first cluster in Table~\ref{tab:categories_of_meta_rl} is the \textit{few-shot multi-task setting}.
In this setting, an agent must quickly -- within just a few episodes -- adapt to a new MDP sampled from the task distribution it was trained on.
This requirement captures the central idea in meta-RL that we want to use the task distribution to train agents that are capable of learning new tasks from that distribution using as few environment interactions as possible.
The \emph{shots} in the name of the setting echo the shots in \emph{few-shot classification}~\citep{vinyals2016matching,snell2017prototypical,finn2017model}, where a model is trained to recognize new classes given only a few samples from each.
In meta-RL, the number of exploration episodes $K$ is more or less analogous to the shots in classification.
This setting is discussed in detail in Section~\ref{sec:fast_adaptation}.

While few-shot adaptation directly tackles the motivating question for meta-RL, sometimes the agent faces an adaptation problem so difficult that it is unrealistic to hope for success within a small number of episodes.
This might happen, for example, when adapting to tasks that do not have support in the task distribution.
In such cases, the inner-loop may require thousands of episodes or more to produce a good policy for a new task, but we would still like to use meta-learning to make the inner-loop as data-efficient as possible.
We discuss this \textit{many-shot multi-task} setting in Section~\ref{sec:long_task_horizon_meta_rl}.

Meta-RL is mostly concerned with the multi-task setting, where the inner-loop can exploit similarities between the tasks to learn a more data-efficient adaptation procedure.
In contrast, in standard RL, agents are often trained to tackle a single complex task over many optimization steps.
Since this is often highly data-inefficient, researchers have investigated  whether meta-learning methods can improve efficiency even without access to a distribution of related tasks.
In such cases, the efficiency gains must come from transfer within the single lifetime of the agent or from adaptation to the local training conditions during training.
Methods for this \textit{many-shot single-task} setting tend to resemble those in the many-shot multi-task setting and are therefore also discussed in Section~\ref{sec:long_task_horizon_meta_rl}.

The fourth cluster is the \textit{few-shot single-task} setting, where a hypothetical meta-learner would accelerate learning on a single task within a few episodes without transfer from other related tasks.
The short lifetime of the agent is unlikely to leave enough time for the inner-loop to learn a data-efficient adaptation procedure and produce a policy using it.
Therefore, there is to our knowledge no research targeting this setting.

\section{Related Work}
\label{subsec:related}
\paragraph{Related Fields}
The objective of this survey is to introduce, summarize, and highlight research in the field of meta-reinforcement learning.
Several existing surveys \citep{thrun1998learning, hospedales2020meta, wang2020generalizing, parker2022automated, kirk2023survey} include some meta-RL methods, but differ in a few key ways.
\citet{thrun1998learning, hospedales2020meta} both include meta-RL methods, but focus on meta-learning more broadly, including meta-supervised learning.
\citet{wang2020generalizing} survey methods for few-shot learning, but meta-RL is broader than few-shot learning, since meta-RL also includes many-shot problems, and meta-RL additionally requires a task-distribution for learning in the few-shot setting.
\citet{parker2022automated} survey methods for AutoRL, but also include non-learned algorithms for adapting to new MDPs.
\citet{kirk2023survey} survey methods for generalization in RL, which includes meta-RL methods, but also includes both non-learned algorithms and problems that do not require adaptation, i.e., problems where a single policy can perform optimally without specialization to the given task.
Of all the related machine learning fields, meta-supervised learning (meta-SL) and multi-task RL, are the most closely related.

\paragraph{Meta-Supervised Learning}
There are many points of similarity between meta-RL and meta-SL research.
Like meta-RL, meta-SL considers a distribution of tasks.
However, whereas the tasks in meta-RL are defined by MDPs, the tasks in meta-SL are defined by fixed datasets.
Some algorithms have been proposed as methods for both meta-SL and meta-RL~\citep{finn2017model, mishra2018a}.
Moreover, many algorithmic choices have analogues in both settings.
For example, in the few-shot setting, the use of attention over prior states in meta-RL \citep{ritter2018been, fortunato2019generalization, team2023human}, has analogues in the kernel-based methods of supervised meta-SL \citep{santoro2016meta, vinyals2016matching, sung2018compare}.
These methods are covered later in Section~\ref{sec:task_inf}.
Additionally, in the many-shot setting, meta-learning optimizers and objective functions has been covered in both meta-RL \citep{houthooft2018evolved,kirsch2019improving,oh2020discovering,bechtle2021meta,lu2022discovered, lan2023learning} and meta-SL \citep{andrychowicz2016learning, li2017learning, bechtle2021meta}.
These methods are covered in Section~\ref{subsection:long-horizon-methods}.
Finally, imitation learning is considered a closely related pursuit to RL, since both consider sequential decision making problems.
Therefore, we discuss meta-supervised learning for imitation learning in Section~\ref{sec:supervision}.

In addition to similarities in the problem settings and methods, meta-RL and meta-SL have some key differences.
In particular, meta-SL generally considers fixed datasets while meta-RL does not.
In meta-RL, the adapted policy is responsible for collecting more data, which is fed back into the dataset for adaptation.
This introduces a problem of data collection for adaptation to the given task.
Moreover, since the meta-RL agent must adapt quickly, this requires that the agent also explores quickly.
Fast exploration and the resulting challenges are unique to meta-RL and are a key theme discussed at length in Section~\ref{sec:fast_adaptation}.

\paragraph{Multi-Task RL} Multi-task RL, like meta-RL, considers a distribution of MDPs.
However, whereas a meta-RL agent must identify the MDP that it encounters, in multi-task RL, the agent has access to a ground-truth representation of the task.
While some solutions from multi-task RL, such as PopArt \citep{hessel2019multi}, are immediately applicable to meta-RL \citep{grigsby2023amago}, differences in evaluation also prevent some methods from being applicable.
For example, multi-task RL is often evaluated over a finite set of discrete tasks, whereas meta-RL agents often encounter new tasks at test time \citep{yu2020meta, teh2017distral}.
This allows multi-task methods to train separate networks for each task \citep{teh2017distral}, whereas such approaches are not immediately applicable in meta-RL.

Generally, multi-task RL can be seen as an easier version of meta-RL, in the sense that the MDP representation is known, so there is no need to learn it from data, nor to explore to get data for adaptation.
In fact, an entire category of meta-RL methods tries to explicitly infer the MDP identity in order to reduce the meta-RL problem to the easier multi-task RL problem, which is discussed in Section \ref{sec:task_inf}.
However, there can also be cases in which meta-RL is possible and multi-task RL is not.
Since generalization to different tasks in multi-task RL occurs without conditioning on data at test time, it is always zero-shot generalization.
Such generalization may not be possible, in practice, if the task representation is not sufficiently informative.
For example, if the task is represented as a one-hot categorical vector, then it may be impossible to generalize to categories not seen during training.
However, while multi-task RL would fail, meta-learning may still be viable in this case, if a more informative representation of the task can be inferred from data, or if additional learning at test time can compensate for sparsity in the training distribution.

\chapter{Few-shot Meta-RL}
\label{sec:fast_adaptation}
In this section, we discuss few-shot adaptation, where the agent meta-learns across multiple tasks and at meta-test time must quickly adapt to a new, but related task in a few timesteps or episodes.

\enlargethispage{\baselineskip}
As a concrete example, recall the robot chef learning to cook in home kitchens.
Training a new policy to cook in each user's home using reinforcement learning from scratch would require many samples in each kitchen, which can be wasteful as general cooking knowledge (e.g., how to use a stove) transfers across kitchens.
Wasting many data samples in the homes of customers who purchased a robot may be unacceptable, particularly if every action the robot takes risks damaging the kitchen.
Meta-RL can automatically learn a procedure, from data, for adapting to the differences that arise in new kitchens (e.g., the location of the cutlery).
During meta-training, the robot may train in many different kitchens in simulation or a setting with human oversight and safety precautions.
Then, during meta-testing, the robot is sold to a customer and deployed in a new kitchen, where it must quickly learn to cook in it.
However, training such an agent with meta-RL involves unique challenges and design choices.

In particular, here we summarize the literature in terms of the inner-loop, the exploration, the supervision, and whether the RL algorithm is model-based or model-free.
We categorize by the type of inner-loop, since the majority of research considers different inner-loop parameterizations.
We consider exploration, since learning exploration is unique to meta-RL compared to meta-SL, as discussed in Section~\ref{subsec:related}.
We classify closely related problem settings, which utilize imitation learning, and some other forms of supervision, by the supervision available.
Finally, we survey model-based approaches to meta-RL, since they present different trade-offs to model-free methods.
However, we only discuss model-based methods briefly, as most meta-learning literature concerns the model-free setting.

\paragraph{Meta-parameterization}
In this section, we first discuss three common categories of methods in the multi-task few-shot setting.
Methods in these categories are further classified in Table~\ref{tab:short_horizon_methods}.
Recall that meta-RL itself learns a learning algorithm $f_\theta$.
This places unique demands on $f_\theta$ and suggests particular representations for this function.
We call this design choice the \textit{meta-parameterization}, with the most common ones being the following:

\begin{itemize}
    \item \textbf{Parameterized policy gradient} methods build the structure of existing policy gradient algorithms into $f_\theta$. We review these in Section~\ref{sec:ppg}.
    \item \textbf{Black box} methods impose little to no structure on $f_\theta$. We review these in Section~\ref{sec:black_box}.
    \item \textbf{Task inference} methods structure $f_\theta$ to explicitly infer the unknown task. We review these in Section~\ref{sec:task_inf}.
\end{itemize}

\begin{table}[ht!]
	\centering
    \footnotesize
    \caption{
    Representative examples of few-shot meta-RL research categorized by method. The majority of methods fall into one of three clusters: Parameterized policy gradient, black box, or task inference. These categories determine how the inner-loop is parameterized. Within each cluster, we further categorize the methods. Explanations of these methods can be found in Sections \ref{sec:ppg}, \ref{sec:black_box}, and \ref{sec:task_inf}, respectively.}
    \label{tab:short_horizon_methods}
    \resizebox{\textwidth}{!}{
    \begin{tabular}{C{0.2\textwidth}C{0.8\textwidth}}
            \multicolumn{2}{c}{\textbf{Parameterized Policy Gradients}}  \\ \hline
            MAML-like & \citet{finn2017model,zintgraf2019fast,park2019meta,raghu2020rapid} \\ \hline
            Adapt Policy Distribution & \citet{yoon2018bayesian,gupta2018meta,Wang2020Bayesian,zou2020gradientem,ghadirzadeh2021bayesian} \\ \hline
            Meta-gradient estimation & \citet{al2017continuous,stadie2018some,foerster2018dice,fallah2020provably,tang2022biased} \\
            \hline
            Alternative outer-loop algorithms & \citet{sung2017learning,medonca2019guided,song2020maml} \\
            \hline \hline
            \multicolumn{2}{c}{\textbf{Black Box}}  \\ \hline
            RL2-Like & \citet{heess2015memory,duan2016rl,wang2016learning} \\ \hline
            Adapt Policy Using Context Vector & \citet{duan2016rl,wang2016learning,zintgraf2019fast,humplik2019meta,zintgraf2020varibad,fakoor2020meta,liu2021decoupling} \\ \hline
            Inner-Loop with Hebbian Learning & \citet{miconi2018differentiable,miconi2018backpropamine,najarro2021metalearning,chalvidal2022metareinforcement,rohani2022bimrl} \\ \hline
            Inner-Loop with Attention & \citet{oh2016control,ritter18been,mishra2018a,fortunato2019generalization,ritter2021rapid,luckeciano2022transformers,xu2022prompting,team2023human,elawady2024relic} \\
            \hline \hline
            \multicolumn{2}{c}{\textbf{Task Inference}}   \\ \hline
            Multi-task pre-training & \citet{humplik2019meta,kamienny2020learning,raileanu2020fast,peng2021linear,liu2021decoupling} \\ \hline
            Without privileged information & \citet{guo2018neural,rakelly2019efficient,zintgraf2020varibad,fu2021towards,zhang2021metacure,luo2022adapt} \\ \hline
            Permutation invariance & \citet{rakelly2019efficient,raileanu2020fast,korshunova2020exchangeable,imagawa2022off,beck2024splagger} \\ \hline
            Adapt Policy Using Hypernetworks & \citet{peng2021linear, beck2022hyper, beck2023recurrent} \\
            \hline \hline
    \end{tabular}}
\end{table}

While each of these categories represents a discrete cluster of research, other authors cluster the research differently and may use different names to refer to these clusters.
For example, parameterized policy gradient methods are sometimes referred to as \textit{gradient-based} methods \citep{finn2017model,rakelly2019efficient}, or are referred to as part of a larger category of methods in meta-SL called \textit{optimization-based} methods \citep{hospedales2020meta}.
Additionally, black box methods and task-inference methods are sometimes referred to collectively as \textit{context-based} methods \citep{rakelly2019efficient}.

Along with the parameterization of the inner-loop, related design choices include the representation of the base policy and the choice of the outer-loop algorithm.
Each method must additionally specify which base policy parameters are adapted by the inner-loop, and which are meta-parameters.
We call this design choice the adapted policy parameters.
For each of the three types of meta-parameterizations in this section, we discuss the inner-loop, the outer-loop, and the adapted policy parameters.

\paragraph{Exploration}
While all these methods are distinct, they also share some challenges.
One such challenge is that of \emph{exploration}, the process of collecting data for adaptation.
In few-shot learning, exploration determines how an agent takes actions during its few shots.
Subsequently, an agent must decide how to adapt the base policy using this collected data.
For the adaptation to be sample efficient, the exploration must be efficient as well.
Specifically, the exploration must target differences in the task distribution (e.g., different locations of cutlery).
In Section~\ref{sec:explore_basics}, we discuss the process of exploration, along with ways to add structure to support it.
While all few-shot methods must learn to explore an unknown MDP, there is an especially tight relationship between exploration methods and task inference methods.
In general, exploration may be used to enable better task inference, and conversely, task inference may be used to enable better exploration.
One particular way in which task inference may be used to enable better exploration is by quantifying uncertainty about the task, and then choosing actions based on that quantification.
This method may be used in order to learn optimal exploration.
We discuss this in Section~\ref{sec:optimal_exploration}.

\paragraph{Supervision}
Meta-RL methods also differ in the assumptions that they make about the available
 \emph{supervision}. In the standard meta-RL problem setting, rewards are available during both meta-training and meta-testing.
However, providing rewards in each phase presents challenges.
For example, it may be difficult to manually design an informative task distribution for meta-training, and it may be impractical to measure rewards with expensive sensors during deployment for meta-testing.
In this case, unique methods must be used, such as automatically designing rewards for the outer-loop, or creating an inner-loop that does not need to condition on rewards.
Alternatively, supervision that is more informative than rewards can be provided.
We review such settings, challenges, and methods in Section~\ref{sec:supervision}.

\paragraph{Model-Based Meta-RL}
Some meta-RL methods explicitly learn a model of the MDP dynamics and reward function.
Such methods are called \emph{model-based} methods, in contrast to  \emph{model-free} methods that do not explicitly learn a model of the environment.
Model-based methods confer advantages such as sample-efficient and off-policy meta-training.
Moreover, using an off-the-shelf planning algorithm with the learned model can be easier for some task distributions than learning a complicated policy directly.
However, model-based methods generally require the implementation of additional components and can have lower asymptotic performance.
We discuss these trade-offs and the applications of model-based RL to meta-RL in Section~\ref{subsection:model-based}.
We keep discussion in this section brief, since most meta-RL literature considers mode-free RL, the relevant trade-offs are similar in meta-RL and RL at large, and most model-based meta-RL methods have an analogous model-free method that will have already been discussed by that point.

\paragraph{Theory of Meta-RL}
Finally, we consider theoretical research on meta-RL.
While meta-RL is a relatively new area of research, several studies have found interesting theoretical challenges in it.
These insights relate well to the empirical findings, which constitute the majority of the research in meta-RL.
In Section~\ref{subsec:theory}, we survey important theoretical works in meta-RL.
We believe there is a lot more to explore in the theory of meta-RL and therefore hope this survey can motivate future research.

\section{Parameterized Policy Gradient Methods}\label{sec:ppg}
\enlargethispage{\baselineskip}
\looseness=-1
Meta-RL learns a learning algorithm $f_\theta$, the inner-loop.
This inner-loop is learned to maximize the meta-RL objective given by Equation \ref{eq:meta-rl-objective}, over samples from a task distribution, $p(\mathcal{M})$.
However, during training, we generally only assume access to samples from this distribution.
These samples may be limited, and the distribution over which we want to generalize may be broad.
Moreover, the distribution on which the meta-RL is evaluated for meta-testing may differ from the distribution on which the meta-RL agents is meta-trained.
Each of these motivates the need for learning to take place in the inner-loop algorithm $f_\theta$.
This section discusses methods for building such structure into the inner-loop itself.

We call the parameterization of $f_\theta$ the \emph{meta-parameterization}.
In this section, we discuss one way of parameterizing the inner-loop that builds in the structure of existing standard RL algorithms.
\emph{Parameterized policy gradients (PPG)} are a common class of methods which parameterize the learning algorithm $f_\theta$ as a policy gradient algorithm.
These algorithms generally have an inner-loop of the form
\begin{equation*}
\phi_{j+1} = f_\theta(\phi_{j}, \D_j) = \phi_j + \alpha_\theta \nabla_{\phi_j} \hat{J}_\theta(\D_j, \pi_{\phi_j}),
\end{equation*}
where $\hat{J}_\theta(\D_j, \pi_{\phi_j})$ is an estimate of the returns of the policy $\pi_{\phi_j}$.
For example, recall the MAML algorithm discussed in Section~\ref{sec:example_algos}.
In MAML, $\theta=\phi_0$, and so the initialization is the meta-learned component.
It is also possible to add additional pre-defined components to the inner-loop, such as sparsity-inducing regularization \citep{lou2021meta}.
In general, whatever structure is not predefined, is a parameter in $\theta$ that is meta-learned.
In addition to initialization, the meta-learned structure can include components such as hyper-parameters \citep{li2017meta}.
Some methods also meta-learn a preconditioning matrix to approximate the curvature of the objective, inspired by second-order optimization methods.
These methods generally have the form $ \phi_{j+1} = \phi_j + \alpha_\theta M_\theta \nabla_{\phi_j} \hat{J}_\theta(\D_j, \pi_{\phi_j}) $ \citep{park2019meta,flennerhag2020meta}.
While a value based-method could be used to parameterize the inner-loop instead of a policy gradient~\citep{zou2021learning}, value based-methods generally require many more steps to propagate reward information \citep{mitchell2020offline}, and so are typically reserved for the many-shot setting, discussed in Section~\ref{sec:long_task_horizon_meta_rl}.
In this section, we focus on PPG methods.
We begin by discussing different parameters of the base policy that the inner-loop may adapt.
Then, we discuss options for outer-loop algorithms and optimization.
We conclude with a discussion of the trade-offs associated with PPG methods.

\paragraph{Adapted policy parameters}
PPG algorithms commonly meta-learn an initialization, and then adapt that initialization in the inner-loop.
Instead of adapting a single initialization, several PPG methods learn a full distribution over initial policy parameters, $p(\phi_0)$ \citep{yoon2018bayesian,gupta2018meta,Wang2020Bayesian,zou2020gradientem,ghadirzadeh2021bayesian}.
This distribution allows for modeling uncertainty over policies.
The distribution over initial parameters can be represented with a finite number of discrete particles \citep{yoon2018bayesian}, or with a Gaussian approximation fit via variational inference \citep{gupta2018meta,ghadirzadeh2021bayesian}.
Moreover, the distribution itself can be updated in the inner-loop, to obtain a posterior over (a subset of) model parameters \citep{yoon2018bayesian,gupta2018meta}.
The updated distribution may be useful both for modeling uncertainty \citep{yoon2018bayesian}, and for temporally extended exploration, if policy parameters are resampled periodically \citep{gupta2018meta}.

\begin{figure}[ht!]
    \centering
    \includegraphics[height=4cm,alt={Illustration of meta-RL without (left) and with (right) a context vector.}]{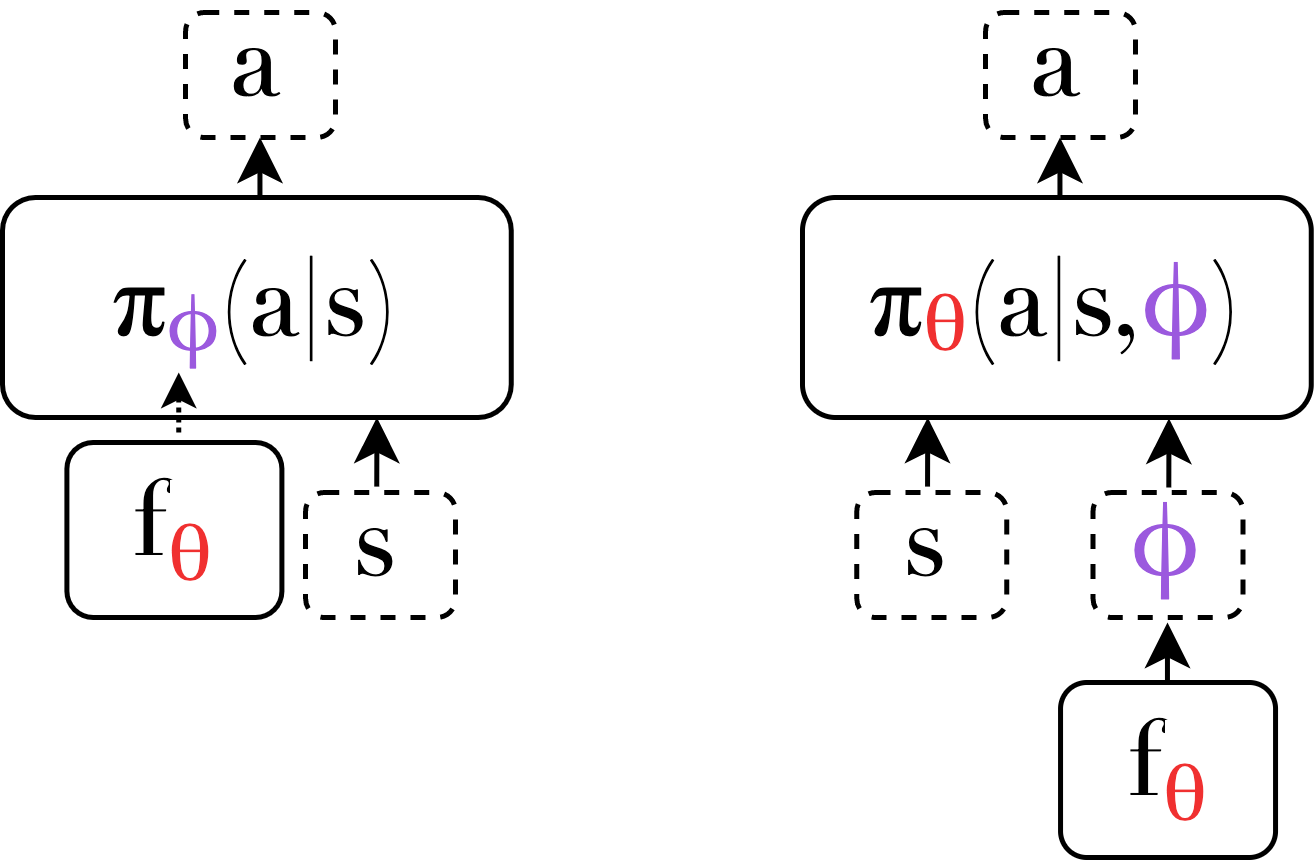}
    \caption{
        Illustration of meta-RL without (left) and with (right) a context vector. The context vector can be updated with backpropagation in a PPG method or by a neural network in a black box method. When using a context vector, some of the policy parameters are not adapted by the inner-loop, and are instead meta-parameters set by the outer-loop.
    }
    \label{fig:context_vector}
\end{figure}

Alternately, some PPG methods adapt far fewer parameters in the inner-loop.
Instead of adapting all policy parameters, they adapt a subset \citep{zintgraf2019fast,raghu2020rapid}.
For example, one method adapts only the weights and biases of the last layer of the policy \citep{raghu2020rapid}, while leaving the rest of the parameters constant throughout the inner-loop.
Another method adapts only a vector, called the \textit{context vector}, on which the policy is conditioned \citep{zintgraf2019fast}.
In this case, the input to the policy itself parameterizes the range of possible behavior.
We write the policy as $\pi_\theta(a|s,\phi)$, where $\phi$ is the adapted context vector.
The weights and biases of the policy, as well as the initial context vector, constitute the meta-parameters that are constant in the inner-loop.
We visualize the use of a context vector in Figure~\ref{fig:context_vector}.

\begin{sloppypar}
Consider the algorithm Context Adaptation via Meta-learning (CAVIA) \citep{zintgraf2019fast}.
CAVIA is similar to MAML but it only adapts a context vector, which is initialized to the zero vector:
\end{sloppypar}

\noindent
\begin{align*}
& \phi_{0} = \mathbf{0}, \\
& \phi_{j+1} = \phi_j + \alpha_\theta \nabla_{\phi_j} \hat{J}_\theta(\D_j, \pi_{\theta}(\cdot|\phi_j)).
\end{align*}
While the update to $\phi$ is the same as in MAML, here, $\phi$ represents a vector passed as an input to the policy network, while all the weights and biases of the network remain fixed throughout adaptation.
One benefit of adapting a subset of parameters in the inner-loop is that it may mitigate overfitting in the inner-loop, for task distributions where only a small amount of adaption is needed \citep{zintgraf2019fast}.

\paragraph{Meta-gradient estimation in outer-loop optimization}
Effective learning in the outer-loop of PPG algorithms requires access to estimates of \emph{meta-gradients}, i.e., gradients of the outer-loop objective with respect to the meta-parameters.
Most commonly, the meta-gradients are computed by directly optimizing Equation~\ref{eq:meta-rl-objective} with a policy gradient algorithm.
However, naively applying a policy gradient method in the outer-loop can lead to a suboptimal bias-variance trade-off.
Significant research has considered how to improve this trade-off when estimating  meta-gradients for PPG methods~\citep{stadie2018some,foerster2018dice,rothfuss2018promp,mao2019baseline,liu2019taming,fallah2020provably,vuorio2021no,tang2022biased,fallah2021convergence,tang2021unifying,tack2022metalearning,liu2022a}.
Next, we discuss the bias-variance trade-offs in meta-gradient estimation for PPG methods and the algorithms arising from choosing different points in the trade-off.

The unbiased gradient of the meta-RL objective given by Equation~\ref{eq:meta-rl-objective}
with respect to the meta-parameters $\theta$ can be computed as follows.
Since the distribution of tasks does not depend on the meta-parameters, we can estimate the meta-gradient with respect to each task separately and then combine.
The expression for the meta-gradient for a single task is given by

\noindent
\begin{align*}
    \nabla_{\theta} \E_{\mathcal{D}} \bigg[ \sum_{k=K}^{H} G(\D_k) \bigg| \pi_{f_\theta} \bigg]
    =\nabla_{\theta} \sum_{k=K}^{H} \int G(\D_k) \prod_{i=0}^{k} p(\D_i ; \phi_i) d \D_i.
\end{align*}
Here, we take into account that the returns from previous trajectories are not influenced by the policies of future trajectories. Therefore, we restricted the product of trajectory probabilities to range only from $i=0$ to $k$.
Since the probabilities of the trajectories depend on $\theta$, we use the log-derivative trick to get the gradient of the product
\begin{align*}
    &= \sum_{k=K}^{H} \int G(\D_k) \prod_{i=0}^{k} p(\D_i ; \phi_i) \nabla_{\theta} \log  \prod_{i=0}^{k} p(\D_i ; \phi_i)  d \D_i.
\end{align*}
Finally, we use the chain rule to get the meta-gradient:
\begin{align*}
    &= \sum_{k=K}^{H} \int G(\D_k) \prod_{i=0}^{k} p(\D_i ; \phi_i) \sum_{j=0}^{k} \nabla_{\theta} \phi_j \nabla_{\phi_j} \log p(\D_j ; \phi_j)  d \D_i
\end{align*}
Naively applying a policy gradient algorithm to the returns $\sum_{k=K}^{H} G(\D_k)$ does not compute this meta-gradient.
To see why, we reorganize the terms to get

\small
\vspace{-0.8\baselineskip}
\begin{align*}
    &= \sum_{k=K}^{H} \mathbb{E}_{\mathcal{D}} \bigg[ G(\D_k) \bigg( \nabla_{\theta} \phi_k \nabla_{\phi_k} \log p(\D ; \phi_k)
    + \sum_{j=0}^{k-1} \nabla_{\theta} \phi_j \nabla_{\phi_j} \log p(\D_j ; \phi_j) \bigg)
    \bigg],
\end{align*}\normalsize

\noindent where the first gradient term is the regular policy gradient on the $k$th trajectory w.r.t.\ the meta-parameters, and the gradient terms in the sum are the gradients of the policies on the $j$th trajectories for $0 \leq j \leq k-1$ w.r.t.\ the meta-parameters.
These latter terms, sometimes called \emph{sampling-correction} terms, are often omitted in practice, yielding biased meta-gradient estimates~\citep{al2017continuous,stadie2018some}.

Recall that PPG methods produce a sequence of policies, and generally optimize Equation~\ref{eq:meta-rl-objective} with $K=H-n$, where $n$ is the number of episodes collected by the final policy.
Actions taken by any one policy affect the data seen by the next, and thus these actions affect the expected return of the later policies.
The sampling-correction terms account for this dependence.
Ignoring them amounts to ignoring actions from earlier policies when training the inner-loop to optimize the meta-RL objective and thus introduces bias to the meta-gradient estimation~\citep{al2017continuous}.
This bias can sometimes be detrimental to the meta-learning performance but the high variance of the unbiased estimator often dominates~\citep{stadie2018some,fallah2020provably,vuorio2021no}.

Significant research has considered a meta-gradient estimator derived originally for computing higher-order meta-gradients~\citep{foerster2018dice,rothfuss2018promp,farquhar2019loaded,liu2019taming,mao2019baseline,tang2021unifying,tang2022biased,liu2022theoretical}.
In practice, PPG algorithms use a sample-based policy gradient algorithm for updating the policy parameters in the inner-loop.
The higher-order meta-gradients of the sample-based inner-loop are different from those of an inner-loop that uses the expected policy gradient.
\citet{foerster2018dice,rothfuss2018promp} argue that the higher-order meta-gradients of the sample-based inner-loop should approximate those of the expected inner-loop and derive alternative sample-based inner-loop update functions for that purpose.
While these meta-gradient estimators are still biased, their variance has a more benign dependence on the inner-loop sample size than the unbiased meta-gradient estimator and therefore can achieve a better point in the bias-variance trade-off than either the naive or the unbiased meta-gradient estimators~\citep{tang2021unifying,liu2022theoretical}.
To further reduce the variance of this class of meta-gradient estimators, \citet{rothfuss2018promp,farquhar2019loaded,liu2019taming,mao2019baseline} propose to ignore certain high-variance terms in the estimator and introduce baselines.

\enlargethispage{\baselineskip}
Alternatively, some PPG methods do not require higher-order meta-gradients.
For example, \citet{finn2017model} use a first-order approximation of the meta-gradient, whereas \citet{song2020maml} use gradient-free optimization
(see \cite{nichol2018first} for another first-order approximation proposed for supervised meta-learning, and recently used in meta-RL, \cite{ren2022leveraging}.)
Furthermore, when meta-learning an initialization as in MAML, for a limited number of tasks or limited amount of data in the inner-loop, it may even be preferable to set the inner-loop to take no steps at all during meta-training, i.e., without adapting to each task \citep{gao2020modeling}.
In this case, task adaptation occurs only through fine-tuning at test-time.
The lack of explicit meta-learning can be seen as a limitation of the meta-learning approach and is discussed further in Section~\ref{sec:open_problems}.
Additionally, it is possible to use a value-based algorithm in the outer-loop, instead of a policy gradient algorithm, which avoids meta-gradients altogether \citep{sung2017learning}.

\paragraph{Outer-loop algorithms}
While most PPG methods use a policy-gradient algorithm in the outer-loop, other alternatives are possible \citep{sung2017learning,medonca2019guided}.
For example, one can train a critic, $Q_\theta(s,a,\mathcal{D})$, using-TD error in the outer-loop, then reuse this critic in the inner-loop \citep{sung2017learning}.
Alternatively, instead of estimating meta-gradients via backpropagation, \emph{evolution strategies} (ES)~\citep{rechenberg1971evolutionsstrategie,wierstra2014natural,salimans2017evolution} may be used~\citep{song2020maml}.
Additionally, one can train task-specific experts and then use these for imitation learning in the outer-loop.
While neither directly learn exploratory behavior by optimizing Equation~\ref{eq:meta-rl-objective}, they can work in practice.
We provide a more complete discussion of outer-loop supervision in Section~\ref{sec:supervision}.

\begin{figure}[ht!]
    \centering
    \includegraphics[width=\textwidth,alt={Out-of-distribution generalization metrics for MAML and RL2 on environment variants from Figure 1.1.}]{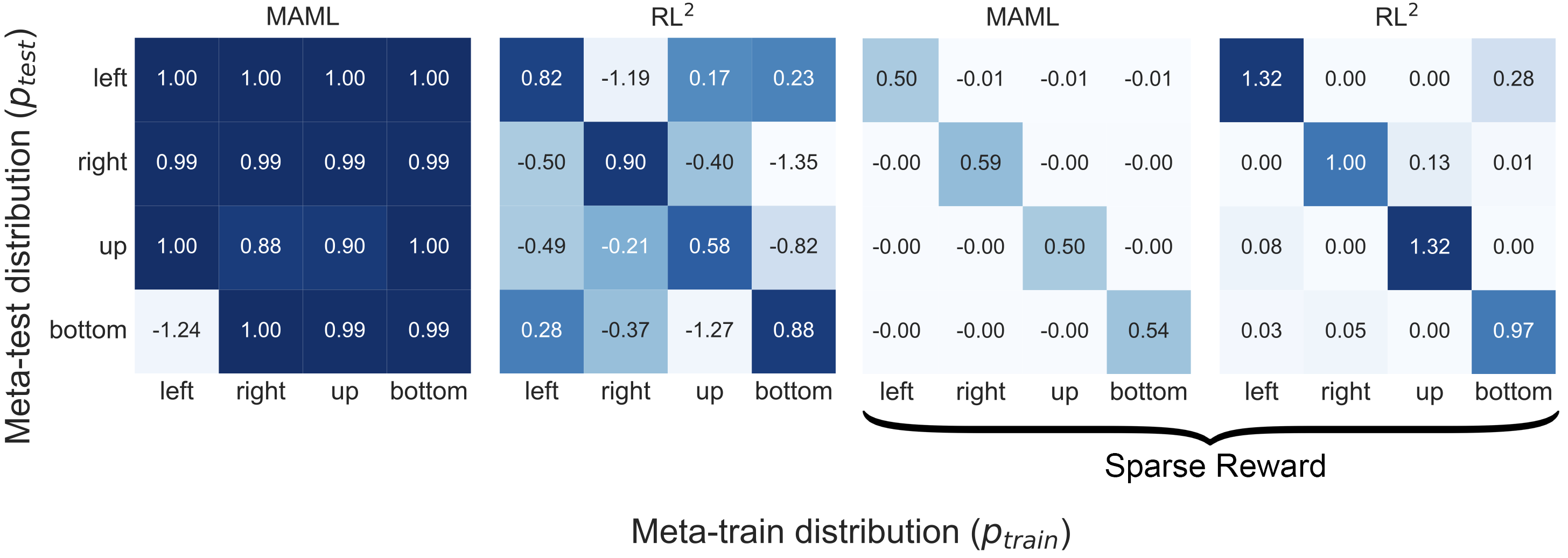}
    \caption{
    Out-of-distribution generalization metrics reproduced from \citet{xiong2021on}. MAML and RL$^2$ are evaluated on variants of the environment depicted in Figure \ref{fig:intro_example_short}, with greater scores (darker blue) indicating a greater degree of recovered performance. While the canonical PPG method, MAML, is able to generalize better to out-of-distribution than the black-box method, RL$^2$, on environments with dense reward, MAML generalizes worse on some environments in practice. MAML performs especially poorly on sparse reward environments requiring systematic exploration that is highly specific to the task distribution.
    }
    \label{fig:consistent}
\end{figure}

\paragraph{PPG trade-offs}\enlargethispage{-\baselineskip}
One of the primary benefits of PPG algorithms is that they produce an inner-loop that converges to a locally optimal policy, even when relatively few samples are available for meta-training or when the task distribution differs from meta-training to meta-testing.
PPG inner-loops are generally guaranteed to converge under the same assumptions as standard policy gradient methods.
For example, the MAML inner-loop converges with the same guarantees as REINFORCE \citep{williams1992simple}, since it simply runs REINFORCE from a meta-learned initialization.
Even convergence bounds are possible \citep{fallah2021convergence}.
However, as with any stochastic gradient algorithm, the guarantees given by REINFORCE are rather weak.
Stochastic gradient algorithms, with a sufficiently small learning rate, only converge to a local optimum, and this can be problematic in practice in meta-RL \citep{xiong2021on}.
Moreover, in some cases, a step of REINFORCE can even make the policy worse on the task~\citep{deleu2018effects}.
We reproduce empirical results from \citet{xiong2021on} in Figure \ref{fig:consistent}.
While the PPG method, MAML, eventually converges to performant policies on many out-of-distribution tasks, it fails to do so in practice on some environments, such as those with sparse rewards.
Moreover, \citet{xiong2021on} show that black-box methods can be made to converge similarly in practice by continuing to meta-train with gradient steps at meta-test time.
Having a learning algorithm that eventually adapts to a novel task is desirable, since it reduces the dependence on seeing many relevant tasks during meta-training.

Although parameterizing $f_\theta$ as a policy gradient method may ensure that adaptation generalizes, this structure also presents a trade-off.
Typically, the inner-loop policy gradient has high variance and requires a value estimate covering the full episode, so estimating the gradient generally requires many episodes.
Hence, PPG methods are generally not well-suited to few-shot problems that require stable adaption at every timestep or within very few episodes in the inner-loop.
This can result in decreased performance relative to black-box methods, as seen, for example, in the sparse reward environments that require highly specialized and systematic exploration, as depicted in Figure \ref{fig:consistent}.
Moreover, PPG methods are often sample-inefficient during meta-training as well, because the outer-loop generally relies on on-policy evaluation, rather than an off-policy method that can reuse data efficiently.

\begin{figure}[ht!]
    \centering
    \includegraphics[height=4cm,alt={The meta-parameterization spectrum, from structured methods to black-box methods.}]{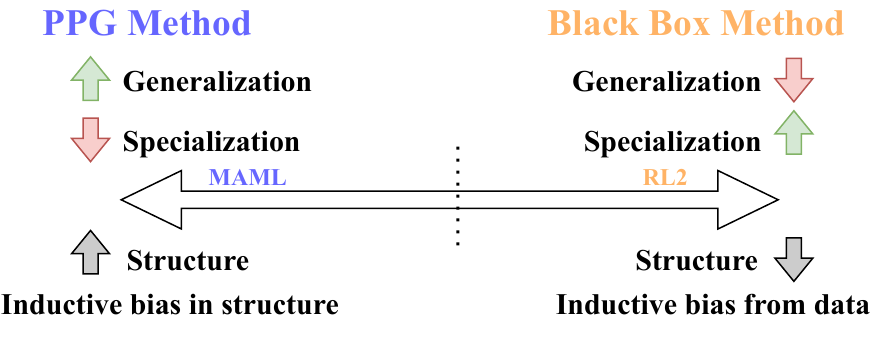}
    \caption{
        The meta-parameterization spectrum. On the left are methods like MAML that hard-code the structure of policy gradients into the inner-loop. Such methods generalize better. On the right are black box methods, which learn all of the structure of the inner-loop from data. Such methods can specialize to the task distribution better.
    }
    \label{fig:param_spectrum}
\end{figure}

In general, there is a trade-off between generalization to novel tasks and specialization over a given task distribution.
How much structure is imposed by the parameterization of $f_\theta$ determines where each algorithm lies on this spectrum.
The structure of PPG methods places them near the end of the spectrum that ensures generalization.
This is visualized in Figure~\ref{fig:param_spectrum}.
While this spectrum summarizes current methods, the trade-off is not necessarily inherent to the problem setting, and future work could investigate methods that achieve the best of both worlds.
In the next section, we discuss methods at the other end of the spectrum.

\section{Black Box Methods}\label{sec:black_box}

In contrast to PPG methods, \emph{black box methods} are near the other end of the spectrum.
Black box methods represent the inner-loop, $f_\theta$, as a sequence model to process data for learning.
Learning in the inner-loop occurs solely in the activations of this sequence model, a phenomenon sometimes called \textit{in-context learning}.
Meta-learning $f_\theta$ is a way to explicitly elicit the phenomenon of in-context learning in a sequence model.
In principle, black-box methods can learn any arbitrary learning procedure, since they represent $f_\theta$ with a neural network as a universal function approximator.
This places fewer constraints on the function $f_\theta$ than with a PPG method.
Since $f_\theta$ represents an arbitrary function of history, the modified policy can represent an arbitrary history-dependent policy.
As discussed in Section \ref{section:background}, such a policy is sufficient for learning an optimal policy for a POMDP, and thus for the meta-RL problem as well.
Recall the RL2 algorithm \citep{duan2016rl} from Section~\ref{sec:example_algos}.
In this case, $f_\theta$ is represented by a recurrent neural network, whose outputs are an input vector to the base policy.
While the use of a recurrent network to output a context vector is common,
black box methods may also structure the base policy and inner-loop in other ways.
In this section we survey black box methods.
We begin by discussing architectures that are designed for different amounts of diversity in the task distribution.
Each of these architectures has different adapted policy parameters.
Then, we discuss distinct ways to structure the inner-loop and alternative outer-loop algorithms.
Finally, we conclude with a discussion of the trade-offs associated with black box methods.

\paragraph{Adapted policy parameters}
On one hand, some task distributions may require little adaptation for each task, as discussed in Section~\ref{sec:ppg}.
For example, if a navigation task varies only by goal location, then it may be that not many policy parameters need to be adapted.
Many black box methods have been applied to such task distributions \citep{wang2016learning,humplik2019meta,zintgraf2020varibad}.
In such a setting, it is sufficient to only adapt a vector, used as an input to the base policy, instead of all parameters of the base policy \citep{duan2016rl,wang2016learning,zintgraf2019fast,humplik2019meta,zintgraf2020varibad,fakoor2020meta,liu2021decoupling}.
This vector, $\phi$, is called a \textit{context vector}, just as in PPG methods.
This distinction is visualized in Figure~\ref{fig:context_vector}.
The context is produced by a recurrent neural network, or any other network that conditions on history, such as a transformer or memory-augmented  network.
Using a recurrent neural network, the inner-loop can be written $f_\theta(\mathcal{D})=\textsc{rnn}_\theta(\mathcal{D})$.
As discussed in Section~\ref{subsec:pomdp-formalization}, when $f_\theta$ is represented as a sequence model, $f_\theta$ and $\pi_{f_\theta(\mathcal{D})}(a|s)$ can be viewed together as a single history-dependent policy, $\pi_\theta(a|s,\mathcal{D})$.
Using $\phi$ to represent the RNN hidden state, we can write the recurrent policy as $\pi_\theta(a|s,\phi)$.
This is a common architecture for few-shot adaptation methods \citep{duan2016rl,wang2016learning,humplik2019meta,zintgraf2020varibad,fakoor2020meta,liu2021decoupling}.

\begin{figure}[ht!]
    \centering
    \includegraphics[height=5cm,alt={Illustration of black-box meta-RL with a hypernetwork (left) and a context vector (right), showing how an RNN generates policy parameters or a context vector for conditioning.}]{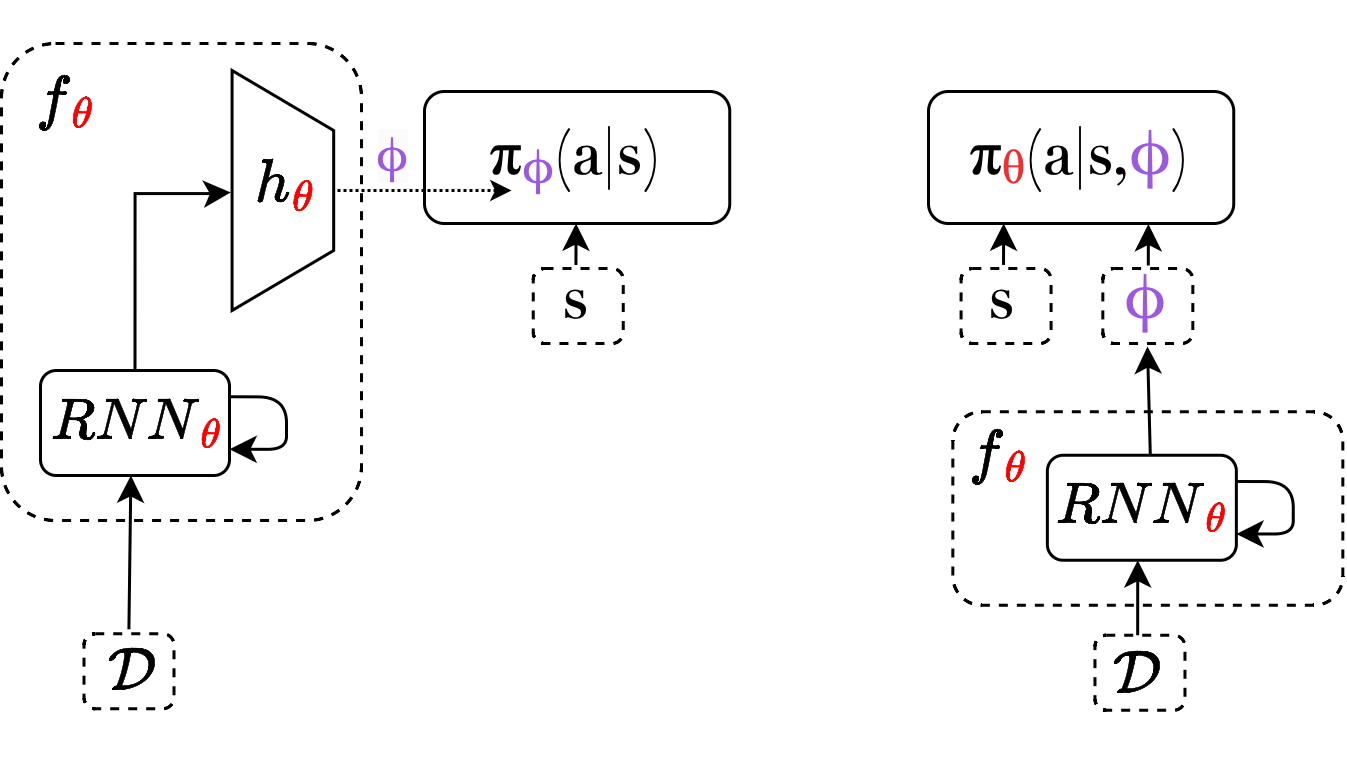}
    \caption{
        Illustration of black box meta-RL with hypernetwork (left) and with a context vector (right). When the inner-loop is a recurrent neural network, generating all the parameters of the policy, including its weights and biases, results in a hypernetwork. Alternatively, the RNN can output a context vector on which the policy is conditioned.
    }
    \label{fig:hypernet}
\end{figure}

On the other hand, some task distributions may require significant differences in behavior between the optimal policies.
By conditioning a policy on a context vector, all of the weights and biases of $\pi$ must generalize between all tasks.
However, when significantly distinct policies are required for different tasks, forcing base policy parameters to be shared may impede adaptation \citep{beck2022hyper}.
Alternatively, just as in PPG methods, there may be no context vector, but the inner-loop may adapt the weights and biases of the base policy, directly \citep{beck2022hyper}.
The weights and biases here are the task parameters, $\phi$, output by the inner-loop.
The inner-loop may produce all of the parameters of a feed-forward base policy \citep{beck2022hyper}, or may modulate the weights of a recurrent base learner \citep{flennerhag2020meta}.
In these cases, using one network to map all data in the trial, $\mathcal{D}_j$,
to weights and biases of another network defines a hypernetwork \citep{ha2017hypernetworks}.
The architecture can be written as $\pi_\phi(a|s)$, where $\phi=h_\theta(\textsc{rnn}_\theta(\mathcal{D}))$, $h$ is a hypernetwork, and $\phi$ are the weights and biases of the policy.
See Figure~\ref{fig:hypernet} for an illustration.

\paragraph{Inner-loop representation}
While many black box methods simply use recurrent neural networks as the inner-loop representation \citep{heess2015memory,duan2016rl,wang2016learning,fakoor2020meta}, alternative representations have also been investigated.
Some of these representations are biologically inspired \citep{bellec2018long, miconi2018differentiable}.
One of the most common biologically inspired representations leverages Hebbian learning \citep{miconi2018differentiable,miconi2018backpropamine,najarro2021metalearning,chalvidal2022metareinforcement,rohani2022bimrl}.
Hebbian learning \citep{hebb1949organization} is a biologically inspired method in which weight updates are a function of the associated activations in the previous and next layers.
The update to the weight ($w^k_{i,j}$) from the $i$th activation in layer $k$ ($x^k_i$), to the $j$th activation in layer $k+1$ ($x^{k+1}_j$) generally has the form
\begin{equation*}
w^k_{i,j} := w^k_{i,j} + \alpha(ax^k_ix^{k+1}_j+bx^k_i+cx^{k+1}_j+d),
\end{equation*}
where $\alpha$ is a learning rate and $\alpha$, $a$, $b$, $c$, $d$ are all meta-learned parameters of the inner-loop learned in $\theta$.
While Hebbian learning is one of the most common biologically inspired representations, most methods aligned with recent machine learning literature use some form of attention \citep{graves2014neural,vaswani2017attention} to parameterize the inner-loop \citep{oh2016control,ritter18been,mishra2018a,fortunato2019generalization,ritter2021rapid,wang2021alchemy,emukpere2021successor,luckeciano2022transformers,xu2022prompting,team2023human, elawady2024relic}.

Attention can be thought of as a soft form of a key-value lookup in a dictionary.
Specifically, it is a mechanism for combining different vectors based on the similarity of their associated key vectors to a given query vector.
Given a query vector, $q$, a matrix, $V$, where each row is one of $n$ value vectors over which to attend, and a matrix $K$, where each row is one of $n$ key vectors, then attention can be written
$$ \mathrm{Attention}(q, K, V) = \sum_{i=0}^{i=n} w_i(q,K_i)V_i $$ where $w \in \Delta^n$ is a weight vector defining the convex combination of value vectors \citep{graves2014neural}.
Generally, attention computes the weights as $w = \mathrm{softmax}(Kq)$, leading to an attention mechanism that simplifies to
$$ \mathrm{Attention}(q, K, V) = V^T\mathrm{softmax}(Kq) = (\mathrm{softmax}(q^TK^T)V)^T .$$
Additionally, computing multiple queries, with each as a row vector in $Q$, we can write this as
$$ \mathrm{Attention}(Q, K, V) = \mathrm{softmax}(QK^T)V .$$

\begin{figure}[ht!]
    \centering
    \includegraphics[trim=450 0 0 0, clip, height=7cm,alt={Example of attention over past recurrent states in the inner-loop of some black-box methods.}]{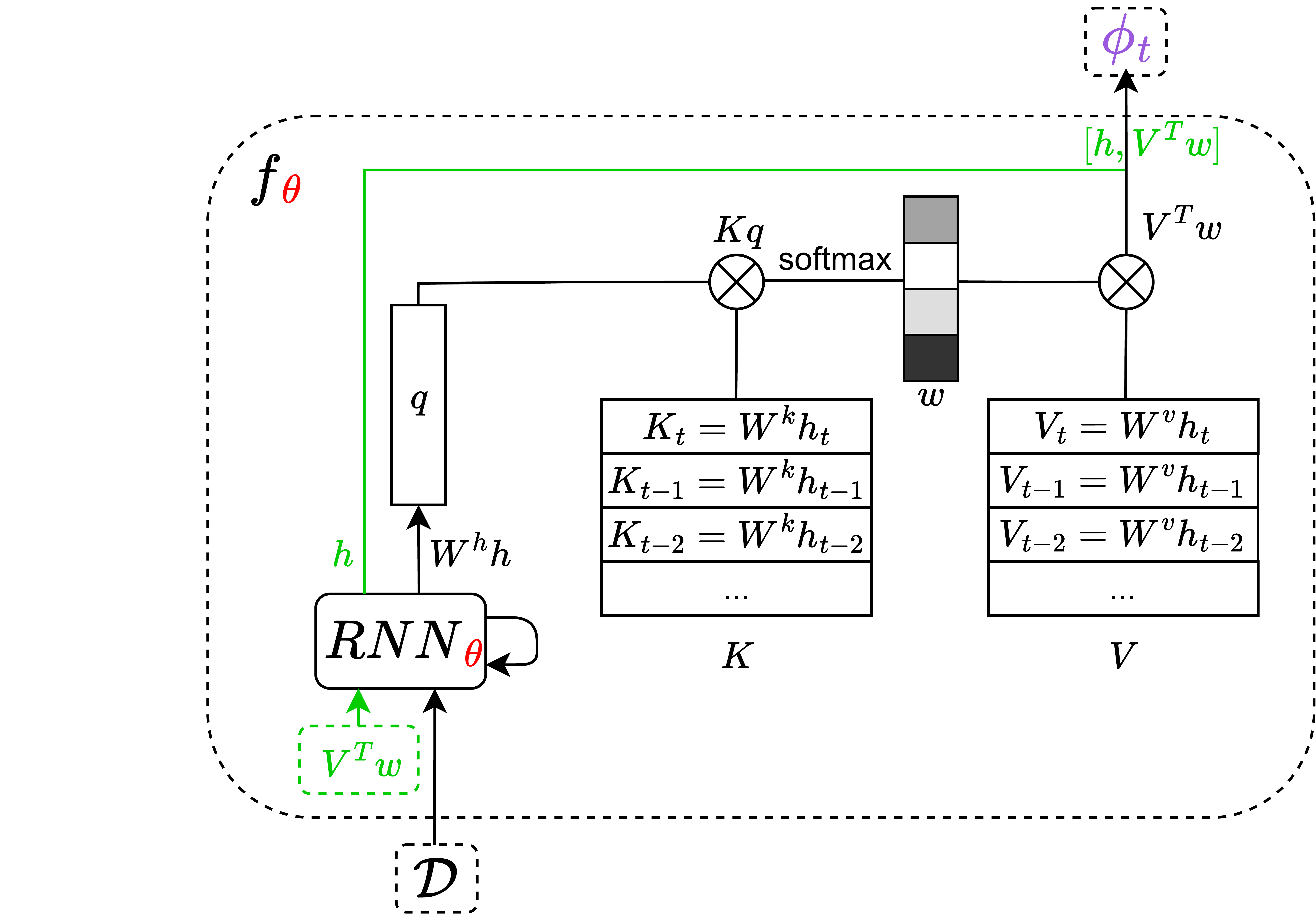}
    \caption{
        Attention over past recurrent states in the inner-loop of black box methods. One approach is to use the current hidden state as a query in attention over past recurrent states. The keys ($K$) and values ($V$) are all linear projections of the past recurrent states. The output, $\phi$, is a a convex combination of the projected hidden states ($V$), where the combination is specified by this weight vector ($w$), computed from the similarity between the keys ($K$) and query ($q$). Components in green are optional. In order to provide long-term context to the RNN, the output from attention over past hidden states, $V^Tw$, can be passed as an input to the RNN at the next timestep. Additionally, the output, $\phi$, can be $V^Tw$, $h$, or a concatenation of both. Passing the RNN state, $h$, as an output, with $V^Tw$ as an input, allows multiple steps of attention to be integrated into the final output.
    }
    \label{fig:attention}
\end{figure}

Some methods combine attention with convolution \citep{mishra2018a}, or use attention over past recurrent states \citep{ritter18been,fortunato2019generalization, team2023human, oh2016control}, while others use self-attention alone \citep{ritter2021rapid,wang2021alchemy,team2023human}.
For example, to attend to past recurrent states, $q$ may be a function of the current hidden state of a recurrent network, while $K$ and $V$ may be (two different linear projections of) all prior hidden states computed over $\mathcal{D}$.
A generalization of such methods can be seen in Figure~\ref{fig:attention}.
In contrast, self-attention models $Q$, $K$, and $V$ all as linear projections of the inputs in $\mathcal{D}$ first, then as linear projections of the previous attention layer, for multiple attention layers in a row.

Attention mechanisms seem to aid in generalization to novel tasks outside of the distribution, $p(\mathcal{M})$, \citep{fortunato2019generalization,luckeciano2022transformers} and self-attention may be useful for complex planning \citep{ritter2021rapid}.
Still, attention is computationally expensive: whereas recurrent networks use $O(1)$ memory and compute per timestep, attention generally requires $O(t^2)$ memory and compute per timestep $t$, which may cross many episodes.
While fast approximations of attention  exist \citep{katharopoulos2020transformers}, solutions in meta-RL often simply maintain only memory of a fixed number of the most recent transitions \citep{mishra2018a,fortunato2019generalization}.
Nonetheless, both transformers, which use self-attention, and recurrent neural networks can still struggle to meta-learn simple inductive biases, particularly for complex task distributions generated by simple underlying rules in low-dimensional spaces \citep{kumar2020meta,chan2022transformers}.

\paragraph{Outer-loop algorithms}
While many black box methods use on-policy algorithms in the outer-loop \citep{duan2016rl,wang2016learning,zintgraf2020varibad},
it is  straightforward to use off-policy algorithms \citep{rakelly2019efficient,fakoor2020meta,liu2021decoupling}, which bring increased sample efficiency to RL.
For discrete actions, it is also straightforward to use $Q$-learning \citep{fakoor2020meta,liu2021decoupling,zhang2022temporal}, in which case the inner-loop must change as well.
In this case, the inner-loop estimates $Q$-values instead of directly parameterizing a policy.
The policy can then act greedily with respect to these $Q$-values at meta-test time.
This approach can be thought of as modifying recurrent $Q$-networks \citep{hausknecht2015deep} to fit the meta-RL setting, which compares favorably compared to other state-of-the-art meta-RL methods \citep{fakoor2020meta}.

\paragraph{Black box trade-offs}
One key benefit of black box methods is that they can
rapidly alter their policies in response to new information, whereas PPG methods generally require multiple episodes of experience to get a sufficiently precise inner-loop gradient estimate.
For example, consider an agent that must learn which objects in a kitchen are hot at meta-test time.
While estimating a policy gradient, a PPG method may touch a hot stove multiple times before learning not to.
In contrast, a black box method may produce an adaptation procedure that never touches a hot surface more than once.
Black box methods can learn such responsive adaptation procedures
because they represent the inner-loop as an arbitrary function that maps from the cumulative task experience to the next action.

However, black-box methods also present a trade-off.
While black-box methods can tightly fit their assumptions about adaptation to a narrow distribution of data, $p(\mathcal{M})$, increasing specialization, they often struggle to generalize outside of  $p(\mathcal{M})$ \citep{wang2016learning,finn2018meta,fortunato2019generalization,xiong2021on}.
Consider the robot chef: while it may learn to not touch hot surfaces, it is unlikely a black box robot chef will learn a completely new skill, such as how to use a stove, if it has never seen a stove during meta-training.
In contrast, a PPG method could still learn such a skill at meta-test time with sufficient data.
While there have been efforts to both broaden the set of meta-training tasks and manually add inductive biases to black-box methods for generalization,
which we discuss in Section~\ref{sec:long_task_horizon_meta_rl}, whether to use a black-box method or PPG method generally depends on the amount of generalization and specialization in the problem to be solved.

Additionally, there exist trade-offs in outer-loop optimization challenges between PPG and black box methods.
On one hand, as discussed in Section~\ref{sec:ppg}, PPG methods often estimate a meta-gradient, which is difficult to compute \citep{al2017continuous}, especially for long horizons \citep{wu2018understanding}.
On the other hand, black box methods do not have the structure of optimization methods build into them, so they can be harder to train from scratch, and have associated outer-loop optimization challenges even for short horizons.
Black box methods generally make use of recurrent neural networks, which can suffer from
 vanishing and exploding gradients \citep{pascanu2013understanding}.
Moreover, the optimization of recurrent neural networks can be especially difficult in reinforcement learning \citep{Beck2020AMRL}, while transformers can be even more problematic to train~\citep{parisotto2020stabilizing,luckeciano2022transformers}.
While large transformers have shown some notable success in meta-RL, such solutions require the use of curriculum learning and distillation to train stably \citep{team2023human}.
Thus, the rapid updates of black box methods in meta-RL, while enabling fast learning, also present
challenges for meta-learning.

Some methods use both PPG and black-box components \citep{vuorio2019multimodal,xiong2021on,ren2022leveraging}.
In particular, even when training a fully black-box method, the policy or inner-loop can be fine-tuned with policy gradients at meta-test time \citep{lan2019meta,xiong2021on,imagawa2022off}.

\section{Task Inference Methods} \label{sec:task_inf}

Closely related to black box methods are \textit{task inference} methods, which often
share the same parameterization as black box methods and thus can be considered a subset of them.
However, parameterizations of the inner-loop may be specific to task inference methods \citep{rakelly2019efficient,korshunova2020exchangeable,zintgraf2020varibad}, which generally train the inner-loop to perform a different function by optimizing a different objective.

Task inference methods generally aim to identify the MDP, or task, to which the agent must adapt, in the inner-loop.
In meta-RL, the agent must repeatedly adapt to an \textit{unknown} MDP whose representation is not given as input to the inner-loop.
While the agent ultimately acts to maximize reward,
the entire purpose of the inner-loop can be described as identifying the task.
The agent's belief about what it should do can be represented as a distribution over tasks.
As discussed in Section \ref{section:background}, this posterior distribution constitutes a sufficient statistic, or information state \citep{subramanian2022approximate}, for the meta-RL problem.
Since we already know the form of this sufficient statistic, the inner-loop can model it directly, instead of learning a mapping from history to action end-to-end.
\textit{Task inference} is the process of inferring this posterior distribution over tasks, conditional on what the agent has seen so far.

Consider the case where the agent has uniquely identified the task.
Then, at this point, the agent knows the MDP and could in fact use classical planning techniques, such as value iteration, to compute the optimal policy directly.
In this scenario, no further learning or data collection is required.
More practically, if the task distribution is reasonably small and finite, we can avoid even having to explicitly plan, by learning a mapping from the task to the optimal policy directly, during meta-training.
In fact, training a policy over a distribution of tasks, with the policy conditioned on the true task, can be taken as the definition of multi-task RL \citep{yu2020meta}.
In the multi-task case, a mapping is learned from a known task to a policy.
In meta-RL the only difference is that the task is not known.
Thus task inference can be seen as an attempt to move a meta-learning problem into the easier multi-task setting.

When uncertainty remains in the distribution, instead of mapping a task to a base policy, task inference methods generally map a task distribution, given the current data, to a base policy.
This can be seen as learning a policy conditioned on a (partially) inferred task.
In this case, learning becomes the process of reducing uncertainty about the task.
The agent must collect data that enables it to identify the task.
That is, the agent must{explore to reduce uncertainty in the posterior given by task inference.
Task inference is therefore a useful way to frame exploration, and many task inference methods are framed as tools for exploration, which we discuss in Section~\ref{sec:explore_basics}.
Optimally handling uncertainty in the task distribution is difficult, and is discussed in Section~\ref{sec:optimal_exploration}.

In this Section we discuss two methods for task inference that use supervised learning but also require assumptions about the information available for meta-training.
We then discuss alternative methods without such assumptions, and how the inner-loops are usually represented.
Finally, we conclude with a discussion of the trade-offs concerning task inference methods.
Table~\ref{tab:short_horizon_methods} summarizes these categories and task inference methods.

\paragraph{Task inference with privileged information}
A straightforward method for inferring the task is to add a supervised loss so that a black box $f_\theta$ predicts some estimate of the task, $\hat{c}_\mathcal{M}$, given some known representation of the task, $c_\mathcal{M}$ \citep{humplik2019meta}.
For example, a recurrent network may predict the task representation conditional on all data collected so far.
Recall that $\phi$, the task parameters, are the adapted policy parameters output by the inner-loop.
Most commonly, $\phi$ is passed directly to the policy as an input vector: $\pi_\theta(a|s,\phi)$.
In task inference, the vector $\phi$ is generally the predicted task estimate: $\pi_\theta(a|s,\hat{c}_\mathcal{M})$, where $\phi=\hat{c}_\mathcal{M}$.
For computing the supervised loss, this task representation must be known during meta-training, and so constitutes a form of \textit{privileged information}.
The representation may, for instance, be a one-hot representation of a task index, if the task distribution is discrete and finite.
Or, it may be some parameters defining the MDP.
For example, if kitchens differ in the location of the stove and refrigerator, the task representation, $c_\mathcal{M}$, could be a vector of all these coordinates.
In this case, $f_\theta$ would predict these coordinates.
The representation may even contain sub-tasks and their hierarchies, or human preferences \citep{ren2022efficient}, when such information is known \citep{sohn2020meta, zhang2022temporal}, or make use of known transition functions to explicitly approximate a Bayesian posterior over the tasks \citep{lee2018bayesian}.
Commonly, the task-inference objective, denoted here as $J_{\mathrm{infer}}$, is given by the maximum likelihood estimate:
\begin{equation*}
J_{\mathrm{infer}}(\theta) = \E_\mathcal{M}[\E_{\mathcal{D} | \pi}[\log p_\theta(c_\mathcal{M})]].
\end{equation*}

When passing the task to the policy, there are a few important representation choices.
First, instead of conditioning the policy on the task representation directly, we can pass a representation with more information about task uncertainty.
This can be accomplished, for instance, by passing to the policy the penultimate layer when predicting $\hat{c}_\mathcal{M}$ \citep{humplik2019meta}.
Then the task parameters, $\phi$, passed to the policy, are trained by inferring $\hat{c}_\mathcal{M}$, but only after a subsequent linear transformation: $\hat{c}_\mathcal{M} = L_\theta(\phi)$.
This hidden layer may contain more information than $\hat{c}_\mathcal{M}$, and $\hat{c}_\mathcal{M}$ can always be computed from it by the policy.
Second, it can be useful to add a stop-gradient to $\phi$ before passing it to the policy \citep{humplik2019meta,zintgraf2021varibad}, to prevent conflicting gradients.
Finally, when training $f_\theta$ with $J_{\mathrm{infer}}$, as a supervised objective, the dataset $\D$ may contain off-policy data without bias, which may given a particular advantage in the off-policy setting, as compared to $J_{\mathrm{meta}}$ given by Equation~\ref{eq:meta-rl-objective} \citep{humplik2019meta}.

\paragraph{Task inference with multi-task training}
\label{para:TI with MT training}
Some research uses the multi-task setting to improve task inference with privileged information \citep{humplik2019meta,kamienny2020learning,raileanu2020fast,liu2021decoupling,peng2021linear}.
The task representation may contain little task-specific information (e.g., if it is one-hot representation) or task-specific information that is irrelevant to the policy (e.g the amount of oxygen in the kitchen).
For example, consider the more concrete task of navigation to a goal on the perimeter of a circle, as discussed in Section \ref{section:introduction}.
In this case, if the task representation is one-hot, it may be useful to instead have access to the $(x_\textrm{goal}, y_\textrm{goal})$ coordinates of the goal.
Additionally, if the task representation contains the location of an additional irrelevant object, $(x_\textrm{goal}, y_\textrm{goal}, x_\textrm{object}, y_\textrm{object})$, then if may be useful to instead have access to a more parsimonious task descriptor, $(x_\textrm{goal}, y_\textrm{goal})$, that contains the goal location alone.
In general, the entire MDP does not need to be uniquely identified.
The agent only needs to identify the variations between MDPs that occur in the task distribution, $p(\mathcal{M})$.
Even more specifically, the agent only needs to identify the subset of those variations that change the optimal policy.

\looseness=-1
To address uninformative and irrelevant task information, representations can be learned by pre-training in the multi-task setting \citep{humplik2019meta,liu2021decoupling}.
Let $g_\theta(c_\mathcal{M})$ be a function that encodes the task representation.
First an informed policy, $\pi^{\mathrm{multi}}_\theta(a|s,g_\theta(c_\mathcal{M}))$, can be trained.
This is the multi-task phase, and enables the learning of $g_\theta$.
Since this representation, $g_\theta(c_\mathcal{M})$, is learned end-to-end, it contains the information relevant for solving the task.
For example, $g_\theta$ could transform $c_\mathcal{M}$ from $(x_\textrm{goal}, y_\textrm{goal}, x_\textrm{object}, y_\textrm{object})$ to $(x_\textrm{goal}, y_\textrm{goal})$.
Often, an information bottleneck \citep{alemi2017deep} is used to ensure it contains only this information  \citep{humplik2019meta,liu2021decoupling}.
After this, the inner-loop can infer the learned representation, $g_\theta(c_\mathcal{M})$, from the meta-trajectory, which we write as $\hat{g}_\theta(\D)$ (see Figure~\ref{fig:task_inference}).
For example, in the circle navigation task, after learning the coordinate representation, $(x_\textrm{goal}, y_\textrm{goal})$, the inferred $\hat{g}_\theta$ could be represented by the inferred coordinates, $(\hat{x}_\textrm{goal}, \hat{y}_\textrm{goal})$.
Alternatively, the informed multi-task RL may be performed concurrently with the meta-RL \citep{kamienny2020learning}.

\begin{figure}[ht!]
    \centering
    \includegraphics[height=4cm,alt={Illustration of normal meta-RL (left), task inference using privileged information (middle), and task inference using multi-task training (right).}]{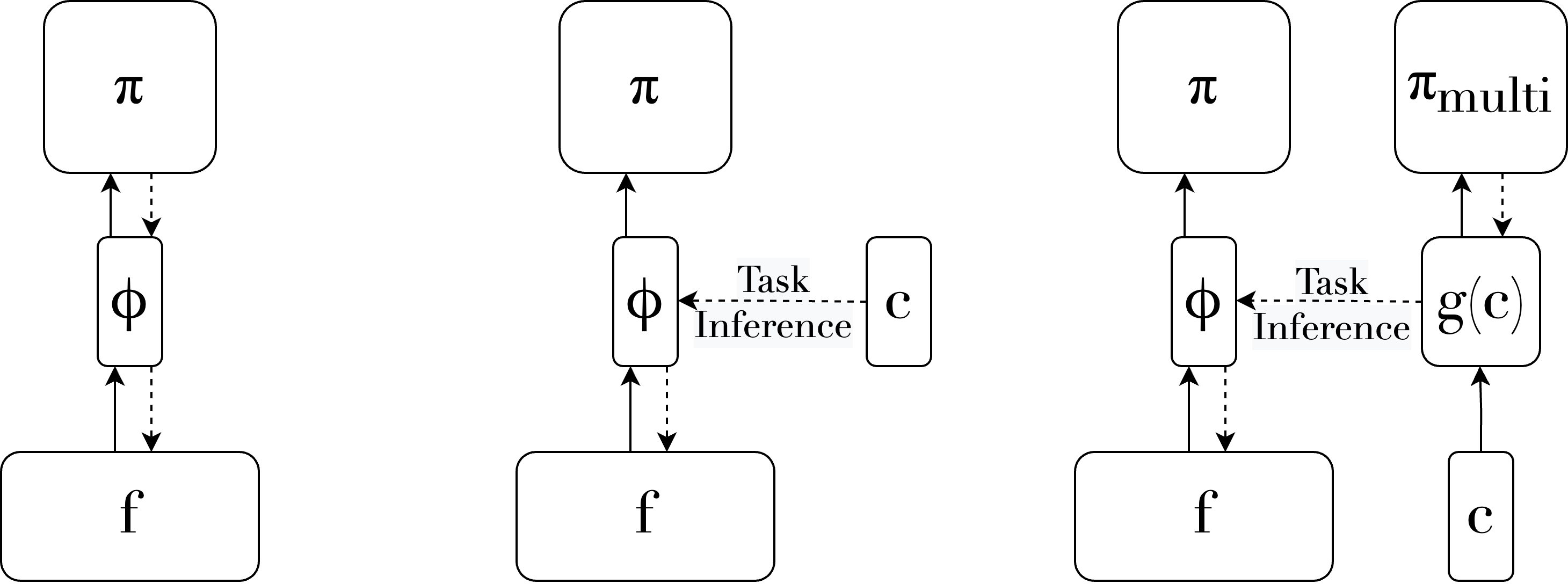}
    \caption{
        Illustration of normal meta-RL (left), task inference using privileged information (middle), task inference using multi-task training (right). Solid arrows represent forward propagation, and dashed arrows represent backpropagation. Task inference with privileged information uses a true task representation, $c$, and then backpropagates that inference to train $\phi$. Task inference with multi-task training learns an encoding of the task, $g(c)$, to maximize the return of an informed policy, $\pi_{\mathrm{multi}}$, then uses inference of this learned encoding to train $\phi$.
    }
    \label{fig:task_inference}
\end{figure}

For example, the informed policy regularization algorithm (IMPORT, \cite{kamienny2020learning}) follows the simultaneous-training paradigm.
In this case, the auxiliary inference objective is given by
\begin{equation*}
J_{\mathrm{infer}}(\theta) = \E_\mathcal{M}[\E_{\mathcal{D} | \pi}[(g_\theta(c_\mathcal{M})-\hat{g}_\theta(\D))^2]].
\end{equation*}
This inference objective forces the encoding of $g$ and $\hat{g}$ to look similar.
While optimizing this objective, the algorithm also optimizes the normal meta-learning objective (see Equation~\ref{eq:meta-rl-objective}) but with the policy conditioned on the inferred task representation: $\pi_\theta(a|s,\hat{g}_\theta(\D))$.
Additionally, IMPORT simultaneously optimizes the meta-RL objective but with the multi-task policy conditioned on the learned task representation: $\pi_\theta^{\mathrm{multi}}(a|s,g_\theta(c_\mathcal{M}))$.

\enlargethispage{\baselineskip}
Some task distributions even allow for significant shared behavior between the informed multi-task agent and the uninformed meta-RL agent.
This sharing is generally possible when little exploration is needed for the meta-RL policy to identify the task.
In this case instead of only inferring the privileged task information, the meta-RL agent may imitate the multi-task agent through distillation \citep{weihs2021bridging,nguyen2022leveraging}, which we cover in Section~\ref{sec:supervision}, or by direct parameter sharing of policy layers \citep{kamienny2020learning,peng2021linear}.
For instance, in the IMPORT algorithm, $\pi_\theta$ and $\pi_\theta^{\mathrm{multi}}$ share parameters by using the same components of the meta-parameters vector, $\theta$.
In this case, the entire policy is shared, such that $\pi = \pi^{\mathrm{multi}}$.
Even when the multi-task and meta-RL policies are computed sequentially, the pre-trained multi-task agent may still be used as an initialization for the meta-RL agent \citep{beck2023recurrent}.

In contrast, when task distributions require taking sufficiently many exploratory actions to identify the task, sharing policies becomes less feasible.
For example, in the circle navigation task, the informed multi-task policy, with access to the goal, never needs to explore, whereas the primary behavior of the meta-RL policy is exploration around the circumference of the circle.
If one were to reuse the multi-task policy conditioned on the inferred goal, $\pi_\theta=\pi_\theta^{\mathrm{multi}}(a|s,\hat{x}_\textrm{goal}, \hat{y}_\textrm{goal})$, as the meta-RL policy, then the inference would not have sufficient data to be accurate.
The initial inferred goal location, for example, may be $(0,0)$, which is not in the training distribution of the multi-task agent.
In this case, the behavior of the multi-task agent would be undefined.
However, sharing some parameters, simultaneously training the shared policy on inferred representation \citep{kamienny2020learning}, or fine-tuning the shared policy as an initialization for the meta-RL policy \citep{beck2023recurrent}, can solve this issue.
For example, if training the shared policy simultaneously with known and inferred goal locations, $(x_\textrm{goal}, y_\textrm{goal})$ and $(\hat{x}_\textrm{goal}, \hat{y}_\textrm{goal})$, the agent can learn to execute exploratory behavior whenever the inferred task is $(0,0)$, or has high uncertainty in the posterior distribution.
Still, collecting data sufficient for inferring the task correctly may be difficult.
Often, intrinsic rewards are additionally needed to even be able to collect the data enabling task inference.
Such exploration in task inference methods is discussed in Section~\ref{sec:explore_basics}.

\enlargethispage{-\baselineskip}
\paragraph{Task inference without privileged information}
Other task inference methods do not rely on privileged information in the form of the known task representation.
If the algorithm is allowed to know whether two trajectories where generated by the same task, then one option is to use this label as a learning signal for the encoder \citep{fu2021towards,mu2022domino,choshen2023contrabar,guo2018neural, luo2022adapt}.
More commonly, the representation can be (a sample from) a latent variable parameterizing a learned reward function or transition function \citep{zhou2018environment,zintgraf2020varibad,zhang2021metacure,zintgraf2021varibad,he2022learning}.
These methods use only information already observable and train $f_\theta(\mathcal{D})$ to represent a task distribution, given trajectories in $\mathcal{D}$.

For example, \citet{zintgraf2020varibad} propose to learn the task parameters, $\phi$, as a latent variable parameterizing the reward and transition function.
In this case, the latent is trained in a self-supervised manner by reconstructing trajectories.
This is accomplished using only observable information by predicting rewards and next states for each transition, given each $\phi_t$:

\small
\vspace{-0.8\baselineskip}
\begin{align*}
     J_{\mathrm{infer}}(\theta) & = \E_\mathcal{M}[\E_{\mathcal{D} | \pi}[\sum_{t=0}^{HN-1} \log p_\theta(\mathcal{D}|\phi_t)]] \\
     & = \E_\mathcal{M}[\E_{\mathcal{D} | \pi}[\sum_{t=0}^{HN-1} \sum_{s,a,r,s' \in \mathcal{D}} [ \log p_\theta(s'|s,a,\phi_t) +  \log p_\theta(r|s,a,\phi_t)]]],
\end{align*}\normalsize

\noindent where the inner summation is taken over consecutive transitions, $(s,a,r,$ $s')$, in $\mathcal{D}$.
An extension of this algorithm has even been proposed with a hierarchical latent variable to accommodate the additional structure in distributions of POMDPs \citep{akuzawa2021estimating}.

\paragraph{Inner-loop representation}
Generally, task inference is implemented by adding an additional loss function, and not by any particular meta-parameterization of $f_\theta$.
While task inference methods do not require a particular meta-parameterization, most implementations use a ``black box,''
such as a recurrent neural network \citep{humplik2019meta,zintgraf2020varibad,kamienny2020learning,liu2021decoupling,zintgraf2021exploration}.
Since many task inference methods infer a latent variable, it is also common for $f_\theta$ to explicitly model this distribution using a variational information bottleneck \citep{humplik2019meta,rakelly2019efficient,zintgraf2020varibad,liu2021decoupling,zintgraf2021exploration}.

In this case, the variational distribution defining the latent variable conditions on $\D$.
We write this as $q_\theta(z|\D)$.
This variational distribution has its own parameters inferred from the dataset, such as a mean, $\mu_\theta(\mathcal{D})$, and a covariance matrix $diag(\sigma_\theta(\mathcal{D}))$.
First, the task representation is reconstructed, e.g., with a linear projection of samples from $q_\theta$: $\hat{c}_\mathcal{M} = L^c_\theta(z \sim q_\theta)$.
Then, the inference is trained by maximizing $J_{\mathrm{infer}}$, also called the reconstruction loss, over samples from the variational distribution:
\begin{equation*}
    J_{\mathrm{infer}}(\theta) = \E_\mathcal{M}[\E_{\mathcal{D} | \pi}[p_\theta(c_\mathcal{M}|\hat{c}_\mathcal{M})]],
\end{equation*}
along with a KL-Divergence to a prior,
\begin{equation*}
    J_{\mathrm{prior}}(\theta) = \E_\mathcal{M}[\E_{\mathcal{D} | \pi}[D(q_\theta(z)||p_\theta(z)]].
\end{equation*}

Once the task inference is trained, the parameters passed to the policy, $\phi$, must be chosen.
One option is to represent $\phi$ as a projection of the full distribution of the latent variable, e.g., its mean and variance, in order to capture uncertainty about the task \citep{zintgraf2020varibad, imagawa2022off}.
This can be written $\pi_\theta(a|s, L^\phi_\theta(\mu,\sigma))$, where $\phi=L^\phi_\theta(\mu,\sigma)$.
Combining this representation of the latent distribution with the $J_{\mathrm{infer}}$ for trajectory reconstruction above, we arrive at a canonical method, variational Bayes-Adaptive Deep RL (VariBAD, \cite{zintgraf2021varibad}).
This method is depicted in Figure~\ref{fig:varibad}.

\begin{figure}[ht!]
    \centering
    \begin{subfigure}[t]{0.40\textwidth}
        \centering
        \caption{TI with IB}
        \includegraphics[trim=60 0 0 0, clip, width=\linewidth,alt={Task inference with an information bottleneck (IB).}]{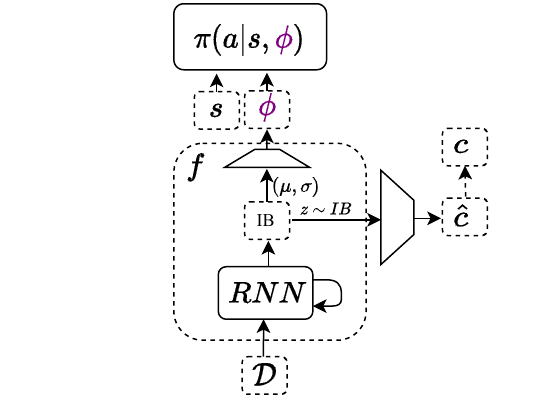}
    \end{subfigure}
    \begin{subfigure}[t]{0.44\textwidth}
        \centering
        \caption{VariBAD \cite{zintgraf2021varibad}}
        \includegraphics[trim=60 0 0 0, clip, width=\linewidth,alt={Task inference with an IB without privileged information.}]{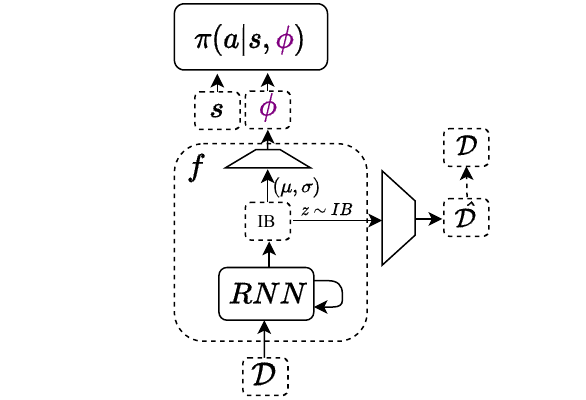}
    \end{subfigure}
    \caption{Task inference with an information bottleneck (IB) (a), task inference with an IB and without privileged information, as in \citet{zintgraf2021varibad} (b). In either case, the parameters of a latent distribution are passed to the policy. When using privileged information, the task representation, $c$, is reconstructed from samples of this distribution; otherwise, the set of trajectories, $\D$, is reconstructed.}
    \label{fig:varibad}
\end{figure}

Alternatively, it is possible for the policy to condition on samples from the information bottleneck (i.e., $\pi_\theta(a|s,z \sim q_\theta)$, where $\phi=z$), instead of conditioning on the parameters of the information bottleneck directly \citep{rakelly2019efficient}.
In this case, the task inference can be trained entirely from the actor and critic loss given by the meta-learning objective, rather than a distinct task inference objective.
\citet{rakelly2019efficient,wen2022improved} use such probabilistic embeddings for actor-critic RL, as introduced by the PEARL method \citep{rakelly2019efficient}.

Several methods also make use of the exchangeability, or permutation invariance, of the transitions implied by the Markov property in task inference \citep{rakelly2019efficient,korshunova2020exchangeable,raileanu2020fast,imagawa2022off,beck2024splagger}.
PEARL specifically models the task distribution conditioned on $\mathcal{D}$ as a product of individual distributions conditioned on each transition in $\mathcal{D}$:
\begin{equation*}
    z \sim q_\theta(z|\mathcal{D}) \propto \prod_{s,a,r,s' \in \mathcal{D}} q_\theta(z|s,a,r,s').
\end{equation*}
While it is not necessary to model this permutation invariance, standard sequence models are sensitive to the ordering of their inputs, which can can be detrimental to learning when the order does not matter \citep{Beck2020AMRL}.
For this reason, it may be useful to embed the structure of permutation invariance into the task inference model as an inductive bias.
Specifically, PEARL uses a product of Gaussians, $q_\theta(z|s,a,r,s')=\mathcal{N}(z;\mu_\theta(s,a,r,s'),diag(\sigma_\theta(s,a,r,s')))$, which has a simple closed form for the joint mean and covariance.
However, other methods forgo the product entirely, and use more general permutation invariant representations, such as neural processes \citep{garnelo2018neural,wang2022learning} and transformers \citep{vaswani2017attention}.

Finally, some work uses hypernetworks in the inner-loop \citep{peng2021linear, beck2022hyper, beck2023recurrent}.
It is possible to use a neural network to map an inferred task to base-network weights and biases: $\pi_{h_\theta(\hat{c}_\mathcal{M})}(a|s)$ \citep{beck2022hyper}.
In this case, the mapping is a hypernetwork, and it can be trained end-to-end with the meta-learning objective.
Alternatively, it is also possible to use the weights and biases of existing experts trained in the multi-task setting as explicit targets for the hypernetwork.
In this case, the using these targets for direct supervision can be seen as (additionally) training with a task-inference objective \citep{peng2021linear,beck2023recurrent}.
Representing the task as the weights and biases of the expert polices has only been investigated as supervision for hypernetworks, but it could, in principle, apply as a task representation for any task-inference method.
Note that even when the hypernetwork has no direct supervision, it is still possible to make use of pre-training in the multi-task setting.
For example, the meta-learned hypernetwork can be pre-trained in the multi-task setting \citep{beck2023recurrent}.

\paragraph{Task inference trade-offs}
Task inference methods present trade-offs in comparison to other methods.
First, we consider the comparison to PPG methods then we consider the comparison to black box methods.
In comparison to PPG methods, task inference methods impose less structure on the inner-loop.
On the one hand, PPG methods generalize well to novel tasks because of the additional structure.
In the case where a novel task cannot be represented using the task representations learned during meta-training, task inference methods fail~\citep{rimon2022meta}, whereas PPG methods generally adapt to the novel task using a policy gradient.
On the other hand, PPG methods are unlikely to recover an algorithm as efficient as task inference.
For distributions where task inference is possible, fitting such a method to the task distribution may enable faster adaptation.
If there are not many tasks in the distribution they are easily inferable from a few consecutive transitions, inferring a latent task using a task inference method can be more sample efficient than learning a new policy with a PPG method.
This is the trade-off between generalization and specialization depicted in Figure~\ref{fig:param_spectrum}.

In comparison to black box methods, task inference methods impose more structure.
On the one hand, task inference methods often add additional supervision through the use of privileged information \citep{humplik2019meta,kamienny2020learning,liu2021decoupling} or the use of self-supervision \citep{zintgraf2020varibad,zintgraf2021exploration}, which may make meta-training more stable and efficient \citep{humplik2019meta,zintgraf2020varibad}.
This is particularly useful since recurrent policies, often used in task inference and black box methods, are difficult to train in RL \citep{Beck2020AMRL}.
On the other hand, there is evidence that black-box training can be stabilized in meta-RL with the use of a hypernetwork architecture and reasonable initialization \citep{beck2023recurrent}, and training the inner-loop to accomplish any objective that is not Equation~\ref{eq:meta-rl-objective} may be suboptimal with respect to that meta-RL objective over the given task distribution.
An empirical comparison between black-box methods and task-inference methods can be seen in Table \ref{tab:ti_tradeoffs}.
Beyond comparisons to PPG methods and black box methods, task inference methods provide some additional advantages.
Task inference methods that model the reward and dynamics can be used to sample additional (imagined) tasks \citep{rimon2022meta}, which can be seen as a type of model-based meta-RL, discussed later in Section~\ref{subsection:model-based}.
And, as we show in Section~\ref{sec:explore_basics}, task inference methods also are useful for exploration.

\begin{table}[t!]
\footnotesize
\centering
\caption{Average return across continuous control environments reproduced from \citet{zintgraf2021varibad,beck2023recurrent}. We compare the black-box method, RL$^2$, and task-inference method, VariBAD, with and without hypernetworks. All methods are evaluated over the training distribution and evaluated after a single episode for adaptation (0-shot with $H=1$). Results are rounded to the nearest hundred. We see that the task-inference method generally outperforms the black-box method without hypernetworks, but with hypernetworks, the black-box methods perform equivalently or better. (Note that the Walker environment is excluded from this table, since the experimental setups on that environment differ between \citet{zintgraf2021varibad,beck2023recurrent}. PPG methods do not compare favorably in this setting; for PPG comparisons, see Figure \ref{fig:consistent} and Table \ref{tab:bb_tradeoffs}.)}
\label{tab:ti_tradeoffs}
\begin{tabular}{lccc}
\hline
    & Ant-Dir & Cheetah-Dir & Cheetah-Vel \\
    \hline  \hline
RL$^2$                 & 1,200 & 1,200 & 0 \\
VariBAD                & \textbf{1,300} & \textbf{2,000} & 0 \\ \hline
RL$^2$ with Hypernetwork    & \textbf{1,400} & \textbf{2,400} & 0 \\
VariBAD with Hypernetwork   & \textbf{1,400} & 2,200 & 0 \\
\hline
\end{tabular}
\end{table}

\section{Exploration and Meta-Exploration} \label{sec:explore_basics}
Exploration is the process by which an agent collects data for learning.
In standard RL, exploration should work for any MDP and may consist of random on-policy exploration, epsilon-greedy exploration, or methods to find novel states.
In meta-RL, this type of exploration still occurs in the outer-loop, which is called \textit{meta-exploration}.
However, there additionally exists exploration in the inner-loop, referred to as just \textit{exploration}, which is where we begin our discussion.
This inner-loop exploration is specific to the distribution of MDPs, $p(\mathcal{M})$.
To enable sample efficient adaptation, the meta-RL agent uses knowledge about the task distribution to explore efficiently.
For instance, instead of taking random actions, the robot chef may open every cabinet to learn about the location of food items and utensils when first entering a new kitchen.
This exploration is targeted and used to provide informative trajectories in $\mathcal{D}$ that enable few-shot adaptation to the MDP within the task distribution.

\begin{table}[t!]
\footnotesize
\centering
\caption{Average return across continuous control environments reproduced from \citet{zintgraf2021varibad}. We compare the black-box method, RL$^2$, and E-MAML \cite{stadie2018some}, a variant of the PPG method, MAML, designed to encourage exploration. Results are rounded to the nearest hundred. (See \citet{zintgraf2021varibad} for details.) RL$^2$ and E-MAML are evaluated over the training distribution and evaluated after a single episode for adaptation (0-shot with $H=1$), which favors black-box methods. Note that even E-MAML after 19 episodes of adaptation (19-shot with $H=20$) is still outperformed by black-box methods on these domains. (For environments where PPG methods are favorable, see the out-of-distribution evaluations in Figure \ref{fig:consistent}.)}
\label{tab:bb_tradeoffs}
\begin{tabular}{lcccc}
\hline
    & Ant-Dir & Cheetah-Dir & Cheetah-Vel & Walker \\
    \hline \hline
RL$^2$ ($H=1$)  &  \textbf{1,200} &   \textbf{1,200}  & \textbf{0}    &  \textbf{300} \\
E-MAML ($H=1$)  &  300   &   0      & -200 &  200 \\
E-MAML ($H=20$) &  300   &   300    & -100 &  200 \\
    \hline
\end{tabular}
\end{table}

Recall that in the few-shot adaptation setting, on each trial, the agent is placed into a new task and is allowed to interact with it for a few episodes (i.e., its few shots $K$), before being evaluated on solving the task in the next few episodes (i.e., over the $H - K$ episodes in Equation~\ref{eq:meta-rl-objective}).
An illustration can be seen in Figure~\ref{fig:exploration}.
Intuitively, the agent must explore to gather information during the first few shots that enables it to best solve the task in later episodes.
More generally, there is an exploration-exploitation trade-off, where the agent must balance taking exploratory actions to learn about the new task (potentially even beyond the initial few shots) with exploiting what it already knows to achieve high rewards.
It is always optimal to explore in the first $K$ episodes, since no reward is given to the agent.
However, the optimal amount of exploration in the remaining $H - K$ shots depends on the size of the evaluation period $H - K$:
When $H - K$ is large, more exploration is optimal, as sacrificing short-term rewards to learn a better policy for higher later returns pays dividends,
while when $H - K$ is small, the agent must exploit more to obtain any reward it can, before time runs out.
In this section, we survey approaches that navigate this trade-off.
Table~\ref{tab:short_horizon_explore} summarizes these categories.
As discussed in Section \ref{section:background}, there is an optimal solution to this exploration problem that maximizes the meta-RL objective.
In the next section, we discuss a framework that formalizes making this trade-off optimally.

\begin{figure}[t!]
    \centering
    \includegraphics[height=4cm,alt={Illustration of free exploration (yellow), penalized exploration (gray), and exploitation (white) as an agent (A) learns to reach a goal (X).}]{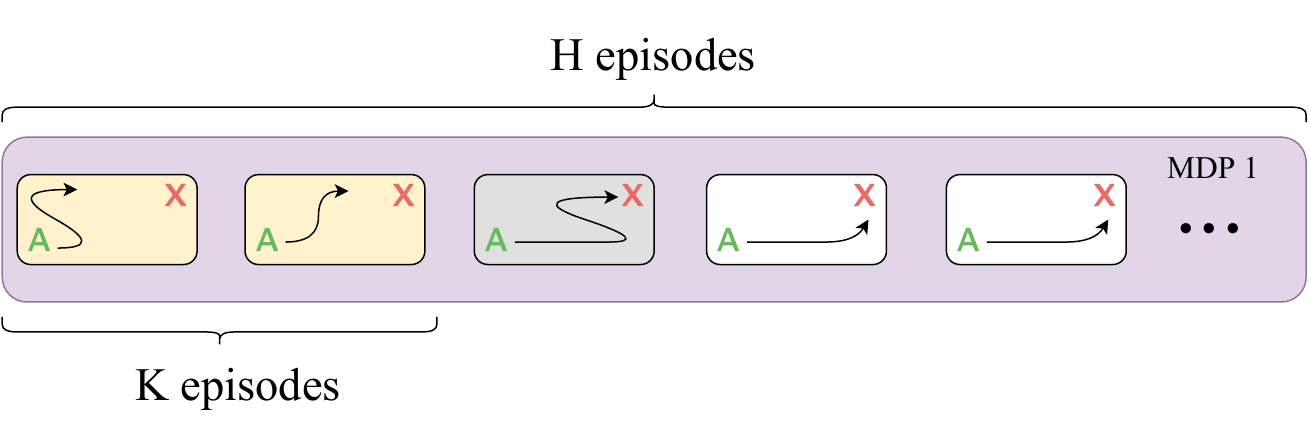}
    \caption{
        Illustration of ``free'' exploration in first $K$ episodes (\textcolor{orange}{yellow}), followed by not free exploration (\textcolor{gray}{gray}), followed by exploitation (white). An agent (A) is trying to identify how to get to a goal location (X). The agent has K shots, or free episodes to explore. However, $K$ episodes may not be enough, so in the $(K+1)$th episode, the agent is still exploring the map to find the goal, and is penalized for this. In the remaining two episodes, the agent has learned to navigate to the goal optimally, once the goal has been found, and no longer needs to explore.
    }
    \label{fig:exploration}
\end{figure}

\paragraph{End-to-end optimization}
Perhaps the simplest approach is to learn to explore and exploit \emph{end-to-end} by directly maximizing the meta-RL objective (Equation~\ref{eq:meta-rl-objective}) as done by black box meta-RL approaches~\citep{duan2016rl,wang2016learning,mishra2018a,stadie2018some,boutilier2020differentiable}. Approaches in this category implicitly learn to explore, as they directly optimize the meta-RL objective whose maximization requires exploration.
More specifically, the returns in the later $K - H$ episodes can only be maximized if the policy appropriately explores in the first $K$ episodes, so maximizing the meta-RL objective can yield optimal exploration in principle.
However, when more complicated exploration strategies are required, learning exploration this way can be extremely sample inefficient.
One issue is that learning to exploit in the later $K - H$ episodes requires already having explored in the first $K$ episodes, but exploration relies on good exploitation to provide reward signal~\citep{liu2021decoupling}.
For example, in the robot chef task, the robot can only learn to cook (i.e., exploit) when it has already found all of the ingredients, but it is only incentivized to find the ingredients (i.e., explore) if doing so results in a cooked meal.
Hence, it is challenging to learn to exploit without already having learned to explore and vice-versa, and consequently, end-to-end methods may struggle to learn tasks requiring sophisticated exploration in a sample-efficient manner, compared to methods with more structure, discussed later in this section.

\begin{table}[t!]
    \footnotesize
    \centering
    \caption{
    Few-shot meta-RL research categorized by exploration method as described in Section~\ref{sec:explore_basics}. End-to-End methods learn to explore implicitly, by directly maximizing the meta-RL objective. Posterior sampling maintains a distribution over possible tasks and acts optimally with respect to samples from this distribution. Task inference guides exploration in order to enable better inference of the task. Meta-exploration concerns exploration in the outer-loop.}
    \label{tab:short_horizon_explore}
    \begin{tabular}{C{0.2\textwidth}C{0.6\textwidth}}
        Sub-topic & Papers \\ \hline \hline
        End-to-End Components & \citet{stadie2018some, garcia2019metamdp, boutilier2020differentiable} \\ \hline
        Posterior Sampling & \citet{gupta2018meta,rakelly2019efficient,kveton2021metathompson,simchowitz2021bayesian} \\ \hline
        Task Inference &  \citet{zhou2018environment,gurumurthy20mame,wang2020learning,liu2021decoupling,fu2021towards,zhang2021metacure} \\ \hline
        Meta-Exploration & \citet{zintgraf2021exploration,grewal2021variance} \\ \hline
    \end{tabular}
\end{table}

One method that learns exploration end-to-end also modifies the reward to encourage exploration.
E-RL$^2$ \citep{stadie2018some} is an end-to-end method that sets all rewards in the first $K$ episodes to zero in the outer-loop.
While ignoring these rewards does introduce sparsity, it may be helpful when myopically maximizing an immediate, dense reward prevents exploration needed for a longer-term reward.
In general, many methods add more structure over end-to-end optimization of the meta-RL objective in order to solve task distributions that demand complicated exploration behavior.

\paragraph{Posterior sampling}
To circumvent the challenge of implicitly learning to explore, \citet{rakelly2019efficient} propose to directly explore via posterior sampling, an extension of Thompson sampling~\citep{thompson1933likelihood} to MDPs.
The inner-loop of this method, PEARL, is described in Algorithm~\ref{alg:pearl}.
When the agent is placed in a new task, the general idea is to maintain a distribution over what the identity of the task is, and then to iteratively refine this distribution by interacting with the task until it roughly becomes a point mass on the true identity.
Posterior sampling achieves this by sampling an estimate of the task identity from the distribution on each episode, and acting as if the estimated task identity were the true task identity for the episode.
Then, the observations from the episode are used to update the distribution either with black box methods~\citep{rakelly2019efficient} or directly via gradient descent~\citep{gupta2018meta}.

\begin{algorithm}[ht!]
\caption{PEARL Inner-Loop}
\label{alg:pearl}
\begin{algorithmic}[1]
\STATE Sample task $\mathcal{M} \sim p(\mathcal{M})$
\STATE Initialize empty meta-trajectory $\mathcal{D}$
\FOR{episode $0, ..., H-1$ sampled from $\mathcal{M}$}
    \STATE Sample $z \sim q_\theta(z|\mathcal{D})$
    \STATE Roll out $\pi_\theta(a|s,z)$ to collect trajectory $\tau$ for this episode
    \STATE Update $\mathcal{D}$ with episode data, $\tau$
\ENDFOR
\end{algorithmic}
\end{algorithm}

Posterior sampling does, however, also have a drawback.
First, since the policy employed is always conditioned on a sampled task, all of the exploration in such a method is executed by a policy that assumes it knows the task it is in.
This means that the same policy is used for both exploration and exploitation.
This can lead to suboptimal exploration
 in terms of Equation~\ref{eq:meta-rl-objective}.
Consider a robot chef that has to find a stove along a curved kitchen counter.
Optimally exploring, the chef walks along the perimeter of the counter until it finds the stove.
If the chef must be reset to its initial position, e.g., to charge its battery at the end of an episode, then the robot resumes exploration where it last left off.
In contrast, a chef using posterior sampling, in each episode, simply walks to a different point along the counter that it has not yet checked, repeating this process until it finds the stove.

This comparison is depicted in Figure~\ref{fig:thompson}.
Other methods for exploration in meta-RL exist that add structure to exploration without this limitation.

\begin{figure}[ht!]
    \centering
    \includegraphics[width=0.9\textwidth,alt={Exploration for a robot chef (green) finding a stove (red) along a circular counter, comparing different methods.}]{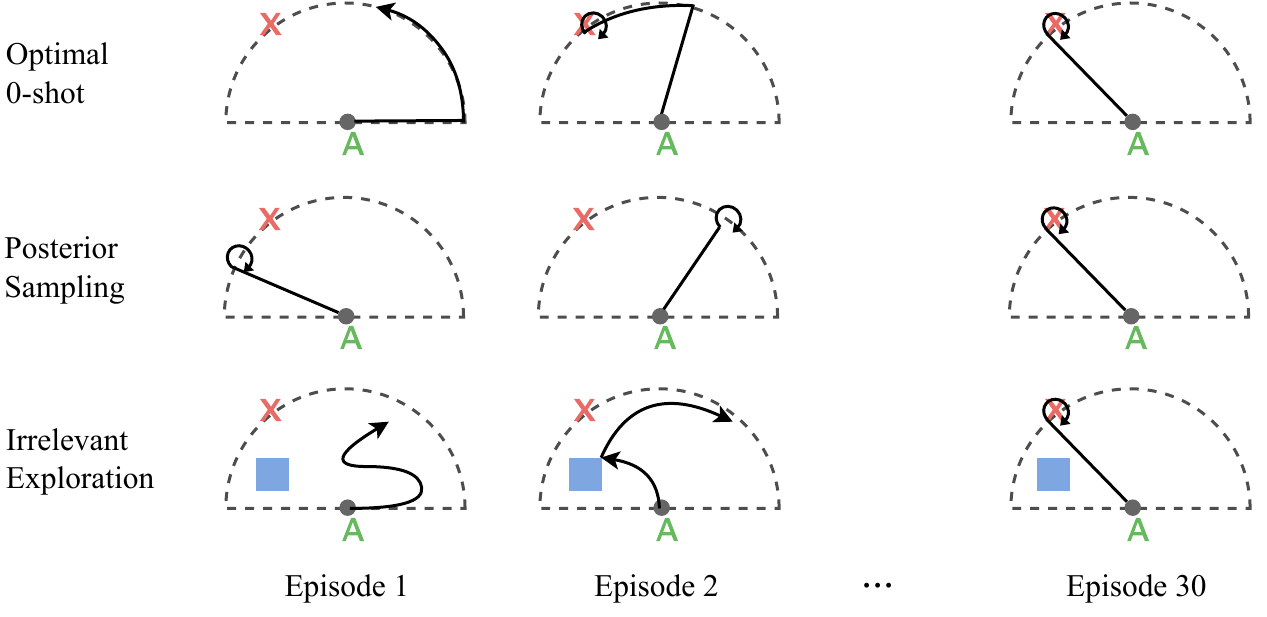}
    \caption{
        Exploration for a robot chef (\textcolor{green}{green}) finding a stove (\textcolor{red}{red}) along a circular counter. The first two rows compare optimal exploration and posterior sampling. The third row shows excessive exploration (\textcolor{blue}{blue}) when irrelevant information is available in the sampled MDP.
    }
    \label{fig:thompson}
\end{figure}

\paragraph{Task inference}
Another way to avoid the challenges of implicitly learning to explore is to directly learn to explore using task inference objectives that encourage exploration \citep{zhou2018environment,gurumurthy20mame,wang2020learning,liu2021decoupling,fu2021towards,zhang2021metacure}.
Some, but not all, task inference methods make use of such an objective to encourage exploration.
Exploration methods that use task inference generally add an intrinsic reward to gather information that removes uncertainty from the task distribution.
In other words, these methods train the policy to explore states that enable predicting the task.
Specifically, these intrinsic rewards generally incentivize improvement in transition predictions (i.e., in adapting the dynamics and reward function) \citep{zhou2018environment} or incentivize information gain over the task distribution \citep{fu2021towards,liu2021decoupling,zhang2021metacure}.
The idea is that recovering the task is sufficient to learn the optimal policy, and hence achieve high returns in the later episodes.
While task inference rewards may incentivize more exploration than necessary, as we discuss in this section in the context in meta-exploration, the rewards can be annealed so that the policy is ultimately optimized end-to-end \citep{zintgraf2021exploration}.

\clearpage
Most of these methods use separate policies for exploration and for exploitation, particularly when exploration episodes are freely available \citep{zhou2018environment,gurumurthy20mame,liu2021decoupling,fu2021towards}.
The intrinsic reward is used to train the exploration policy, while the standard meta-RL objective given by Equation~\ref{eq:meta-rl-objective} is used to train the exploitation policy.
The exploration policy explores for the first $K$ episodes, and then the exploitation policy exploits for the remaining $H - K$ exploitation methods, conditioned on the data collected by the exploration policy.

One method, DREAM~\citep{liu2021decoupling}, first identifies the information that is useful to the exploitation policy, and then trains the exploration policy directly to recover only this information.
This is critical when the number of shots $K$ is too small to exhaustively explore all of the dynamics and reward function, much of which may be irrelevant.
For example, exploring the decorations on the wall may provide information about the task dynamics, but are irrelevant for a robot chef trying to cook a meal.
Learning the task representation in this way can be seen as multi-task training, as described in the Section~\ref{sec:task_inf}, which addresses uninformative (e.g., a one-hot encoding), or irrelevant task information.
This multi-task training is particularly beneficial in the context of exploration since the policy used in the multi-task phase to learn the task representation can also be reused as an exploitation policy.
While, the exploration and exploitation policies could be meta-trained entirely in sequence, in practice they are trained simultaneously for stability \citep{liu2021decoupling, fu2021towards}.
Meta-testing and meta-training for DREAM are described in Algorithms \ref{alg:dream_test} and \ref{alg:dream_train}, respectively.
Still, DREAM has some drawbacks.
For instance, DREAM requires privileged information in the form of a known task representation, and DREAM may be suboptimal when it is not possibly to explore sufficiently in the given $K$ shots (e.g., in a $0$-shot setting).

\begin{algorithm}[ht!]
\caption{DREAM Meta-Testing}
\label{alg:dream_test}
\begin{algorithmic}[1]
\STATE Sample task $\mathcal{M} \sim p(\mathcal{M})$
\STATE Initialize empty meta-trajectory $\mathcal{D}$
\FOR{each exploration episode $k = 0,...,K$ sampled from $\mathcal{M}$}
        \STATE Roll out exploration policy $\pi_\theta(a|s,\mathcal{D})$ to collect trajectory $\tau$
    \STATE Update $\mathcal{D}$ with $\tau$
\ENDFOR
\STATE Infer $ z \sim q_\theta(z|\mathcal{D})$
\FOR{each exploitation episode $h = K,...,H-1$ sampled from $\mathcal{M}$}
    \STATE Roll out exploitation policy $\pi^{\mathrm{multi}}_\theta(a|s,z)$
\ENDFOR
\end{algorithmic}
\end{algorithm}

\begin{algorithm}[ht!]
\caption{DREAM Meta-Training}
\label{alg:dream_train}
\begin{algorithmic}[1]
\WHILE{not done meta-training}
    \STATE Sample M tasks, $\mathcal{M} \sim p(\mathcal{M})$
    \FOR{each task index $i=0,...M$}
        \STATE Initialize meta-parameters, $\theta$, and empty meta-trajectory $\mathcal{D}^i$
        \STATE Sample $z \sim g_\theta(z|c_\mathcal{M})$, a task embedding with IB
        \FOR{each exploitation episode $h = K,...,H-1$}
            \STATE Every other episode, replace true $z$ with inferred $\hat{z} \sim q_\theta$, to make the meta-training task distribution closer to the meta-test distribution for stability
            \STATE Roll out exploitation policy $\pi^{\mathrm{multi}}_\theta(a|s,z)$
            \STATE Update $\pi^{\mathrm{multi}}_\theta$ and $g_\theta$ using task reward and prior
        \ENDFOR
        \FOR{each exploration episode $k = 0,...,K-1$}
            \STATE Roll out exploration policy $\pi_\theta(a|s,\mathcal{D}^i)$ to collect trajectory $\tau$
            \STATE Update $\mathcal{D}^i$ with $\tau$
            \STATE Update $ q_\theta(z|\mathcal{D}^i)$ to infer $z$ using variational inference
            \STATE Compute $r_t^{\mathrm{explore}} \gets -||z - \hat{z_t}||_2^2 + ||z - \hat{z_{t-1}}||_2^2 - c$ for constant $c$, inferred task $\hat{z}$, and all timesteps $t$
            \STATE Update $\pi_\theta$ using exploration reward, $r_t^{\mathrm{explore}}$, i.e. the information gain in $q$ for each transition
        \ENDFOR
    \ENDFOR
\ENDWHILE
\end{algorithmic}
\end{algorithm}

\paragraph{Meta-exploration}
Finally, in meta-RL, there is still the process of acquiring data for the outer-loop learning, just as in standard RL.
This is called \textit{meta-exploration}, since it must explore the space of exploration strategies.
While meta-exploration can be considered exploration in the outer-loop, both loops share data, and exploration methods may affect both loops, so the distinction may be blurry.
Often, sufficient meta-exploration occurs simply as a result of the exploration of the standard RL algorithm in the outer-loop.
However, a common method to specifically address meta-exploration is the addition of an intrinsic reward.
In fact, the addition of any task inference reward, discussed in the previous paragraph, can be considered meta-exploration.
This is particularly apparent when considering that this intrinsic reward can be used to train a policy exclusively for off-policy data collection during meta-training.
However, sometimes adding a task inference reward is not enough.
In this case, intrinsic rewards can be added that function similarly to those in standard RL.
For example, using random network distillation (RND) \citep{burda2018exploration}, a reward may add an incentive for novelty \citep{zintgraf2021exploration}.
In this case, the novelty is measured in the joint space of the state and task representation, instead of just in the state representation, as in standard RL.
For example, HyperX \citep{zintgraf2021exploration} adds the reward,
\begin{align*}
    r^{\mathrm{hyper}}(\phi,s) = ||f(\phi,s)-h(\phi,s)||,
\end{align*}
where $\phi=L_\theta(\mu,\sigma)$ represents a projection of the task distribution, as in VariBAD; $f$ represents the predictor network in RND; and $h$ represents the random prior network in RND.

Additionally, an intrinsic reward may add an incentive for getting data where the task inference has high error and is still not well trained \citep{zintgraf2021exploration}, or add an incentive for getting data where TD-error is high \citep{grewal2021variance}.
For example, HyperX \citep{zintgraf2021exploration} adds the reward:
\begin{align*}
    r^{\mathrm{error}}(\phi,s) & = -\log p_\theta(s'|s,a,\phi) - \log p_\theta(r|s,a,\phi) \\
    & \propto ||s'-\hat s'||_2^2 + ||r'-\hat r'||_2^2
\end{align*}
Many of these rewards incentivize behavior that should not occur at test time, and in any case, the additional reward changes the optimal policy as suggested by Equation~\ref{eq:meta-rl-objective}.
To address this, the reward bonus can be annealed, letting the portion incentivizing meta-exploration go to zero \citep{zintgraf2021exploration},
so that learning is still ultimately optimized end-to-end, or meta-training could occur off-policy.

\section{Bayes-adaptive Optimality}\label{sec:optimal_exploration}

Our discussion so far reveals two key intuitions about exploration.
First, exploration reduces the uncertainty about the dynamics and reward function of the current task.
Crucially, it is not optimal to indiscriminately reduce all uncertainty.
Instead, optimal exploration only reduces uncertainty that increases expected \emph{future} returns, and does not reduce uncertainty over distracting or irrelevant parts of the state space.
Second, there is a tension between exploration and exploitation: gathering information to decrease uncertainty and increase future returns can sacrifice more immediate returns.
In these cases, it can be worthwhile for the agent to fall back on behaviors shared across all tasks, instead of adapting to the task, particularly when time for exploration is limited \citep{lange2020learning}.
Therefore, maximizing the returns across a period of time requires carefully balancing information gathering via exploration and exploiting this information to achieve high returns.
These raise an important question: What is an optimal exploration strategy?
To answer this question, we next introduce the Bayes-adaptive Markov decision process~\citep{duff2002optimal,ghavamzadeh2015bayesian}, a special type of MDP whose solution is a Bayes-optimal policy, which optimally trades off between exploration and exploitation.
Then, we discuss practical methods for learning approximate Bayes-optimal policies, and analyze the behavior of algorithms introduced in the previous section from the perspective of Bayes-optimality.

\paragraph{Bayes-adaptive Markov decision processes.}
To determine the optimal exploration strategy, we need a framework to quantify the expected returns achieved by a policy when placed into an MDP with unknown dynamics and reward function.
From a high level, the \emph{Bayes-adaptive Markov decision process} (BAMDP), a special type of MDP, models exactly this:
At each timestep, the BAMDP quantifies the current uncertainty about an MDP and returns next states and rewards based on what happens in expectation under the uncertainty.
Then, the policy that maximizes returns under the BAMDP maximizes returns when placed into an unknown MDP.
Crucially, the dynamics of the BAMDP satisfy the Markov property by augmenting the states with the current uncertainty.
In other words, the optimal exploration strategy explicitly conditions on the current uncertainty to determine when and what to explore and exploit.

More formally, the BAMDP characterizes the current uncertainty as a distribution over potential transition dynamics and reward functions based on the current observations.
Intuitively, a peaky distribution that places most of its mass on only a few similar dynamics and reward functions encodes low uncertainty, while a flat distribution encodes high uncertainty, as there are many different dynamics and reward functions that the agent could be in.
Specifically, the belief at the $t^\text{th}$ timestep $b_t = p(r, p \mid \tau_{:t})$ is a posterior over dynamics $p$ and reward functions $r$ consistent with the observations $\tau_{:t} = (s_0, a_0, r_0, \ldots, s_t)$ so far, and the initial belief $b_0$ is a prior $p(r, p)$.
Then, the states of the BAMDP are hyperstates $s_t^+ = (s_t, b_t)$ composed of a state $s_t$ and a belief $b_t$, which effectively augments the state with the current uncertainty.
Conditioned on the hyperstate, the policy can satisfy the Markov property, making the hyperstate a sufficient statistic of the history seen by the agent.
The hyperstate is a natural sufficient statistic for Meta-RL, both for its relevance in the BAMDP, and since it corresponds to the belief-state of the meta-RL POMDP described in Section \ref{subsec:pomdp-formalization}.
However, other sufficient statistics for meta-RL exist as well \citep{choshen2023contrabar,subramanian2022approximate}.

\enlargethispage{-\baselineskip}
As previously mentioned, the transition dynamics and reward function of the BAMDP are governed by what happens in expectation under the current uncertainty.
Specifically, the BAMDP reward function is the expected rewards under the current belief:
\begin{align*}
    R^+(s_t, b_t, a_t) = \mathbb{E}_{R \sim b_t}\left[R(s_t, a_t) \right].
\end{align*}
The transition dynamics of the BAMDP are similar.
The probability of ending up at a next state $s_{t + 1}$ is the expected probability of ending up at that state under the belief, and the next belief is updated according to Bayes rule based on the next state and reward from the \emph{underlying MDP} and not the BAMDP:
\begin{align*}
    P^+(s_{t + 1}, b_{t + 1} & \mid s_t, b_t, a_t) \\ &=
    \mathbb{E}_{R, P \sim b_t} \left[P(s_{t + 1} \mid s_t, a_t)  \delta\left(b_{t + 1} = p(R, P \mid \tau_{:t + 1})\right) \right].
\end{align*}
In other words, the BAMDP can be interpreted as interacting with an unknown MDP and maintaining the current uncertainty (i.e., belief).
Taking an action $a_t$ at timestep $t$ yields a next state $s_{t + 1}$ and reward $r_{t + 1}$ from the MDP, which are used to update the belief $b_{t + 1}$.
The next state in the BAMDP is then $s_{t + 1}^+ = (s_{t + 1}, b_{t + 1})$, but the BAMDP reward is the expected reward under the current belief $r_t^+ = R^+(s_t, b_t, a_t) = \mathbb{E}_{R \sim b_t}\left[R(s_t, a_t) \right]$.

The standard objective on a BAMDP is to maximize the expected rewards over some horizon of $H$ timesteps:
\begin{align}\label{eqn:bamdp_objective}
      \mathcal{J}(\pi) = \mathbb{E}_{b_0, \pi}\left[\sum_{t = 0}^{H - 1} R^+(s_t, b_t, a_t) \right].
\end{align}
As H increases, the agent is incentivized to explore more, as there is more time to reap the benefits of finding higher reward solutions.
Notably, this objective exactly corresponds to the standard meta-RL objective (Equation~\ref{eq:meta-rl-objective}), where the number of shots $K$ is set to 0.

\paragraph{Learning an approximate Bayes-optimal policy}
Directly computing Bayes-optimal policies requires planning through hyperstates.
As the hyperstates include beliefs, which are distributions over dynamics and reward functions, this is generally intractable for all but the simplest problems.
However, there are practical methods for learning approximately Bayes-optimal policies~\citep{lee2018bayesian,humplik2019meta,zintgraf2020varibad,arumugam2022planning}.
The main idea is to learn to approximate the belief and simultaneously learn a policy conditioned on the belief to maximize the BAMDP objective (Equation~\ref{eqn:bamdp_objective}).

As a concrete example, variBAD~\citep{zintgraf2020varibad} learns to approximate the belief with variational inference.
Since directly maintaining a distribution over dynamics and reward functions is generally intractable, variBAD  represents the approximate belief with a distribution $b_t = p(m \mid \tau_{:t})$ over latent variables $m$.
This distribution and the latent variables $m$ can be learned by rolling out the policy to obtain trajectories $\tau = (s_0, a_0, r_0, \ldots, s_{H})$ and maximizing the likelihood of the observed dynamics and rewards $p(s_0, r_0, s_1, r_1, \ldots, s_H \mid a_0, \ldots, a_{H - 1})$ via the evidence lower bound.
Simultaneously, a policy $\pi(a_t \mid s_t, b_t = p(m \mid \tau_{:t}))$ is learned to maximize returns via standard RL.

\paragraph{Connections with other exploration methods}
While not all the methods described in Section~\ref{sec:explore_basics} aim to learn Bayes-adaptive optimal policies, the framework of BAMDPs can still offer a helpful perspective on how these methods explore.
We discuss several examples below.

First, black box meta-RL algorithms such as \rlsquared learn a recurrent policy that not only conditions on the current state $s_t$, but on the history of observed states, actions, and rewards $\tau_{:t} = (s_0, a_0, r_0, \ldots, s_t)$, which is typically processed through a recurrent neural network to create some hidden state $h_t$ at each timestep.
Notably, this history is sufficient for computing the belief state $b_t = p(r, p \mid \tau_{:t})$, and hence black box meta-RL algorithms can in principle learn Bayes-adaptive optimal policies by encoding the belief state in the hidden state $h_t$.
Indeed, variBAD can be seen as adding an auxiliary loss to a black box meta-RL algorithm that encourages the hidden state to be predictive of the belief state, though practical implementations of these approaches differ, as variBAD typically does not backpropagate through its hidden state, whereas \rlsquared does.
However, in practice, black box meta-RL algorithms may struggle to efficiently learn Bayes-adaptive optimal policies in domains requiring exploration that is temporally-extended and qualitatively different from the exploitation behavior.
Learning such sophisticated exploration is difficult due to challenges in end-to-end optimization.

While \citet{team2023human} demonstrate that is possible to learn complicated exploration strategies end-to-end, doing so may also require the use of curriculum learning, distillation, and a large number of samples for meta-training \citep{team2023human}.
Moreover, the meta-exploration problem in their task distribution may not have presented a difficult meta-exploration problem as their agent ``does not use the trial conditioning information it observes to adjust its behaviour'' \citep{team2023human}.
End-to-end training on the meta-RL objective alone presents a difficult optimization problem.
\citet{liu2021decoupling} highlight one such optimization challenge for black box meta-RL algorithms, where learning to explore and gather information is challenging without already having learned how to exploit this information, and vice-versa.

Second, many exploration methods discussed in the previous section consider the few-shot setting, where the agent is given a few ``free'' episodes to explore, and the objective is to maximize the returns on the subsequent episodes.
Likewise, the BAMDP objective can be modified to include free exploration episodes by setting initial rewards to zero.
Depending on the amount of free exploration, the optimal policies for the BAMDP can encourage fairly different exploration behaviors.
For example, in the robot chef task, the optimal few-shot exploration may be to first exhaustively look through the drawers and pantry for the best culinary utensils and ingredients in the initial few shots before beginning to cook.
In contrast, optimal zero-shot behavior may attempt to locate the utensils and ingredients while cooking (e.g., as a pot of water boils), as spending the upfront time may be too costly.
This may result in using less suitable utensils or ingredients, though, especially when optimized at lower discount factors.

More generally, methods designed for the few-shot setting attempt to reduce the uncertainty in the belief state in the initial few free episodes, and then subsequently exploit the relative low uncertainty to achieve high returns.
This contrasts behavior in the zero-shot setting, which may involve interleaving exploration and exploitation.
For example, the posterior sampling exploration in PEARL maintains a posterior over the current task, which is equivalent to the belief state.
Then, exploration occurs by sampling from this distribution and pretending the sampled task is the current task, and then updating the posterior based on the observations, which aims to collapse the belief state's uncertainty.
Similarly, by learning an exploration policy that gathers all the task-relevant information during the few free exploration episodes, DREAM also attempts to collapse the belief state to include only dynamics and rewards that share the same optimal exploitation policy.

\section{Supervision} \label{sec:supervision}

In this section, we discuss most of the different types of supervision considered in meta-RL.
In the standard setting discussed so far, the meta-RL agent receives reward supervision in both the inner- and outer-loops of meta-training, as well as meta-testing.
However, this might not always be the case.
Many variations have been considered, from the unsupervised case (i.e., complete lack of rewards during meta-training or testing), to stronger forms of supervision (e.g., access to expert trajectories or other privileged information during meta-training and/or testing).
Each of these presents a different problem setting, with unique methods, visualized in Table~\ref{tab:supervision}.
Below, we discuss these settings, from those with the least supervision to those with the most.

\begin{table}[ht!]
    \centering
    \footnotesize
    \caption{
    Problem settings by supervision available at meta-training and meta-testing time as categorized in Section~\ref{sec:supervision}. Most of the literature in few-shot meta-RL considers the problem setting with rewards provided at meta-train and meta-test time. There are three additional variations on this supervision in meta-RL addressed in this section. Meta-imitation learning is a related but separate problem, briefly covered in this survey.}
    \label{tab:supervision}
    \resizebox{\textwidth}{!}{
    \begin{tabular}{C{0.16\textwidth}C{0.16\textwidth}C{0.16\textwidth}C{0.38\textwidth}}
         & Meta-Train & Meta-Test & Papers \\
        \hline \hline
        Standard Meta-RL & Rewards & Rewards & See, e.g., Table~\ref{tab:short_horizon_methods} \\
        \hline \hline
        Unsupervised Meta-RL  & Unsupervised & Rewards & \citet{gupta2018unsupervised,jabri2019unsupervised,mutti2022unsupervised} \\ \hline
        \vspace{1cm} Meta-RL with Unsupervised Meta-Testing & \vspace{1cm} Rewards (Sparse) & \vspace{1cm} Unsupervised (Sparse) & \vspace{-1cm} \citet{sung2017learning,miconi2018differentiable,yang2019norml,yan2020multimodal,najarro2021metalearning} \\
        & & & \vspace{-.5cm} Sparse: \citet{gupta2018meta,rakelly2019efficient,zhao2021meld,wan2022hindsight}\\ \hline
        Meta-RL via Imitation & Supervised & Rewards & \citet{medonca2019guided,weihs2021bridging,sharaf2021meta,nguyen2022leveraging}\\ \hline
        Mixed Supervision & Rewards and/or Supervision & Rewards and/or Supervision & \citet{zhou2020watch,prat2021peril,dance2021demonstration,rengarajan2022enhanced} \\ \hline \hline
        Meta-Imitation Learning & Supervised & (Un)- supervised & \citet{duan2017oneshot,finn2017oneshot,james2018taskembedded,yu2018oneshot,xu2019learning,dasari2020transformers,bonardi2020learning,singh2020scalable,brown2020language,jang2021bcz,li2021mural,chowdhery2022palm} \\
    \end{tabular}}
\end{table}

\paragraph{Unsupervised meta-RL}
The first problem setting provides the least supervision: no reward information is available at meta-train time, but it is available at meta-test time \citep{gupta2018unsupervised,jabri2019unsupervised,mutti2022unsupervised}.
For example, a robot chef may be meta-trained in a standardized kitchen and then sold to customers, who may each have their own reward functions for the chef.
However, the company training the robot may not know the desires of the customers.
In this case, it is difficult to design the reward functions for the distribution of MDPs needed for meta-training.
It is difficult even to define a distribution under which we might expect the test tasks to have support.
One solution is simply to create rewards that encourage maximally diverse trajectories in the environment.
Then, it is likely that what an end user desires is similar to one of these trajectories and reward functions.
\citet{gupta2018unsupervised,jabri2019unsupervised} attempt to learn a set of diverse reward functions by rewarding behaviors that are distinct from one another.
In general, a set of tasks can be created using an off-the-shelf unsupervised RL method \citep{eysenbach2019diversity, park2024metra}.
After this set of tasks is created, meta-RL can easily be performed as normal.

For example, \citet{gupta2018unsupervised} use the method Diversity is All You Need (DIAYN, \cite{eysenbach2019diversity}), to create this distribution over reward functions.
Specifically, DIAYN first learns a latent variable, $Z$, to parameterize the reward function, along with a multi-task policy, $\pi^{\mathrm{multi}}(a|s,z)$.
DIAYN does so by maximizing the mutual information between the state and the latent variable, ensuring partitioning of the states, and the entropy of the policy:

\noindent
\begin{align*}
    J_\theta & = H(A|S,Z) + I(S,Z), \\
    & = H(A|S,Z) + H(Z) - H(Z|S)
\end{align*}
In practice, this optimization amounts to using soft actor critic \citep{haarnoja2018soft} to maximize $H(A|S,Z)$, while setting $r(s,z) = \log q_\theta(z|s) - \log p(z)$ to maximize the other terms in expectation \citep{eysenbach2019diversity}.
Here, $q_\theta$, is a learned discriminator predicting the probability of the latent variable given the state, and $p(z)$ is a known latent variable distribution.
Note that $q_\theta$ can be learned by maximizing the same objective, $J_\theta$, since optimizing the only term dependent on $q_\theta$, $\log q_\theta(z|s)$, corresponds to a maximum likelihood estimate.
After learning $q_\theta(z|s)$, this discriminator is then reused as a reward function for meta-RL.
Specifically, a separate meta-RL agent (MAML) is trained over this latent distribution using $r(s,z) = \log q_\theta(z|s)$.

Once trained, such meta-RL agents can adapt more quickly than RL from scratch and are competitive, on several navigation and locomotion tasks, with meta-training over a hand-designed training distribution.
Still, these domains are simple enough that diverse trajectories cover the task space, and a gap remains between these domains and more realistic domains such as the robot chef.
Methods from reward-free RL~\citep{touati2021learning}, are highly relevant here, since they can also make use of access to the dynamics alone to learn representations.
However, in that setting, the reward function is given to the agent at test-time, or estimated manually, rather than being inferred by a meta-learned algorithm.

\paragraph{Meta-RL with unsupervised meta-testing}
A second setting assumes rewards are available at meta-train time but none are available at meta-test time.
For example, perhaps the company producing a robot chef is able to install many expensive sensors in several kitchens in a lab for meta-training.
These sensors detect when a counter is scratched, or water damage occurs, or furniture is broken.
All of these collectively are used to define the reward function.
However, it may be prohibitively expensive to install these sensors in the house of each customer.
In this case, rewards are not available at meta-test time.
Rewards are in the outer-loop, but they are never used in the inner-loop, and it is assumed that reward information is not needed to identify the tasks.
In fact, only the dynamics vary across tasks in this setting.

In order to learn
without rewards at meta-test time, many methods remove rewards from the inner-loop entirely.
While the inner-loop cannot condition on reward, it may learn to maximize reward by maximizing its correlates.
While black box methods are applicable in this setting off the shelf, assuming other inputs to the inner-loop correlate with reward, specific methods have been investigated.
For example, in PPG normally the inner-loop  requires sampling returns; however, without rewards, this is not possible at meta-test time.
One solution is use a learned estimate of return, conditional on the data collected so far, to replace these returns in the policy gradient estimate.
Here, the returns can be replaced with a learned advantage function, $A_\theta(s_t,a_t,s_{t+1})$ \citep{yang2019norml} or learned critic $Q_\theta(s_t,a_t,\mathcal{D})$ \citep{sung2017learning}.
For example,  No-Reward Meta Learning (NoRML) modifies the MAML update to include a learned advantage function (in addition to offset parameters, $\phi_{\mathrm{offset}}$, that allow the meta-learned initialization to focus exclusively on exploration):
\begin{equation*}
\phi_1^i = \phi_0 + \phi_{\mathrm{offset}} + \alpha \E_{s_t,a_t,s_{t+1} \in \D_0^i} A_\theta(s_t,a_t,s_{t+1}) \nabla_{\phi_0} \log \pi_{\phi_0}(a_t | s_t).
\end{equation*}
Another approach is to leverage manually designed features in a black-box self-supervised inner-loop, while using a fully supervised outer-loop \citep{yan2020multimodal}.
Finally, yet another approach is to leverage Hebbian learning \citep{hebb1949organization}, a biologically inspired method for unsupervised learning in which weight updates are a function of the associated activations in the previous and next layers.
The update to the weight ($w^k_{i,j}$) from the $i$th activation in layer $k$ ($x^k_i$), to the $j$th activation in layer $k+1$ ($x^{k+1}_j$) generally has the form
\begin{equation*}
w^k_{i,j} := w^k_{i,j} + \alpha(ax^k_ix^{k+1}_j+bx^k_i+cx^{k+1}_j+d),
\end{equation*}
where $\alpha$ is a learning rate and $\alpha$, $a$, $b$, $c$, $d$ are all meta-learned parameters in $\theta$.
Since weight updates only condition on adjacent layer activations, if no rewards are passed as input to the policy, this function is both local and unsupervised.
Hebbian learning can be applied both to feed-forward \citep{najarro2021metalearning} and recurrent neural networks \citep{miconi2018differentiable,miconi2018backpropamine}.
Variants of Hebbian learning for meta-RL, which pass rewards as an input to the policy are investigated by~\citet{chalvidal2022metareinforcement,rohani2022bimrl}.

Alternatively, instead of having no rewards at meta-test time, we may have only sparse rewards.
If dense rewards are available at meta-training, standard meta-RL methods can be applied directly by using the dense rewards in the outer-loop and sparse rewards in the inner-loop \citep{gupta2018meta,rakelly2019efficient,zhao2021meld}.
In the case only sparse rewards are available for both meta-training and meta-testing, one approach is to alter the reward function at meta-training.
A common method for this is a type of \emph{experience relabelling} called \emph{hindsight task relabelling} \citep{packer2021hindsight,wan2022hindsight}.
Assuming tasks differ only in rewards, trajectories can be relabeled with rewards from other tasks and still be consistent with the MDP.
Training can proceed by using a standard off-policy meta-RL algorithm \citep{rakelly2019efficient}.
This is particularly useful if, for instance, the trajectory did not reach a goal state in the original task, but did under the relabeled task.
How to choose such a task is one area of investigation \citep{wan2022hindsight}.
Alternatively, if the dynamics differ, one few-shot method allows for policies to be explicitly transferred between tasks, when helpful, by learning to map actions between tasks such that they produce similar state transitions in each task \citep{guo2022learning}.
Yet another way of addressing sparse rewards at meta-test time is by introducing auxiliary rewards that encourage exploration, as discussed in Section~\ref{sec:explore_basics}.

\paragraph{Meta-RL via imitation}
A third setting assumes access to expert demonstrations at meta-training time, which provide more supervision than standard rewards.
For example, a robot chef may have access to labeled supervision provided by human chefs.
This setting can increase sample efficiency and reduce the burden of online data collection.
This problem setting requires access to expert policies or expert data.
If experts are not known in advance, they can be trained, per task.
Alternatively, in the many-shot setting, policies optimized by existing RL algorithms can provide supervision, in a process known as algorithm distillation \citep{laskin2022context, kirsch2023towards}.
In order to improve over the existing RL algorithm, the learning speed can be increased by subsampling the demonstration data instead of learning on the full traces of expert data \citep{kirsch2023towards}.
One method, Guided Meta-Policy Search (GMPS, \cite{medonca2019guided}), proposes to imitate task experts in the outer-loop.
In this case, the outer-loop can make use of supervised learning, while the inner-loop still learns a reinforcement learning algorithm that conditions on rewards at meta-test time.
They specifically investigate the use of expert labels for the final policy of a MAML-style algorithm.
If an expert is not available, the simultaneous training of task specific experts can also result in stable meta-training \citep{medonca2019guided}.

GMPS illustrates meta-RL via imitation.
GMPS rolls out an initial policy to compute a policy gradient for the inner-loop, and then uses supervised learning in the outer-loop for the adapted policy.
The action labels for the supervised learning are given by expert policies for each task, which are assumed to be known.
Here, the dataset aggregation entails using trajectories from the adapted policy but with expert actions labeled by the expert policies.
This cloning loss can be written $J_{\mathrm{BC}}(\D_{\mathrm{agg}}^i, \pi_{\phi_1^i})$, where $\D_{\mathrm{agg}}^i$ is the aggregated dataset with expert actions for the $i$th task, $\pi_{\phi_1^i}$ is the policy adapted to the $i$th task after a single policy-gradient update, and $J_{\mathrm{BC}}$ is defined as
\begin{equation}
    J_{\mathrm{BC}}(\D, \pi) = \E_{s,a \sim \D} [ \log \pi(a|s) ] \label{eq:bc}.
\end{equation}
This method is described in Algorithm~\ref{alg:gmps}.

\begin{algorithm}[ht!]
\caption{GMPS Meta-Training}
\label{alg:gmps}
\begin{algorithmic}[1]
\STATE Initialize meta-parameters $\phi_0 = \theta$
\WHILE{not done}
    \STATE Sample M tasks, $\mathcal{M} \sim p(\mathcal{M})$
    \STATE Initialize outer-loop datasets, $\D^i_{\mathrm{agg}}$, with states and actions from roll-outs of multi-task experts $\pi^{\mathrm{multi}}(a|s,\mathcal{M}^i)$, for each task $\mathcal{M}^i$
    \FOR{each task index, $i = 0,...,M$}
        \STATE Collect data $\D_0^i$ using the initial policy $\pi_{\phi_0}$
        \STATE Adapt policy parameters using a policy gradient step: $\phi_1^i \gets \phi_0 + \alpha \nabla_{\phi_0} \hat{J}_{RL}(\D_0^i, \pi_{\phi_0})$
    \ENDFOR
    \STATE Update $\phi_0$ using the meta-gradient: $\phi_0 \gets \phi_0 + \beta \nabla_{\phi_0} \frac{1}{M} \sum_{i} J_{\mathrm{BC}}(\D_{\mathrm{agg}}^i, \pi_{\phi_1^i})$
    \STATE Update $\D^i_{\mathrm{agg}}$ by adding states from roll-outs of adapted meta-RL agents, $\pi_{\phi_1^i}$, but actions from $\pi^{\mathrm{multi}}$, for each task index $i$
\ENDWHILE
\end{algorithmic}
\end{algorithm}

Meta-RL via imitation is still relatively unexplored.
This may be due to difficulty in learning the correct supervision for exploratory actions.
Obtaining a meta-RL expert requires knowing how to optimally explore in a meta-RL problem, which is generally hard to compute \citep{zintgraf2020varibad}.
Instead of obtaining such a general expert, these papers often make use of task-specific experts, which can be easily obtained by standard RL on each task.
However, these task-specific experts can only provide supervision for the post-exploration behavior, e.g., for the actions taken by the final policy produced by the sequence of inner-loop adaptations in a MAML-style algorithm.
In this case, credit assignment for the exploration policy may be neglected.
Some papers use the same actions to provide supervision for both uninformed policies that must explore and informed policies that must exploit \citep{zhou2020watch,prat2021peril}; however, in most environments actions are generally not the same for both exploration and exploitation.
To get around this, the agent can adaptively switch between optimizing the meta-RL objective end-to-end and cloning the informed multi-task expert into  \citep{weihs2021bridging}.
Additionally, it may be possible to use the multi-task policy to generate a reward encouraging similar state-action distributions between the multi-task policy and the meta-RL policy, particularly in maximum-entropy RL \citep{nguyen2022leveraging}.

\paragraph{Meta-imitation learning}
A fourth commonly studied problem setting is meta-imitation learning (meta-IL) \citep{duan2017oneshot,finn2017oneshot,james2018taskembedded,yu2018oneshot,paine2018one,ghasemipour2019smile,yu2019metainverse,goo2019one,dasari2020transformers,bonardi2020learning,singh2020scalable,jang2021bcz,li2021meta,gao2022transferring}.
While meta-imitation learning is technically not meta-RL, because the inner-loop is not an RL algorithm, it is a closely related problem setting.
Although an extensive survey of meta-IL methods is out of scope for this work, we briefly cover meta-IL here.
This setting assumes access to a fixed set of demonstrations for each task in the inner-loop.
Most methods additionally train the outer-loop through behavioral cloning on fixed data, in a process also called meta-behavioral cloning (meta-BC).
Alternatively, the outer-loop may also perform inverse RL, which we call meta-IRL
\citep{ghasemipour2019smile,yu2019metainverse,xu2019learning,wu2020squirl,iii2022fewshot}.
Another approach for using meta-learning for IRL is to train success classifiers via few-shot learning \citep{xie2018few,li2021mural}.
Meta-IRL is generally performed online and so often requires a simulated environment, in contrast to meta-behavioral cloning.
For both BC and IRL, the inner- and outer-loops generally also assume access to expert-provided actions, but one line of work considers inner-loops that use only sequences of states visited by an expert, potentially a human, despite being deployed on robotic system \citep{finn2017oneshot,yu2018oneshot,dasari2020transformers,bonardi2020learning,jang2021bcz}.

There are many similarities between meta-RL research and meta-IL research.
In meta-IL,
there exist analogues to black-box methods \citep{duan2017oneshot,james2018taskembedded,dasari2020transformers,bonardi2020learning}, PPG methods \citep{finn2017oneshot,yu2018oneshot}, and task-inference methods \citep{jang2021bcz}, as well as sim-to-real methods \citep{james2018taskembedded,bonardi2020learning}.
For example, to adapt MAML to meta-BC, the inner-loop and outer-loop are each computed with a behavioral cloning loss over offline data, instead of using policy gradients.
An adaptation of MAML for meta-BC, similar to \citet{finn2017oneshot}, is shown in Algorithm~\ref{alg:metaIL-maml}.
Likewise, to adapt RL$^2$ to meta-BC, the RNN summarizes an offline dataset, instead of online data, while the outer-loop uses behavioral cloning.
An adaptation of RL$^2$ for meta-BC is shown in Algorithm~\ref{alg:metaIL-rl2}.

\begin{algorithm}[ht!]
\caption{Meta-Training MAML for Meta-BC}
\label{alg:metaIL-maml}
\begin{algorithmic}[1]
\REQUIRE A dataset of expert demonstrations per task index, $i$, \\ for adaptation $\D^i_{\mathrm{inner}}$, and validation, $\D^i_{\mathrm{outer}}$
\STATE Initialize meta-parameters $\phi_0 = \theta$
\WHILE{not done}
    \STATE Sample  M tasks, $\mathcal{M} \sim p(\mathcal{M})$
    \FOR{each task index $i = 0,..,M$}
        \STATE Adapt policy parameters using a step of behavioral cloning on the offline inner-loop data: $\phi_1^i \gets \phi_0 + \alpha \nabla_{\phi_0} J_{\mathrm{BC}}(\D_{\mathrm{inner}}^i, \pi_{\phi_0})$
    \ENDFOR
    \STATE Update $\phi_0$ using the meta-gradient: $\phi_0 \gets \phi_0 + \beta \nabla_{\phi_0} \frac{1}{M}\sum_{i}J_{\mathrm{BC}}(\D_{\mathrm{outer}}^i, \pi_{\phi_1^i})$
\ENDWHILE
\end{algorithmic}
\end{algorithm}

\begin{algorithm}[ht!]
\caption{Meta-Training RL$^2$ for Meta-BC}
\label{alg:metaIL-rl2}
\begin{algorithmic}[1]
\REQUIRE A dataset of expert demonstrations per task index, $i$, \\ for adaptation $\D^i_{\mathrm{inner}}$ of length T, and validation, $\D^i_{\mathrm{outer}}$
\STATE Initialize meta-parameters $\theta$ (RNN and other neural network parameters)
\WHILE{not done}
    \STATE Sample M tasks, $\mathcal{M} \sim p(\mathcal{M})$
    \FOR{each task index $i=0,...,M$}
        \STATE Initialize RNN hidden state as $\phi_0$
        \STATE Adapt task parameters using RNN on offline inner-loop dataset: $\phi_T^i \gets f_\theta(\D^i_{\mathrm{inner}})$
    \ENDFOR
    \STATE Update $\theta$ using the meta-BC objective: $\theta \gets \theta+ \beta \nabla_{\theta} \frac{1}{M} \sum_{i} J_{\mathrm{BC}}(\D_{\mathrm{outer}}^i, \pi_\theta(\cdot|{\phi^i_T}))$
\ENDWHILE
\end{algorithmic}
\end{algorithm}

Many contemporary papers discuss the meta-IL problem setting under the name of \textit{in-context} learning \citep{brown2020language}.
In-context learning generally refers to learning in the activations of a sequence model, and frequently implies the use of transformers \citep{raparthy2023generalization, lee2024supervised}.
In-context learning is a phenomenon, and meta-learning is an explicit way to achieve it.
Black box methods and task inference methods can be seen as two ways to learn to elicit in-context learning.
In-context learning can be used both in reference to in-context IL and in-context RL \citep{lee2024supervised}.
Large language models in particular, having gained significant traction recently \citep{devlin2019bert,brown2020language,chowdhery2022palm}, perform \textit{in-context} learning \citep{brown2020language}, which can be seen as explicit meta-IL where each prompt is treated as a task.
Moreover, these models are capable of emergent meta-IL, i.e., meta-IL learning that is not explicitly in the training procedure, where novel datasets are passed in text as a prompt to the model along with a query, e.g., in \textit{few-shot prompting} \citep{chowdhery2022palm}, which suggests a similar capacity to be investigated in meta-RL.

\paragraph{Mixed supervision}
Beyond the most commonly studied settings above, \citet{zhou2020watch,prat2021peril,dance2021demonstration,rengarajan2022enhanced} consider a few related, but different settings.
In these settings, there is a phase in which the inner-loop receives a demonstration, followed by a phase in which the inner-loop performs trial-and-error reinforcement learning.
The demonstration generally comes from a fixed dataset collected offline, but may additionally be supplemented by online data from a separate policy trained via meta-IL on the fixed dataset \citep{zhou2020watch}.
The demonstration data itself may \citep{prat2021peril} or may not \citep{dance2021demonstration} provide supervision (actions and rewards), or a combination of the two \citep{zhou2020watch}.
Finally, the agent may use reinforcement learning for supervision in the outer-loop during the trial-and-error phase \citep{dance2021demonstration}, it may use imitation learning \citep{zhou2020watch}, or it may use some combination of the two \citep{prat2021peril}.

\section{Model-based Meta-RL}
\label{subsection:model-based}

So far, most of the algorithms discussed have been \textit{model-free}, in that they do not learn a model of the MDP dynamics and reward.
Alternatively, such models can be learned explicitly, and then be used for defining a policy via planning or be used for training a policy on the data generated by the model.
Such an approach is called \textit{model-based} RL, and can play a helpful role in meta-RL \citep{harrison2018control, saemundsson2018meta,  nagabandi2018learning, nagabandi2018deep, galashov2019meta, mendonca2020meta, kaushik2020fast,  zichuan2020model, lee2020context, perez2020generalized, co-reyes2021accelerating, xian2021hyperdynamics, hiraoka2021meta, seo2022masked, wang2022model, shin2022infusing}.
Model-based meta-RL methods are summarized in Table~\ref{tab:model}.
In this section, we briefly survey model-based meta-RL methods.
We keep the discussion limited for several reasons: most model-based methods have an analogous model-free method already discussed in Section~\ref{sec:fast_adaptation}, most trade-offs between model-based and model-free methods that exist in RL are the same in meta-RL, and most of the meta-RL literature considers model-free RL.

\begin{table}[t!]
    \footnotesize
    \centering
    \caption{
    Categories of model-based meta-RL methods from Section~\ref{subsection:model-based}. Many themes are similar to those in the model-free section, such as the use of parameterized gradient descent and recurrent networks. The top half of the table shows key concepts in the representation of the inner-loop or the model adaptation, while the bottom half shows what the learned adaptive model is used for.}
    \label{tab:model}
    \begin{tabular}{C{0.2\textwidth}C{0.6\textwidth}}
        Sub-topic & Papers \\ \hline \hline
        Parameterized gradient descent & \citet{nagabandi2018learning, mendonca2020meta, kaushik2020fast, co-reyes2021accelerating} \\ \hline
        Recurrent network & \citet{nagabandi2018learning,seo2022masked} \\ \hline
        Variational inference & \citet{saemundsson2018meta,perez2020generalized} \\ \hline \hline
        Model-predictive control & \citet{nagabandi2018learning,lee2020context,shin2022infusing} \\ \hline
        Additional Data & \citet{hiraoka2021meta,rimon2022meta,seo2022masked,shin2022infusing} \\ \hline
        Sub-procedure & \citet{clavera2018model,hiraoka2021meta} \\ \hline
        \hline
    \end{tabular}
\end{table}

Many different types of model-based meta-RL have been investigated.
In order to adapt the parameters of the model, some model-based meta-RL papers use gradient descent, as in MAML \citep{mendonca2020meta, kaushik2020fast, co-reyes2021accelerating, nagabandi2018learning}, and some use black-box RNNs, as in RL2 \citep{nagabandi2018learning}.
Alternatively, a fixed number of past transitions can be encoded \citep{lee2020context}, or variational inference can be used \citep{saemundsson2018meta,perez2020generalized}.
Similar themes arise in model-based meta-RL as in model-free meta-RL, such as the use of hypernetworks \citep{xian2021hyperdynamics} and permutation invariance of transitions \citep{galashov2019meta,wang2022model}, as discussed earlier in Sections \ref{sec:black_box} and \ref{sec:task_inf}.
Once the model of the environment is learned, it can be used to select the best sequence of actions by optimizing the sequence over a finite horizon.
Generally, the first action is taken from this sequence, then the actions are re-planned.
This process can be used to solve meta-RL with an off-the-shelf planner, and is called model-predictive control \citep{nagabandi2018learning,lee2020context}.
Alternatively, some methods use the model simply for additional data when training a policy \citep{hiraoka2021meta,rimon2022meta,seo2022masked}, some uses model-based meta-RL as a sub-procedure in a standard RL algorithm to make learning more sample-efficient \citep{clavera2018model,hiraoka2021meta}, and others use the adapted parameters from the learned model as an input to a policy that is then trained using standard RL \citep{mendonca2020meta,lee2020context,zintgraf2020varibad,zhao2021meld}.
However, in the last case there is significant overlap between model-based meta-RL and task-inference methods because learning to infer the task often means learning a latent-variable dynamics model. We discuss task-inference methods with model-free methods in Section~\ref{sec:task_inf}, due to their otherwise similar assumptions.

A simple method, for example, would learn a black-box environment model, without a separate planner, and run a standard RL algorithm, but use the model for additional data in the policy gradient estimation.
The model itself would be trained using a maximum likelihood estimate of the transition function and reward function over the same trajectories collected by the policy.
For pseudocode of such a model, see Algorithm~\ref{alg:mb_rl2}.
On top of this, it is straightforward to condition the policy on parameters adapted for the transition or reward model, which can also be seen as a type of task inference method.

\begin{algorithm}[ht!]
\caption{Black-Box Model-Based Meta-RL}
\label{alg:mb_rl2}
\begin{algorithmic}[1]
\STATE Initialize an empty dataset for the environment model $\D_{\mathrm{model}}$
\STATE Initialize an empty dataset for the policy $\D_{\mathrm{policy}}$
\STATE Initialize meta-parameters, $\theta_m$, for the model and meta-parameters, $\theta_p$, policy
\WHILE{not done}
    \STATE Sample M tasks, $\mathcal{M} \sim p(\mathcal{M})$
    \FOR{each task index $i=0,...,M$}
        \STATE Replace old data in $\D^i_{\mathrm{policy}}$ with roll-outs from $\pi_{\phi}$ using recurrent policy in real environment
        \STATE Add the same roll-outs to $\D^i_{\mathrm{model}}$
    \ENDFOR
    \STATE Add roll-outs from $\pi_{\phi}$ in simulated environment model to $\D_{\mathrm{policy}}$, optionally starting from real trajectories
    \STATE Update policy $\theta_p$ using the meta-RL objective: \\ $\theta_p \gets \theta_p + \beta \nabla_{\theta_p} \frac{1}{M} \sum_{i} \hat{J}_{RL}(\D_{\mathrm{policy}}^i, \pi_{\theta_p}(\cdot|\phi))$
    \STATE Update model $\theta_m$ using the supervised objective: \\ $\theta_m \gets \theta_m + \beta \nabla_{\theta_m} \frac{1}{M} \sum_{i} \E_{s,a,r,s',\phi \sim \D^i_{\mathrm{model}}} \log p_{\theta_m}(s'|s,a,\phi) + \log p_{\theta_m}(r|s,a,\phi))$
\ENDWHILE
\end{algorithmic}
\end{algorithm}

Overall, there are trade-offs between model-free meta-RL and model-based meta-RL.
On one hand, model-based meta-RL methods can be extremely sample efficient when it is possible to learn an accurate model \citep{nagabandi2018learning}.
Model-based methods may also be learned off-policy, because the model can be trained using supervised learning.
On the other hand, model-based meta-RL may require the implementation of additional components, especially for longer-horizon tasks that require more than an off-the-shelf planner, and can have lower asymptotic performance \citep{clavera2018model,wu2022aggressive}.
Finally, in the meta-learning context, model-based RL may confer two unique advantages.
First, when there are too few tasks in the meta-training distribution, model-based meta-RL can allow for supplemental (imagined) tasks to be sampled \citep{rimon2022meta}.
Second, model-based meta-RL may be easier when an off-the-shelf planner is viable and complicated exploration policies are necessary.
In model-free meta-RL, the meta-learning needs to learn what data to collect and how.
However, in model-based RL, the exploration may be offloaded to a planning algorithm.
It may be easier to allow the planner to deal with complicated exploration in the inner-loop rather than learn this directly.

\section{Theory of Meta-RL}
\label{subsec:theory}

The theory of meta-RL aims to understand and formalize the principles governing the learning of RL algorithms.
As meta-RL is a relatively new area, its theoretical exploration is closely informed by insights from the related areas of meta-SL and multi-task representation learning for RL.
However, meta-RL poses additional challenges due to the inner-loop learning an RL algorithm.
For a theoretical understanding of these challenges, many researchers have used the Bayesian framework.
This Section integrates central insights from the theory of meta-SL and representation learning and overviews specific theoretical developments within meta-RL.
The papers discussed in this section are summarized in Table~\ref{table:meta-rl-theory}.

\begin{table}[ht!]
\footnotesize
\centering
\caption{
Categories of theory papers from Section~\ref{subsec:theory}. Papers relating to theory of meta-SL and representation learning for RL are included in addition to pure meta-RL theory papers.}
\label{table:meta-rl-theory}
\begin{tabular}{C{0.3\linewidth}C{0.6\linewidth}}
Area & Papers \\
\hline \hline
Meta-SL & \citet{baxter2000model, finn2019online, khodak2019adaptive, denevi2019learning, tripuraneni2021provable, collins2022maml} \\
\hline
Multi-task Representation Learning for RL & \citet{jin2020provably, hu2021near, cheng2022provable} \\
\hline
Meta-RL & \citet{simchowitz2021bayesian, tamar2022regularization, rimon2022meta, chen2022understanding, ye2023power} \\
\hline
\end{tabular}
\end{table}

\paragraph{Insights from supervised meta-learning}
Meta-learning, both meta-RL and meta-SL, focus on designing algorithms capable of efficiently learning new tasks from a given task distribution.
A key aspect of this learning process is the development of representations or inductive biases that generalize across tasks.
As an early theoretical contribution, \citet{baxter2000model} laid the groundwork by studying the learning of inductive biases in a PAC (probably approximately correct, \cite{valiant1984theory}) setting, where the learner operates on a family related tasks and the goal is to find a hypothesis space that is appropriate for all of the tasks in the family.
More recently, \citet{tripuraneni2021provable} provide a fundamental result on how features learned across tasks improve the efficiency of learning in a new task.
A related result specialized to MAML \citep{finn2017model} is shown by \citet{collins2022maml}.
MAML is further studied in the online convex optimization framework by \citet{finn2019online,khodak2019adaptive,denevi2019learning}, who assume a notion of task similarity under which all tasks are close to a single fixed point in the parameter space for which guarantees can be provided.
These findings, while not directly addressing meta-RL, offer valuable insights into the representation learning aspect crucial for meta-RL algorithms.

\paragraph{Multi-task representation learning for RL}
Multi-task representation learning for RL is closely related to meta-RL in that it considers learning a shared representation across a set of tasks.
However, in representation learning, learning an exploration strategy or more generally learning a learning algorithm are not considered.
Nevertheless, the settings are close enough that the theoretical results for multi-task representation learning can provide helpful guidance for developing a theoretical understanding of meta-RL.

\citet{yang2020impact,hu2021near} consider representation learning in linear MDPs.
They show regret bounds across the set of parallel tasks that improve as the number of related tasks in the set increases.
Additionally, \citet{cheng2022provable} reveal that multitask representation learning in low-rank MDPs \citep{agarwal2020flambe} can significantly enhance sample efficiency when the total number of tasks surpasses a certain threshold.
Their findings emphasize the efficiency of employing learned representations in downstream tasks, thereby illustrating the tangible benefits of multi-task representation learning in reinforcement learning.

\paragraph{Meta-RL specific theoretical advances}
While research specifically investigating the theory of meta-RL is limited, some important results have been obtained.
In particular, results relating to the generalization bounds of meta-RL algorithms offer a principled motivation on the division of meta-RL methods into few-shot and many-shot methods introduced in Section~\ref{section:background}.

\citet{simchowitz2021bayesian,tamar2022regularization,rimon2022meta} focus on generalization in meta-RL.
\citet{tamar2022regularization} provide PAC bounds on the number of training tasks required for learning an approximately Bayes-optimal policy, i.e., a policy that optimizes Equation $\ref{eq:meta-rl-objective}$.
\citet{rimon2022meta} show that the bound depends exponentially on the degrees of freedom of the task distribution $p(\mathcal{M})$.
These results give a principled explanation of why meta-RL methods work well for narrow task distributions but not for broader ones.
Whereas \citet{tamar2022regularization} and \citet{rimon2022meta} focus on in-distribution generalization, \citet{simchowitz2021bayesian} study the problem of out-of-distribution generalization by considering misspecified priors for Thompson sampling algorithms, which are often considered in meta-RL \citep{rakelly2019efficient}.

Continuing from the Bayesian perspective, \citet{chen2022understanding} bound the regret of the Bayes-optimal policy compared to the optimal policy on \textit{any} MDP instance possible under the prior.
While \citet{chen2022understanding} focus on the worst-case regret, \citet{ye2023power} bound the expected regret.
Furthermore, they propose a pre-training and fine-tuning algorithm based on policy elimination that achieves favorable regret.

Finally, there are ongoing theoretical investigations targeted at MAML-like methods~\citep{finn2017model}.
\citet{fallah2021convergence} derive the sample complexity of a variant of the MAML algorithm.
\citet{liu2022theoretical,tang2022biased} investigate the bias and variance of gradient estimators for MAML and the latter propose a new gradient estimator with a favorable bias-variance trade-off.


\chapter{Many-shot Meta-RL}
\label{sec:long_task_horizon_meta_rl}

In this section, we consider the many-shot setting where we want to, for example, learn a loss function that is applied on new tasks for thousands of updates, rather than just a handful.
In other words, the goal is to learn a general purpose RL algorithm, similar to those currently used in practice.
This setting is discussed separately from the few-shot setting presented in Section~\ref{sec:fast_adaptation} because in practice it considers different problems and methods, even though increasing the trial length does not change the setting in principle.
On one hand, a prototypical few-shot meta-RL problem is goal navigation in MuJoCo environments~\citep{todorov2012mujoco}.
In this scenario, the agent adapts to the different rewards defined by the goal but operates within the same environment.
On the other hand, a typical task in many-shot meta-RL might involve learning an objective function for a policy-gradient method for Atari~\citep{bellemare2013arcade} games.

In the few-shot multi-task setting, an adaptive policy can successfully solve unseen tasks in a small number of episodes by making use of a systematic exploration strategy that exploits its knowledge of the task distribution.
This strategy works well for narrow task distributions, e.g., changing the goal location in a navigation task.
For more complex task distributions with more tenuous relationships between the tasks, the few-shot methods tend to not work as well~\citep{rimon2022meta,mendonca2020meta}.
\citet{rimon2022meta} studies the complexity of meta-RL in terms of the degrees of freedom of the task distribution.
See Section~\ref{subsec:theory} for more details.
Meta-RL methods designed for these more complicated conditions often resemble standard RL algorithms, which are in principle capable of solving any task given enough interactions.
The drawback of standard RL algorithms is that they require many samples to solve tasks.
The meta-RL algorithms building on them aim to improve the sample efficiency by explicitly optimizing for a faster algorithm.
This algorithm template is illustrated in Figure~\ref{fig:long_horizon_meta_rl}.
We label this setting \textit{multi-task many-shot meta-RL} and discuss it further in Section~\ref{subsec:multi-task-many-shot}.

\begin{figure}[ht!]
    \centering
    \includegraphics[width=\textwidth,alt={Many-shot meta-RL where meta-parameters define a policy gradient objective for updating policy parameters in the inner-loop, improving policies over long trials.}]{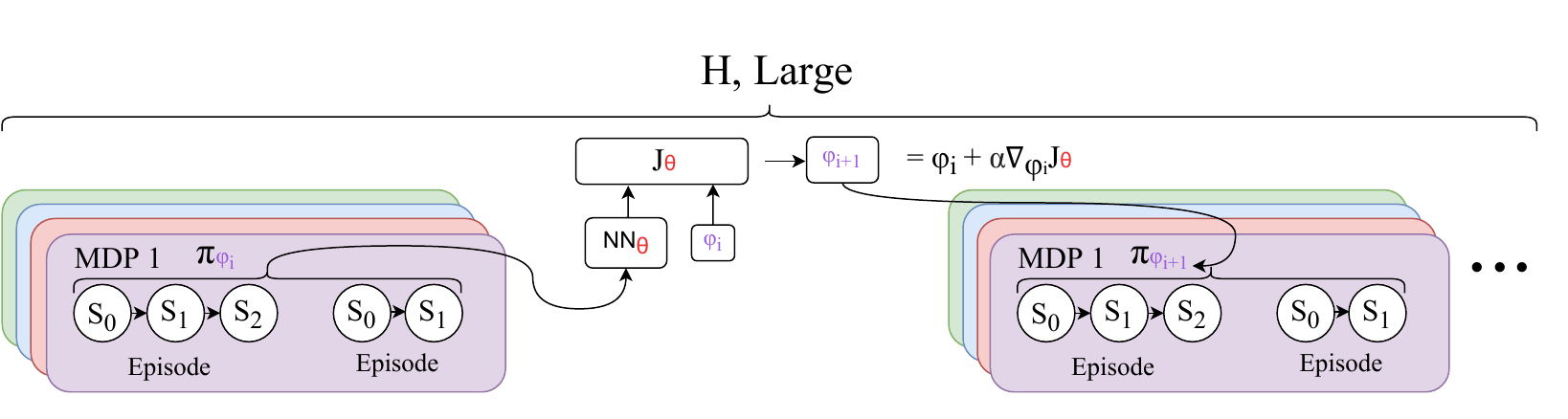}
    \caption{In many-shot meta-RL, the meta-parameters $\theta$ often parametrize a policy gradient objective function $J_{\theta}$ used for updating the parameters of the policy in the inner-loop $\phi$. In this pattern, the inner-loop builds on the inductive bias of a policy gradient algorithm, which helps improve policies over long trials (large $H$. The meta-parameters are updated after a number of inner-loop updates, which may or may not correspond to the trial length depending on specific setting and the algorithm.
    }
    \label{fig:long_horizon_meta_rl}
\end{figure}

Another area where building on standard RL algorithms is helpful is \textit{many-shot single-task meta-RL}.
Instead of seeking greater generalization across complex task distributions, here the aim is to accelerate learning on a single hard task.
Solving hard tasks requires many samples and leaves room for meta-learning to improve sample efficiency during the trial.
We discuss this setting in Section~\ref{subsec:single-task-many-shot}.
While the task distributions in the multi-task and single-task many-shot meta-RL problems are very different, we discuss the methods for both of them together in Section~\ref{subsection:long-horizon-methods} due to the similarity of algorithms used in the settings.

There is less research on the many-shot meta-RL setting compared to its few-shot counterpart.
We believe this disparity is in large part due to the higher computational demands that follow from the very nature of the many-shot meta-RL setting, placing it beyond the reach of many academic groups.
Furthermore, optimization in the many-shot setting is challenging due to high inner-loop computational costs, which limit the kinds of inner-loops that can be efficiently trained.
For example, RL$^2$ could technically be used in the many-shot setting, but optimizing an RNN over millions of timesteps is beyond the current state of the art.
These challenges are open problems in the many-shot meta-RL setting, and are discussed further in Section~\ref{sec:open_problems}.

\section{Multi-task Many-shot Meta-RL}
\label{subsec:multi-task-many-shot}
Many methods in this category are explicitly motivated by the desire to learn RL algorithms that learn quickly on \textit{any} new task.
This requires either a training task distribution that has support for all tasks of interest, or alternatively adaptation to tasks outside the training task distribution.

\enlargethispage{\baselineskip}
The objective for the many-shot multi-task setting is the same as in the few-shot setting and is given by Equation~\ref{eq:meta-rl-objective}.
The differences are the much longer length of the trial $H$ and the broader task distribution, which may include differing action and observation spaces between the tasks.
There is no strict cutoff separating long and short trial lengths but the algorithms targeting each setting are generally easy to distinguish.
On common benchmarks, such as MuJoCo~\citep{todorov2012mujoco}, algorithms targeting the few-shot setting reach maximum performance usually after at most tens of episodes, while algorithms for the many-shot setting are tested on benchmarks like Atari~\citep{bellemare2013arcade} may take tens of thousands or more episodes to converge.
In terms of environment steps, the inner-loop may require tens of thousands to tens of millions of interactions, depending on the environment~\citep{oh2020discovering,meier2022openended}.
For the outer-loop to converge, this can require up to ten billion environment steps in total \citep{oh2020discovering,jackson2023discovering}.
As in the few-shot setting, both meta-training and meta-testing consist of sampling tasks from a task distribution and running the inner-loop for $H$ episodes.

In practice, since a single trial consists of many inner-loop updates, optimizing the meta-parameters over full trials can be challenging.
Running a complete trial in the inner loop requires substantial computation by itself, which can make the meta-parameter updates too expensive to compute.
Furthermore, even if it is feasible to sample full trials for each meta-parameter update, computing such updates may not be possible due to memory requirements or optimization problems like vanishing and exploding gradients.

For example, backpropagating through an agent's entire lifetime necessitates storing all intermediate computations from each inner-loop update, leading to memory usage that grows linearly with the number of updates.
In practice, this means that even processing a single batch can exhaust the memory capacity of standard devices like GPUs.
Additionally, computing meta-gradients involves higher-order derivatives, increasing the computational complexity and often making it quadratic in the number of parameters.

Instead, the meta-learner is commonly updated to maximize a surrogate objective: the performance of the policy after a small number of inner-loop updates starting from the current parameters.
This surrogate objective is a biased estimator of the policy performance at the end of the trial but can still provide gains in practice. The bias from considering a truncated horizon has been analyzed both in supervised learning~\citep{wu2018understanding,metz2019understanding,metz2021gradients} and meta-RL~\citep{vuorio2021no}.
We discuss mitigation strategies for the bias from truncation in Section~\ref{subsection:long-horizon-methods}.

A pseudocode template for a
many-shot meta-RL algorithm is provided in Algorithm~\ref{alg:many-shot-meta-learning}.
In the algorithm, the trials may or may not finish depending on the number of iterations $N$ between each outer-loop update.
Updating the meta-parameters in the middle of a trial corresponds to the truncated objective described above.

\begin{algorithm}[ht!]
\caption{Many-Shot Meta-Learning for Reinforcement Learning}
\label{alg:many-shot-meta-learning}
\begin{algorithmic}[1]
\STATE Initialize meta-parameters $\theta$
\STATE Sample $M$ tasks $\mathcal{M} \sim p(\mathcal{M})$
\STATE Initialize policy parameters $\phi_0^i$ for each task $\mathcal{M}^i$
\STATE Set $j \gets 0$
\WHILE{not done}
\FOR{$N$ iterations}
\FOR{each task index $i = 0, \dots, M$ }
\STATE Collect data $\mathcal{D}^i_j$ using the policy $\pi_{\phi^i_j}$
\STATE Update policy parameters: $\phi^i_{j+1} \gets \phi^i_{j} + \alpha \nabla_{\phi^i_{j}} J_{\theta}(\mathcal{D}^i_j, \pi_{\phi^i_{j}})$
\ENDFOR
\STATE $j \gets j + 1$
\ENDFOR
\STATE Update $\theta$ to maximize the returns of the newest policies: $\theta \gets \theta + \beta \nabla_{\theta} \frac{1}{M} \sum_{i} J(\mathcal{D}^i_j, \pi_{\phi_j^i})$
\STATE For tasks $i$ for which the trial has concluded, re-initialize the policy parameters to $\phi_0$, and sample new tasks $\mathcal{M} \sim p(\mathcal{M})$.
\ENDWHILE
\end{algorithmic}
\end{algorithm}

In the multi-task many-shot setting, the meta-parameters are most commonly parameters of the objective function, which is differentiated with respect to the policy parameters.
This results in a similar algorithm to MAML~\citep{finn2017model} described in Section~\ref{sec:example_algos}.
However, unlike in MAML, where the initialization of the policy is learned, the meta-parameters are in the update function, which enables considering arbitrary policy parameterizations in the inner-loop.
Furthermore, while learning the initialization is a general meta-parameterization, the parametrized update function can be easier to optimize, especially in the many-shot setting.

\section{Single-task Many-shot Meta-RL}
\label{subsec:single-task-many-shot}
\enlargethispage{-\baselineskip}
For some tasks in deep RL, useful training distributions of tasks may not be available.
Moreover, even when they are, the individual tasks may require too many resources to solve on their own making multi-task training across them too hard.
These tasks have motivated researchers to look at whether meta-RL can accelerate learning even when the task distribution consists of a single task, i.e., accelerating learning online during single-task RL training.
Note that this setting is sometimes additionally referred to as \textit{online cross-validation} \citep{sutton1992adapting}.

The goal in the single-task setting is to maximize the final performance of the policy after training has completed.
The final performance is given by the meta-RL objective in Equation~\ref{eq:meta-rl-objective} when we choose a task distribution with only a single task, set the trial length $H$ to a large number, and choose $K$ such that only the last episode counts.
Compared to the few-shot multi-task setting, when meta-learning on a single task, the optimization cannot wait until the inner-loop training has concluded and update only then because there is no further learning on which to use the updated meta-parameters.
Therefore, single-task meta-learning necessarily follows a truncation pattern
similar to the one described in Algorithm~\ref{alg:many-shot-meta-learning}.
In contrast to the many-shot multi-task meta-RL algorithm described in the pseudocode, in the single-task setting there is only one task and as a consequence only one set of policy parameters $\phi_j$.
Those parameters are usually never reset.
Instead, the inner-loop keeps updating a single set of agent parameters throughout the lifetime of the agent.
Therefore, Algorithm~\ref{alg:many-shot-meta-learning} is a fairly good representation of  single-task meta-RL methods when choosing a task distribution with a single task and setting the trial length to the full length of training.

This meta-learning problem is inherently non-stationary, as the data distribution changes with the changing policy.
The ability of the meta-learners to react to the non-stationary training conditions for the policy is often considered a benefit of using meta-learning to accelerate RL, but the non-stationarity itself results in a challenging meta-learning problem.

\looseness=-1
Many of the methods for single-task meta-RL are closely related to methods for online hyperparameter tuning.
Indeed, there is no clear boundary between single-task meta-RL and online hyperparameter tuning, though generally hyperparameters refer to the special case where the meta-parameters $\theta$ are low-dimensional, e.g., corresponding to the learning rate, discount factor, or other hyperparameters of the RL algorithm.
While such hyperparameters are often tuned by meta-RL algorithms \citep{xu2018meta,zahavy2020self},
to limit the scope of our survey, we only consider methods that augment the standard RL algorithm in the inner-loop by introducing meta-learned components that do not have a direct counterpart in the non-meta-learning case.
For a survey of methods for online hyperparameter tuning, see \citet{parker2022automated}.

\section{Many-shot Meta-RL Methods}
\label{subsection:long-horizon-methods}

Algorithms for many-shot meta-RL aim to improve over the plain RL algorithms they build upon by introducing meta-learned components.
The choice of meta-parameterization depends on the problem.
The meta-parameterizations differ in how much structure they can capture from the task distribution, which matters in the multi-task case, and what aspects of the RL problem they address.
The meta-parameterization may tackle problems such as credit assignment, representation learning, etc.
The best choice of meta-parameterization depends on what aspect of the problem is the primary challenge.

Many of the topics discussed below, such as intrinsic rewards and hierarchical RL, are active research areas in RL on their own.
Most of the research on these topics does not consider the bi-level structure present in meta-RL.
We provide a concise description of each topic in general terms but do not to provide details on the bodies of research outside of their intersection with meta-RL.
For learning more about these topics, we provide references to foundational papers and surveys.

In the following, we discuss the different meta-parameterizations considered both in the single-task and multi-task settings.
A summary of the methods discussed in this section is presented in Table~\ref{tab:long_horizon_methods}, where the different methods are categorized by task distribution and meta-parameterization.
The empty categories such as single-task meta-RL for learning hierarchical policies may be promising directions for future work.
We also discuss the different outer-loop algorithms considered for many-shot meta-RL.

\begin{table}[ht!]
    \footnotesize
    \centering
    \caption{
    Many-shot meta-RL methods discussed in Section~\ref{subsection:long-horizon-methods} categorized by the task distribution considered and meta-parameterization.
    }
    \label{tab:long_horizon_methods}
    \begin{tabular}{C{0.2\textwidth}C{0.32\textwidth}C{0.32\textwidth}}
        Meta-parameterization & Multi-task & Single-task  \\ \hline \hline
        Intrinsic rewards & \citet{zheng2020can,alet2020meta,veeriah2021discovery,zou2021learning,meier2022openended} & \citet{zheng2018learning,rajendran2020how} \\ \hline
        Auxiliary tasks & & \citet{veeriah2019discovery,lin2019adaptive,zahavy2020self,flennerhag2021bootstrapped} \\ \hline
        Hyperparameter functions & & \citet{flennerhag2021bootstrapped,almeida2021generalizable,luketina2022metagradients,lu2022discovered} \\ \hline
        Objective functions directly & \citet{houthooft2018evolved,kirsch2019improving,oh2020discovering,bechtle2021meta,jackson2023discovering} & \citet{xu2020meta} \\ \hline
        Optimizers & \citet{chen2017learning,lan2023learning} \\ \hline
        Black-box & \citet{kirsch2021introducing} &  \\ \hline
    \end{tabular}
\end{table}

\paragraph{Learning intrinsic rewards}
The reward function of an MDP defines the task we want the agent to solve.
However, the task-defining rewards may be challenging to learn from because maximizing them may not result in good exploratory behavior, e.g., when rewards are sparse~\citep{singh2009rewards}.
One approach for making the RL problem easier is to introduce a new reward function that can guide the agent in learning how to explore.
These additional rewards are called \emph{intrinsic motivation} or \emph{intrinsic rewards} \citep{aubret2019survey}.
While intrinsic rewards are often designed manually, recently many-shot meta-RL methods have been developed to automate their design~\citep{zheng2018learning,rajendran2020how,veeriah2021discovery}.
One advantage meta-learning has over hand crafting the rewards is that it may be difficult to design reward functions that account for changes happening over time, but it is relatively easy to parameterize the learned rewards as functions of the entire trajectory seen so far~\citep{zheng2020can,alet2020meta}.
Learning an intrinsic reward in the multi-task case can help the agent learn to explore the new environment more quickly~\citep{zheng2020can,alet2020meta,zou2021learning}.
They can also help define skills as part of a hierarchical policy~\citep{veeriah2021discovery} or be used as general reward shaping in the single-task case~\citep{zheng2018learning}.\enlargethispage{\baselineskip}
Beyond the standard settings, \citet{rajendran2020how} learn intrinsic rewards for practicing in extrinsic reward-free episodes in-between evaluation episodes.
Finally, \citet{meier2022openended} present an unsupervised reward learning approach, which learns complex skills in Atari games.

\paragraph{Learning auxiliary tasks}
In some RL problems, learning a good representation of the observations is a significant challenge for which the RL objective alone may provide poor supervision.
One approach for better representation learning is introducing \emph{auxiliary tasks}, defined as unsupervised or self-supervised objectives optimized alongside the RL task~\citep{jaderberg2016reinforcement}.
With auxiliary tasks the inner-loop objective then becomes
\begin{equation*}
    J_{\theta}(\mathcal{D}, \pi_\phi) = J^{\mathrm{RL}}(\mathcal{D}, \pi_\phi) + J^{\mathrm{aux}}_{\theta}(\mathcal{D}, \pi_\phi).
\end{equation*}
When a set of candidate auxiliary tasks is known, the best ones to use can be chosen by meta-learning the weight associated with each task, such that only the tasks that improve the outer-loop objective are assigned high weights~\citep{lin2019adaptive}.
Even when auxiliary tasks are not known in advance, meta-learning can be useful.
\citet{veeriah2019discovery} show that the learned auxiliary tasks can improve sample efficiency over the base algorithm without auxiliary tasks and over handcrafted auxiliary tasks.
The same approach for auxiliary task learning is used by \citet{zahavy2020self} and \citet{flennerhag2021bootstrapped}, who further improve the performance by tuning the hyperparameters of the inner-loop RL algorithm online.
At the time of publication both achieved the state-of-the-art performance of model-free RL on the Atari benchmark~\citep{bellemare2013arcade}.
For comparison of these methods with relevant baselines, see Table~\ref{table:aux-task-atari}.

\begin{table}[ht!]
\centering
\caption{Median human-normalized scores on the Arcade Learning Environment benchmark~\citep{bellemare2013arcade} for many-shot meta-RL methods learning auxiliary tasks.
\label{table:aux-task-atari}
}
\footnotesize
\begin{tabular}{cc}
\textbf{Method} & \textbf{Score (\%)}  \\
\hline
\citet{flennerhag2021bootstrapped} & 611\%  \\
LASER (Not meta-RL)~\citep{schmitt2020off} & 431\%  \\
\citet{zahavy2020self} & 364\%  \\
\citet{xu2018meta} & 287\%  \\
IMPALA (Not meta-RL)~\citep{espeholt2018impala} & 192\%  \\
\hline
\end{tabular}
\end{table}

\paragraph{Learning functions that output hyperparameters}
Methods that learn functions that output hyperparameters of an inner-loop algorithm straddle the gap between hyperparameter optimization and meta-learning. \citet{flennerhag2021bootstrapped,almeida2021generalizable,luketina2022metagradients} learn functions that take as inputs summary statistics of the inner-loop performance such as rewards and TD-error, and output the values of hyperparameters such as the $\lambda$-coefficient used in estimating returns.
In temporal-difference learning, TD($\lambda$) is an algorithm that balances bias and variance by using the parameter $\lambda$ to weight the contribution of bootstrapped multi-step returns, which introduce bias, and the Monte Carlo return estimates, which have high variance.
Furthermore, \citet{lu2022discovered} show that parameterizing the policy update size coefficient featured in algorithms such as proximal policy optimization (PPO)~\citep{schulman2017proximal} by a meta-learned function can be beneficial.
The parameters of these functions are themselves optimized by meta-gradients.

\paragraph{Modifying the RL objective directly}
Learning intrinsic rewards and auxiliary tasks shows that adding meta-learned terms to the RL objective can accelerate RL.
These successes raise the question whether, instead of adding terms to the objectives, modifying the RL objectives directly via meta-RL can improve performance.
To answer this question, \citet{houthooft2018evolved,oh2020discovering,xu2020meta} propose algorithms that replace the return or advantage estimator in a policy gradient algorithm with a learned function of the episode
\begin{equation*}
    \nabla_\phi J_{\theta}(\D, \pi_\phi) \propto \sum_{a_t, s_t \in \D} \nabla_\phi \log \pi_{\phi}(a_t|s_t) f_{\theta}(\D),
\end{equation*}
where $f_{\theta}(\D)$ is some meta-learned function of the trajectory.
An alternative to replacing the advantage estimator is proposed by \citet{kirsch2019improving} and \citet{bechtle2021meta}, who consider a deep deterministic policy gradient (DDPG)-style~\citep{lillicrap2016continuous} objective function, where the critic is learned via meta-RL instead of temporal difference (TD) learning.
\begin{equation*}
    \nabla_\phi J_{\theta}(\D, \pi_\phi) = \sum_{a_t, s_t \in \D} \nabla_\phi Q_{\theta}(s_t, \pi_{\phi}(a_t|s_t)),
\end{equation*}
where $Q_{\theta}$ is the meta-learned critic.
Similar inner-loop is proposed for meta-IL by~\citet{yu2018oneshot}.
These learned RL objectives produce promising results in both multi-task and single-task meta-RL.
In the multi-task setting, \citet{oh2020discovering} demonstrate that an objective function learned on simple tasks such as gridworld can generalize to much more complicated tasks such as Atari~\citep{bellemare2013arcade}.
\citet{jackson2023discovering} improve the generalization of the approach further by replacing the hand-designed training environments with an automated environment design component.
Whereas in the single-task setting, \citet{xu2020meta} show that the learned objective function can eventually outperform the standard RL algorithm (IMPALA, \cite{espeholt2018impala}) it builds upon.

\paragraph{Learning Optimizers}
In most learning systems, the optimizer producing the parameter updates is manually designed.
However, it is also possible to meta-learn an optimizer.
Typically, the inner-loop of meta-learned optimizers conditions on losses and gradients, and outputs parameter updates.
There has been some success in both many-shot and few-shot supervised learning to meta-learn the optimizer \citep{li2016learning, andrychowicz2016learning, ravi2017optimization}.
Some of these meta-learned optimizers use RL for meta-training \citep{li2016learning}, and some deploy on MDPs \citep{chen2017learning}.
Recently, a many-shot method has been proposed to meta-train and meta-test on MDPs, making it the first proper meta-RL method to learn an optimizer \citep{lan2023learning}.
Since then, \citet{goldie2024can} have improved on this technique by additionally conditioning their optimizer on RL-specific heuristics.

\paragraph{Black-box meta-learning}
In few-shot meta-RL, black-box methods that use RNNs or other neural networks instead of stochastic gradient descent (SGD) tend to learn faster than the SGD-based alternatives.
\citet{kirsch2021introducing} argue that many black-box meta-RL approaches, e.g., \citet{duan2016rl,wang2016learning}, cannot generalize well to unseen environments because they can easily overfit to the training environments.
To combat overfitting, they introduce a specialized RNN architecture, which reuses the same RNN cell multiple times, making the RNN weights agnostic to the input and output dimensions and permutations.
The proposed method requires longer trials to learn a policy for a new environment, making it a many-shot meta-RL method, but in return it can generalize to completely unseen environments.

\paragraph{Outer-loop algorithms}

Regardless of the inner-loop parameterization chosen, by definition, algorithms for many-shot meta-RL have to meta-learn over long task-horizons.
Directly optimizing over these long task horizons is challenging because it can result in vanishing or exploding gradients and has infeasible memory requirements~\citep{sutskever2013training,metz2021gradients}.
Instead, as described above, most many-shot meta-RL algorithms adopt a surrogate objective, which considers only one or a few update steps in the inner-loop~\citep{zheng2018learning,veeriah2019discovery,kirsch2019improving,rajendran2020how,zheng2020can,zahavy2020self,oh2020discovering,veeriah2021discovery,bechtle2021meta}.
These algorithms use either A2C~\citep{mnih2016asynchronous}-style \citep{zheng2020can,oh2020discovering,veeriah2021discovery,bechtle2021meta} or DDPG~\citep{lillicrap2016continuous}-style \citep{kirsch2019improving} actor-critic objectives in the outer-loop.
\citet{flennerhag2021bootstrapped} present a different kind of surrogate objective, which bootstraps target parameters for the inner-loop by computing several updates ahead and then optimizing its earlier parameters to minimize distance to that later target using a chosen metric.
This allows optimizing over more inner-loop updates, and with the right choice of metric, it can be used for optimizing the behavior policy in the inner-loop, which is difficult using a standard actor-critic objective.
This surrogate objective has also been used to meta-learning how to prioritize the task distribution for policy improvement in model-based methods \citep{burega2022learning}.

Alternatively \emph{evolution strategies} (ES)~\citep{rechenberg1971evolutionsstrategie,wierstra2014natural,salimans2017evolution}, which are black-box optimization algorithms, are used by \citet{houthooft2018evolved,kirsch2021introducing,lu2022discovered}.
ES works by sampling a set of parameters from a distribution, evaluating the set by running the inner-loop, and updating the parameters of the distribution.
This can be seen as applying REINFORCE~\citep{williams1992simple} on the parameter distribution.
The general approach of using ES in the outer-loop is illustrated in Figure~\ref{fig:es_outer_loop}.
ES suffers less from the vanishing and exploding gradients problem and has more favorable memory requirements at the cost of high variance and sample complexity compared to SGD-based methods~\citep{metz2021gradients}.
Finally, genetic algorithms~\citep{schmidhuber1987evolutionary} and random search are used by \citet{alet2020meta,co-reyes2021evolving,garau2022multi}, who consider discrete parameterizations of the inner-loop objective.

\begin{figure}[t!]
    \centering
    \includegraphics[width=0.45\textwidth,alt={Evolution Strategies (ES) estimates meta-parameter updates by sampling and running the inner-loop independently for each set.}]{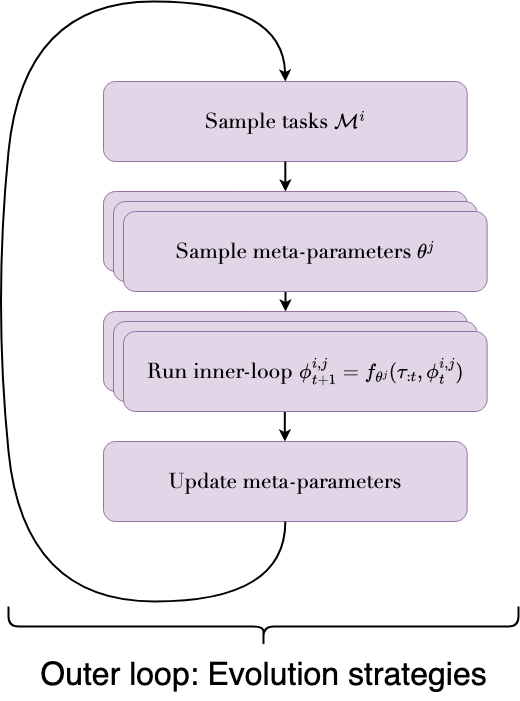}
    \caption{To estimate an update to the meta-parameters, ES samples a set of them and computes the inner-loop for each of them independently. This requires more evaluations of the inner-loop but can work better when the task-horizon is long.
    }
    \label{fig:es_outer_loop}
\end{figure}


\chapter{Applications}
\label{sec:application}
\enlargethispage{\baselineskip}
In many application domains, fast adaptation to unseen situations during deployment is a critical consideration.
By meta-learning on a set of related task, meta-RL provides a promising solution in these domains, such as traffic signal control \citep{zang2020metalight}, building energy control \citep{luna2020information,grewal2021variance}, and automatic code grading in education \citep{liu2022giving}.
Meta-RL has also been used as a subroutine to address non-stationarity in the sub-field of continual RL
\citep{al2018continuous,nagabandi2018deep,riemer2018learning,berseth2021comps,liotet2022lifelong,woo2022structure}.
In continual RL, the task, or MDP, can change throughout the lifetime of an agent.
Meta-RL can help by learning how to address this non-stationarity in the task, given a distribution over sequences of tasks, simply by allowing the MDP to change to closely related MDPs, during a single trial.
Moreover, curriculum learning and unsupervised environment design (UED) have been used to provide the distribution of tasks required for a meta-RL agent \citep{mehta2020curriculum} or an otherwise adaptive agent \citep{dennis2020emergent}.
In this section, we consider where meta-RL has been most widely applied, as well as intersections with other sub-fields where meta-RL has been used successfully to solve problems.
Specifically, we discuss robotics and multi-agent RL.

\begin{figure}[ht!]
    \centering
    \includegraphics[width=\textwidth,alt={Example meta-RL application domains, including building energy control, traffic signal control, multi-agent RL, robot manipulation, robot locomotion, and automatic code grading in education.}]{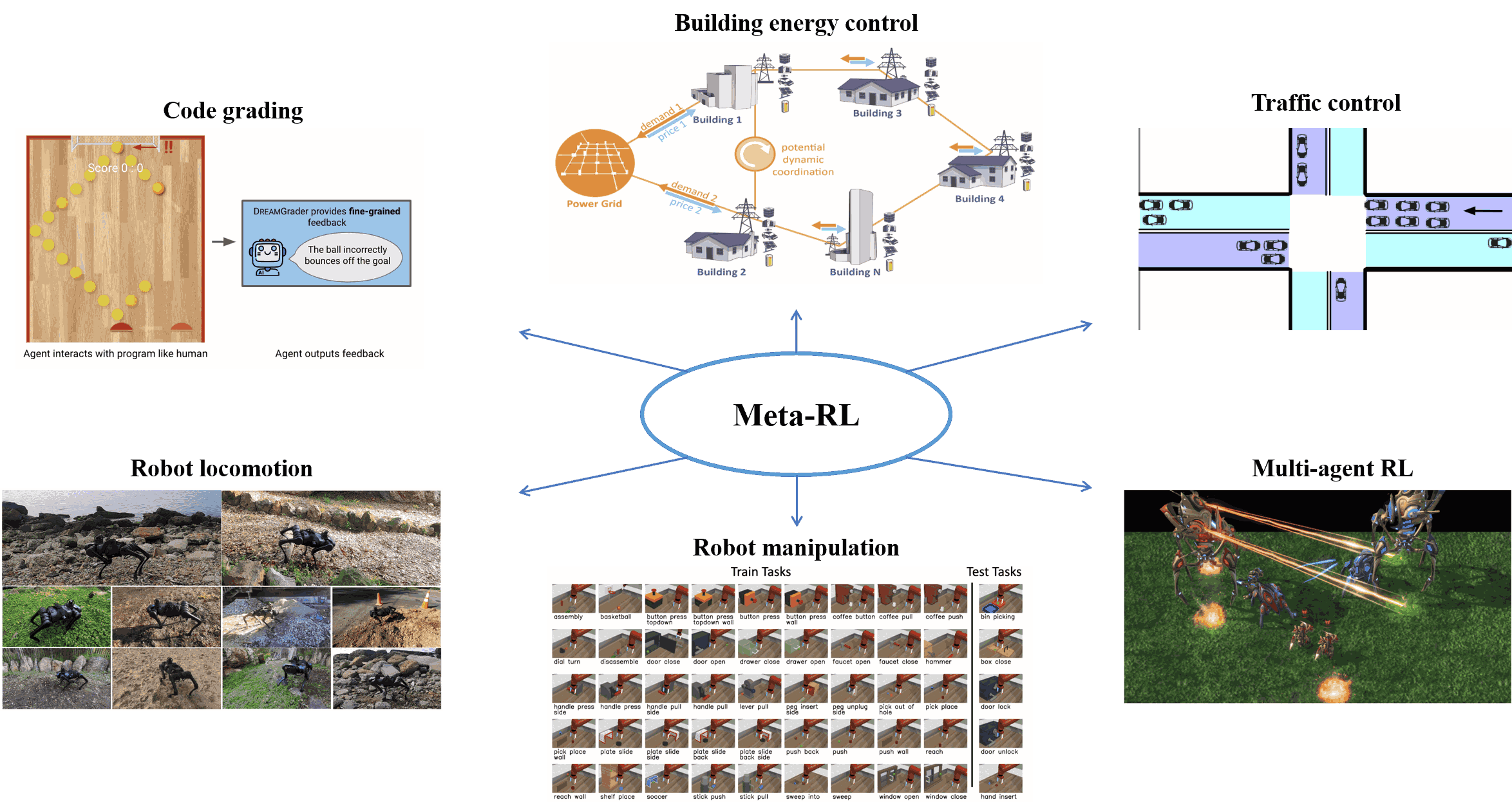}
    \caption{Example meta-RL application domains, including building energy control \citep{luna2020information,grewal2021variance}, traffic signal control \citep{zang2020metalight}, multi-agent RL \citep{al2018continuous,grover18learning,foerster2018learning,papoudakis2020variational,charakorn2021learning,zintgraf2021deep,kim2021policy,feng2021neural,gupta2021dynamic,zhang2021metagrad,yang2022adaptive,gerstgrasser2022metarl,he2022learning,lu2022model}, robot manipulation \citep{yu2017preparing,akkaya2019solving,arndt2020meta,schoettler2020meta,ghadirzadeh2021bayesian,zhao2022offline}, robot locomotion \citep{yu2017preparing,schwartzwald2020sim,yu2020learning,song2020rapidly,kumar2021error,kumar2021rma}, and automatic code grading in education \citep{liu2022giving}. Image credit to \citet{vazquez2019reinforcement,chen2020toward,vazquez2020citylearn,yu2020meta,kumar2021rma,ellis2022smacv2,liu2022giving}. }
    \label{fig:application}
\end{figure}

\section{Robotics}
\label{sec:robot}
\enlargethispage{\baselineskip}
One important application domain of meta-RL is robotics, as a robot needs to quickly adapt to different tasks and environmental conditions
 during deployment, such as manipulating objects with different shapes, carrying loads with different weights, etc.
Training a robot from scratch for each possible deployment task may be unrealistic, as RL training typically requires millions of steps, and collecting such a large amount of data on physical robots is time-consuming, and even dangerous when the robots make mistakes during learning.
Meta-RL provides a promising solution to this challenge by meta-learning inductive biases from a set of related tasks to enable fast adaptation in a new task.

However, while meta-RL enables efficient adaptation during deployment, it comes at the expense of sample-inefficient meta-training on multiple tasks, which is bottlenecked by the cost of collecting an even larger amount of real-world data for meta-training.
Nonetheless, some methods meta-train in the real world \citep{nagabandi2019learning,belkhale2021model,zhao2021meld,zhao2022offline,walke2022dont}, and even learn to manually reset their tasks while doing so \citep{walke2022dont}.
However, most methods
train a meta-RL agent in simulation instead of on physical robots, where different tasks can be easily created by changing the simulator parameters, and then deploy the meta-trained robot in the real world.
This process, known as \emph{sim-to-real transfer} \citep{zhao2020sim}, significantly reduces the meta-training cost and is adopted by most methods that apply meta-RL to robotics \citep{yu2017preparing,akkaya2019solving,arndt2020meta,cong2020self,kaushik2020fast,schoettler2020meta,schwartzwald2020sim,song2020rapidly,yu2020learning,ghadirzadeh2021bayesian,kumar2021error,kumar2021rma,he2022learning}.
We give a taxonomy of these robotic meta-RL papers
in Table~\ref{tab:robotic-paper-taxonomy} and introduce them in more detail below.

\begin{table}[ht!]
    \scriptsize
    \centering
    \caption{Taxonomy of the meta-RL papers that apply meta-RL to robotics from Section~\ref{sec:robot}. Methods for meta-RL can be directly applied to robotics and robotics span the range of meta-RL methods. Task-inference and Model-based methods are common in this section for their sample efficiency. Real-world meta-training is relatively less common.}
    \label{tab:robotic-paper-taxonomy}
    \resizebox{\textwidth}{!}{
    \begin{tabular}{
        >{\centering\arraybackslash}m{0.15\textwidth}
        >{\raggedright\arraybackslash}m{0.14\textwidth}
        >{\centering\arraybackslash}m{0.17\textwidth}
        >{\centering\arraybackslash}m{0.17\textwidth}
        >{\centering\arraybackslash}m{0.17\textwidth}
    }
        & & \textbf{Black-box} & \textbf{Task-inference} & \textbf{PPG} \\
        \hline \hline
        \multirow{4}{*}{\textbf{Model-free}}
        & \textbf{Sim-to-real} & \citet{akkaya2019solving,schwartzwald2020sim} & \citet{yu2017preparing,schoettler2020meta,kumar2021error,kumar2021rma,he2022learning} & \citet{gao2019fast,arndt2020meta,song2020rapidly,yu2020learning,ghadirzadeh2021bayesian} \\
        & \textbf{Real-world meta-training} & - & \citet{zhao2022offline,zhao2021meld,walke2022dont} & - \\
        \hline
        \multirow{4}{*}{\textbf{Model-based}}
        & \textbf{Sim-to-real} & \citet{cong2020self} & - & \citet{kaushik2020fast,anne2021meta} \\
        & \textbf{Real-world meta-training} & \citet{nagabandi2019learning,belkhale2021model} & - & - \\
        \hline
    \end{tabular}}
\end{table}

Model-free meta-RL methods directly adapt the control policy to handle unseen situations during deployment, such as locomotion of legged robots with different hardware (mass, motor voltages, etc.) and environmental conditions (floor friction, terrain, etc.) \citep{yu2017preparing,schwartzwald2020sim,yu2020learning,song2020rapidly,kumar2021error,kumar2021rma}, and manipulation with different robot arms or variable objects \citep{yu2017preparing,akkaya2019solving,arndt2020meta,schoettler2020meta,ghadirzadeh2021bayesian,zhao2022offline}.
On one hand, black-box methods \citep{akkaya2019solving,schwartzwald2020sim} (see Section~\ref{sec:black_box}) and task-inference methods \citep{yu2017preparing,schoettler2020meta,yu2020learning,kumar2021error,kumar2021rma,he2022learning,zhao2022offline} (see Section~\ref{sec:task_inf}), when used in robotics, generally condition the policy on a context vector inferred from historical data to represent the current task.
They mainly differ in what kind of loss function is used to train the task inference.
On the other hand, model-free PPG methods (see Section~\ref{sec:ppg}) in robotics \citep{gao2019fast,arndt2020meta,song2020rapidly,ghadirzadeh2021bayesian} mainly build upon MAML \citep{finn2017model}.
However, they usually require more rollouts for adaptation compared to model-free black-box methods,
which is consistent with our discussion on the efficiency of different methods in Section~\ref{sec:fast_adaptation}.
Moreover, instead of directly using MAML, these methods introduce further modifications to enable sample-efficient training \citep{arndt2020meta,ghadirzadeh2021bayesian} and handle the high noise of the real world \citep{song2020rapidly}.

Another line of works adopts model-based meta-RL (Section~\ref{subsection:model-based}).
Model-based methods may be more suitable for robotic tasks for the following reasons:
(1) Model-based RL can be more efficient than model-free methods, a key consideration for robot deployment \citep{polydoros2017survey}.
(2) Adapting the dynamics model may be much easier than adapting the control policy in some cases, such as in tasks where task difference is defined by various dynamics parameters.
Similar to the model-free case, both black-box methods \citep{nagabandi2019learning,cong2020self,belkhale2021model} and PPG methods \citep{nagabandi2019learning,kaushik2020fast,anne2021meta} have been considered adapting the dynamics model for model-based control.
(We generally avoid classification of methods as task-inference and model-based. See Section~\ref{subsection:model-based}.)

While many meta-RL methods can be applied to robotics, the critical challenges in practice are often different from the conceptual problems studied in the field as a whole.
For example, exploration, which differentiates meta-RL from supervised meta-learning, is not generally the most challenging obstacle in meta-RL for robotics.
Often, it is the case that robots must generalize over a distribution over robot platforms or terrains, which present themselves immediately to the agent.
Instead, the sim-to-real gap, induced by the simulator generally required for meta-raining, presents a significant hurdle.
Some meta-RL methods do mitigate the impact of sim-to-real transfer.
For example, the task inference algorithm, IMPORT, discussed in Section \ref{para:TI with MT training}, leverages privileged information in the simulator.
A nearly identical algorithm has been used successfully in robotics sim-to-real transfer in the robotics literature \citep{kumar2021rma}, despite not being framed as meta-RL in that context.
This phenomenon demonstrates how familiarity with meta-RL could be of immense use to the field at large in practice in the future.

\section{Multi-agent RL}
\label{subsec:marl}
\enlargethispage{-\baselineskip}
Meta-learning has been applied in multi-agent RL to solve a number of problems, from learning with whom to communicate \citep{zhang2021metagrad}, to automating mechanism design by learning agent-specific reward functions \citep{yang2022adaptive}.
Notable problems and solutions are summarized in Table~\ref{tab:multi}.
However, in this section we focus on two main problems that meta-RL can address in the multi-agent setting.
First, we introduce the problems of generalization to unseen agents and non-stationarity,
and discuss how meta-RL can address them in general.
Then, we discuss the types of meta-RL
methods that have been used to address each problem, elaborating on PPG methods that propose additional mechanisms for each problem.

\begin{table}[ht!]
    \footnotesize
    \centering
    \caption{
    Summary of meta-RL papers used for multi-agent RL from Section~\ref{subsec:marl}. Methods use PPG, black-box, and task-inference agents, in addition to non-adaptive agents. As examples, the problems addressed include generalization over other agents policies and non-stationarity induced by other agents.}
    \label{tab:multi}
    \begin{tabular}{C{0.4\textwidth}C{0.5\textwidth}}
        Sub-topic & Papers \\ \hline \hline
        PPG meta-gradients for non-stationarity & \citet{al2018continuous,foerster2018learning,kim2021policy} \\ \hline
        Black box for generalization to teammates & \citet{charakorn2021learning} \\
        Black box for generalization to opponents & \citet{lu2022model} \\ \hline
        Task inference for generalization to teammates &
        \citet{grover18learning,zintgraf2021deep} \\
        Task inference for generalization to opponents &
        \citet{grover18learning,papoudakis2020variational,zintgraf2021deep} \\
        Task inference for generalization to humans & \citet{he2022learning} \\ \hline
        \hline
    \end{tabular}
\end{table}

The first multi-agent problem we consider is generalization over other agents.
In multi-agent RL, many agents act in a shared environment.
Often, it is the case that other agents' policies vary greatly.
This creates a problem of generalization to unseen agents.
This generalization may occur over opponents \citep{papoudakis2020variational,lu2022model}, or over teammates \citep{charakorn2021learning,gupta2021dynamic,he2022learning}, which is sometimes called ad hoc teamwork \citep{stone2010adhoc}.
The other agents may be learned policies \citep{zintgraf2021deep} or even humans \citep{he2022learning}.
By viewing other agents as (part of) the task, and assuming a distribution of agents available for practice, meta-RL is directly applicable.
This view can lead to the application of meta-learning methods \citep{beck2022hyper, zintgraf2020varibad} to the multi-agent setting \citep{zintgraf2021deep, tessera2024hyper}.
Using meta-RL to address generalization over other agents is visualized in Figure~\ref{fig:adhoc}.

\begin{figure}[ht!]
    \centering
    \includegraphics[height=4cm,alt={Using meta-RL to generalize over unseen agents by training on diverse agents and adapting to new ones at meta-test time.}]{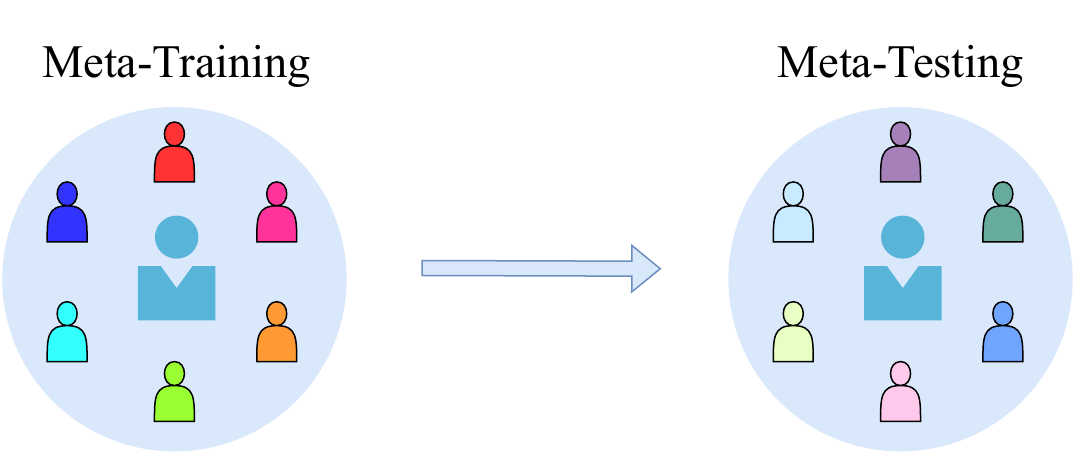}
    \caption{Illustration of using meta-RL to address generalization over unseen agents. By generalizing over other agents at meta-train time, we can adapt to new agents at meta-test time.}
    \label{fig:adhoc}
\end{figure}

The second multi-agent problem we consider is \textit{non-stationarity}.
In multi-agent RL, from the perspective of any one agent, all other agents change as they learn.
This means that from one agent's perspective, if all of the other agents are modeled as part of the environment, then the environment changes -- i.e., the problem is non-stationary \citep{al2018continuous,foerster2018learning,kim2021policy}.
Meta-RL likewise can address non-stationary by treating other learning agents as (part of) the task.
In this case, the learning algorithm of each agent, and what each agent has learned so far, collectively defines the task.
By repeatedly resetting the other learning agents during meta-training, we can meta-learn how to handle the changes introduced by the other agents.
From the perspective of the meta-learning agent, the distribution over other agents remains stationary.
This effectively resolves the non-stationarity of multi-agent RL, which is depicted in Figure~\ref{fig:nonstat}.

Solutions in multi-agent RL make use of all different types of meta-RL methods: PPG methods \citep{al2018continuous,foerster2018learning,kim2021policy}, black-box \citep{charakorn2021learning,lu2022model}, and task-inference methods \citep{grover18learning,papoudakis2020variational,zintgraf2021deep,he2022learning}.
These methods are discussed in Sections \ref{sec:ppg}, \ref{sec:black_box}, and \ref{sec:task_inf}, respectively.
The agent may even use a Markovian (i.e., non-adaptive) policy, if the other agents are learning \citep{lu2022adversarial}.
Most of these methods can be applied without modification to the underlying meta-RL problem to address both generalization over other agents and non-stationarity.
In the case that other agents form the entirety of the task, then the meta-RL objective can be written:

\noindent
\begin{align}
    \mathcal{J}(\theta) = \E_{\pi'_i \sim p(\pi'_i) \forall i=0...A} \bigg[ \E_{\mathcal{D}} \bigg[ G(\mathcal{D}) \bigg| \pi_{f_{\theta}}, \pi'_0...\pi'_A  \bigg]\bigg], \label{eq:marl-objective}
\end{align}
where $A$ is the number of other agents.
However, several PPG papers investigate additional mechanisms for both such generalization and non-stationarity that we discuss next.

\begin{figure}[t!]
    \centering
    \includegraphics[height=4cm,alt={Using meta-RL to train over another agent’s learning in multi-agent RL, addressing non-stationarity.}]{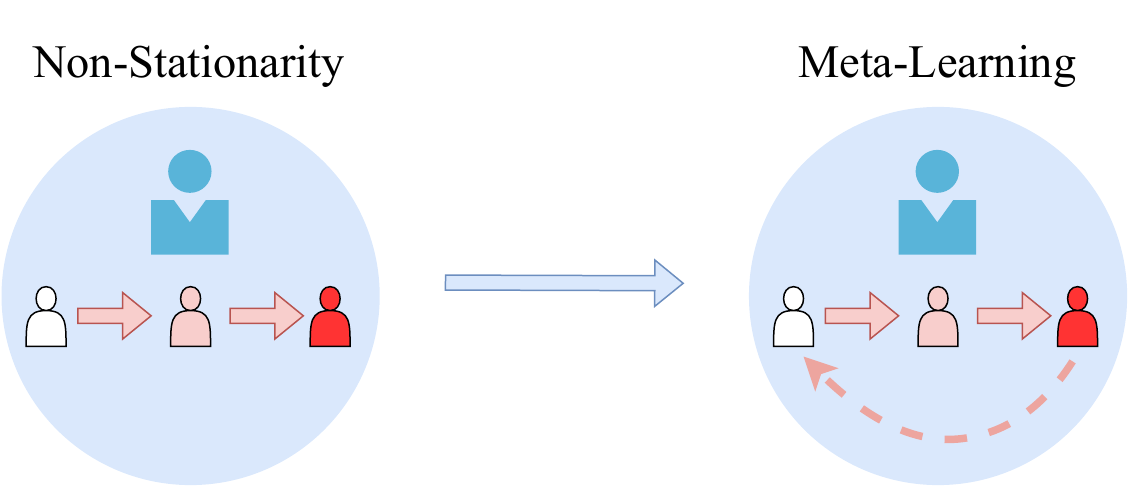}
    \caption{Illustration of using meta-RL to train over another agent's learning in multi-agent RL. By resetting the learning process of other agents, and training over it, we can learn to address the non-stationarity created by the learning of other agents.}
    \label{fig:nonstat}
\end{figure}

Some PPG papers focus on improving the distribution of other agents using meta-gradients, in order to improve generalization to new agents.
See Section~\ref{sec:ppg} for a discussion of meta-gradient estimation.
These papers focus on the curriculum of opponents to best support learning in a population-based training setting.
\citet{gupta2021dynamic} iterates between PPG meta-learning over a distribution of fixed opponents and adding the best-response to the meta-learned agent back to the population.
Alternatively, \citet{feng2021neural} uses meta-gradients or evolutionary strategies to optimize a neural network to produce opponent parameters.
The opponent parameters are chosen to maximize the learning agent's worst case performance.
Both methods use the distribution of agents to create a robust agent capable of generalizing across many other agents.

Finally, some PPG papers introduce additional mechanisms to handle non-stationarity.
In order to resolve all non-stationarity, all other (adaptive or non-adaptive) agents must repeatedly reset to their initial policies eventually.
While this has been explored \citep{papoudakis2020variational,zintgraf2021deep,charakorn2021learning}, PPG methods tend to allow other agents to continue to learn \citep{al2018continuous,foerster2018learning}, or even to meta-learn \citep{kim2021policy}, without resetting.
Each PPG method addresses the non-stationary learning of other agents in a different way.
For example, \citet{al2018continuous} propose meta-learning how to make gradient updates such that performance improves against the subsequent opponent policy, over various pairs of opponent policies.
In contrast, \citet{foerster2018learning} derive a policy gradient update such that one agent meta-learns assuming the rest follow an exact policy gradient, and \citet{kim2021policy} derive a policy gradient update assuming all agents sample data for meta-learning.

\chapter{Open Problems} \label{sec:open_problems}
Meta-RL is an active area of research with an increasing number of applications, and in this nascent field there are many opportunities for future work. In this section, we discuss some of the key open questions in meta-RL.
Following the method categories in this survey, we first discuss some important directions for future work in few-shot and many-shot meta-RL in Section~\ref{subsec:short-horizon open problems} and \ref{subsec:long-horizon open problems} respectively.
Following this, we discuss how to utilize offline data in Section~\ref{subsec:offline data future work} in order to eliminate the need for expensive online data collection during adaptation and meta-training.
Finally, we take a critical stance and evaluate some potential limitations of meta-RL research.

\section{Few-shot Meta-RL Generalization}
\label{subsec:short-horizon open problems}

As discussed in Section~\ref{sec:fast_adaptation}, few-shot meta-RL methods are a promising solution to fast adaptation on new tasks.
However, so far their success is mainly achieved on narrow task distributions, while the ultimate goal of meta-RL is to enable fast acquisition of entirely new behaviors.
Consequently,  future work should focus more on generalization of few-shot meta-RL methods to broader task distributions.
A straightforward way to achieve this goal is to meta-train on a broader task distribution to learn an inductive bias that can generalize to more tasks.
However, training on a broader task distribution also introduces new challenges (such as a harder exploration problem) that are beyond the scope of existing methods.
For an overview of methods for generalization in RL, beyond meta-RL, see \citet{kirk2023survey}.
Moreover, even when trained on a broad task distribution, the agent may still encounter test tasks that lie outside the training distribution.
Generalization to such out-of-distribution (OOD) tasks is important since a meta-RL agent is likely to encounter unexpected tasks in the real world.
For a summary of the task distributions in existing benchmarks, see Table~\ref{tab:benchmarks}.
We discuss these two open problems, illustrated in Figure~\ref{fig:ood}, in more detail below.

\begin{table}[ht!]
    \footnotesize
    \centering
    \caption{Summary of existing benchmarks in meta-RL by setting and number of tasks. MuJoCo and 2D Navigation are perhaps the most common, but also the narrowest meta-training distributions. XLand, Alchemy, NetHack, and Procgen are procedurally generated and so contain many tasks. NetHack and Procgen, however, were not designed to evaluate adaptation in meta-RL, and XLand is private. Additionally, benchmarks that contain both discrete tasks and diverse behavior generally require many tasks in order to generalize to new tasks.}
    \label{tab:benchmarks}
    \resizebox{\textwidth}{!}{
    \begin{tabular}{
        >{\centering\arraybackslash}m{0.2\textwidth}
        >{\raggedright\arraybackslash}m{0.22\textwidth}
        >{\centering\arraybackslash}m{0.19\textwidth}
        >{\raggedright\arraybackslash}m{0.22\textwidth}
    }
        \hline
        Benchmark Name & Citation & Setting & Number of Tasks \\
        \hline \hline
        XLand 2.0 (private) & \citet{team2023human} & Game (Diverse) & $>10^{40}$ \\
        Alchemy & \citet{wang2021alchemy} & Game (Diverse) & $>167,424$ \\
        NetHack & \citet{kuttler2020nethack}  & Game (Diverse) & $>>200,000$ \\
        Procgen & \citet{cobbe2020leveraging}  & Game (Diverse) & $>>200,000$ \\
        Sonic & \citet{nichol2018gotta}  & Game (Diverse) & 58 discrete \\
        Meta Arcade & \citet{staley2021meta}  & Game (Diverse) & 24; continuous (e.g., ball size) \\
        DreamerGrader & \citet{wang2021alchemy} & Game (Narrow) & 3,556 discrete \\
        2D Navigation & e.g., \citet{finn2017model,zintgraf2020varibad} & Game (Narrow) & 25 discrete or continuous goal \\
        RLBench & \citet{james2020rlbench} & Robotic (Diverse) & 100 discrete \\
        Meta-World & \citet{yu2020meta} & Robotic (Diverse) & 45 discrete; continuous (e.g., goal) \\
        MuJoCo~\citep{todorov2012mujoco} Locomotion & e.g., \citet{finn2017model,zintgraf2020varibad} & Robotic (Narrow) & e.g., 2 discrete or continuous goal \\
        \hline
    \end{tabular}}
\end{table}

\begin{figure}[ht!]
    \centering
    \includegraphics[width=\linewidth, alt={Illustration of broad task distributions (left) and out-of-distribution generalization (right).}]{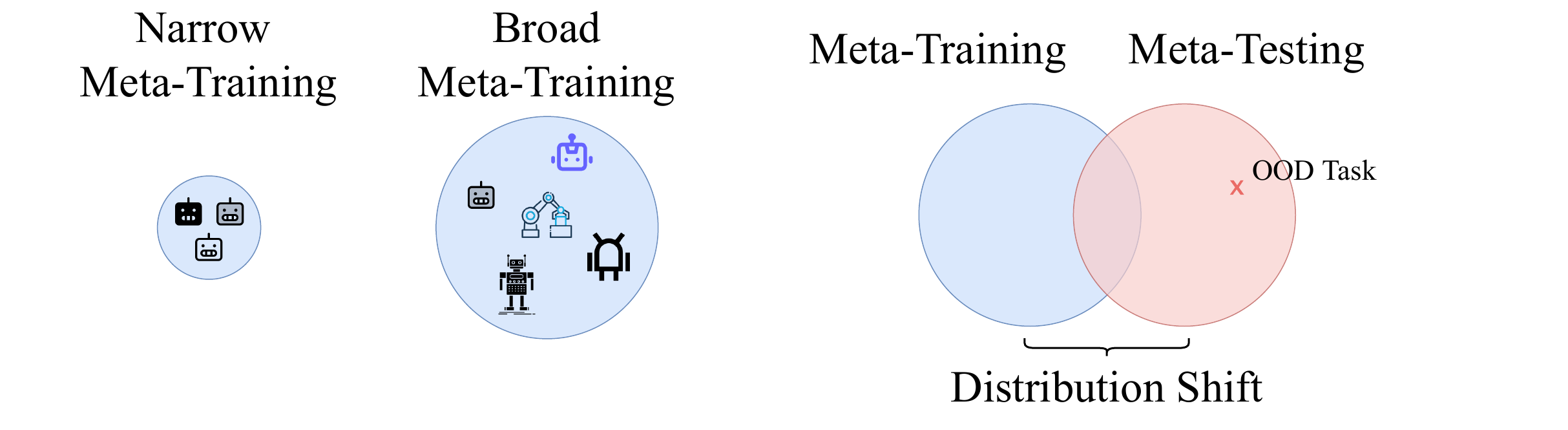}
    \caption{Illustration of broad task distributions (left) and OOD generalization (right). There is a clear need for meta-training distributions with both more tasks and more diverse tasks. Novel methods may be needed for such distributions, including methods that fail gracefully when out of distribution.}
    \label{fig:ood}
\end{figure}

\paragraph{Training on broader task distributions for better generalization}
\label{para:broader task distribution}
The meta-training task distribution plays an important role in meta-RL, as it determines \emph{what} inductive bias we can learn from data, while the meta-RL algorithms determine \emph{how well} we can learn this inductive bias.
The task distribution should be both diverse enough so that the learned inductive bias can generalize to a wide range of new tasks, and clear enough in task structure so that there is indeed some shared knowledge we can utilize for fast adaptation.

However, existing few-shot meta-RL methods are frequently meta-trained on narrow
task distributions, where different tasks are simply defined by varying a few parameters that specify the reward function or environment dynamics \citep{duan2016rl,finn2017model,rakelly2019efficient,zintgraf2020varibad}.
While the structure between such tasks is clear, the narrowness of the distribution poses problems.
For example, in this setting, task inference is generally trivial. Often, the agent can infer the parameters that define the task based on just a few transitions. This makes it hard to evaluate if a meta-RL method can learn systematical exploration behaviors to infer more sophisticated task structures.
In general, the inductive bias that can be learned from a narrow task distribution is highly tailored to the specific training distribution,
which provides little help for faster acquisition of entirely new behaviors in a broader task domain.
Indeed, \citet{mandi2022on} find that meta-RL methods are no better than multi-task pre-training when tested on generalization to complex visual tasks like Atari games.

Consequently, we need to design benchmarks that are diverse, in addition to having clear task structures.
Such a benchmark would better reflect the complex task distribution in real-world problems, and promote the design of new methods that can generalize over these challenging tasks.
For example, some recent robotic benchmarks \citep{james2020rlbench,yu2020meta} introduce a wide range of manipulation tasks in simulation with not only parametric diversity (such as moving an object to different goals), but also non-parametric diversity (such as picking an object and opening a window).
Game benchmarks with procedurally generated environments provide another good testbed to evaluate meta-RL on broader task distributions \citep{nichol2018gotta,cobbe2020leveraging,kuttler2020nethack,staley2021meta,wang2021alchemy}, as novel environments with both diverse and clear task structures can be easily generated by following different game rules.
For example, Alchemy \citep{wang2021alchemy} is a video game benchmark with the goal of transforming stones with potions into more valuable forms, which requires the agent to strategically experiment with different hypotheses for efficient exploration and task inference.
As another example, the Adaptive Agent (Ada) evaluated on XLand 2.0 learns to experiment, use tools, and navigate in a 3D open-ended environment.
In addition to these simulation benchmarks, \citet{liu2022giving} introduce a real-world benchmark which requires systematic exploration to discover the errors in different programs for coding feedback.
Existing meta-RL methods cannot yet achieve satisfactory performance on many of these more challenging benchmarks \citep{alver2020a,yu2020meta,wang2021alchemy,mandi2022effectiveness}, which shows that generalization to a wider task distribution is still an open problem,
and more attention should be paid to these more challenging benchmarks to push the limit of meta-RL algorithms.

Still, both new methods and new benchmarks are still needed in meta-RL.
Novel methods, such as the use of curriculum learning \citep{mehta2020curriculum, team2023human} and active selection of tasks in the distribution based on task descriptions \citep{kaddour2020probabilistic}, will likely be needed as well to address sufficiently broad distributions.
Many of these benchmarks are insufficient as well, with more work needed to improve the task distribution itself.
For example, some of benchmarks with the widest task distributions are private \citep{team2023human}, or not designed to test adaptation in meta-RL \citep{kuttler2020nethack,cobbe2020leveraging}.
It is also possible that some of these benchmarks simply need to be more densely populated with tasks in order for few-shot adaptation to be learnable.
For example, Meta-World \citep{yu2020meta} has benchmarks with 1, 10, or 45 discrete tasks; however, meta-learning an algorithm that can reliably adapt to unseen meta-test tasks may require tens of thousands of meta-training tasks \citep{kirsch2022general}.
Adapting benchmarks \citep{benjamins2021carl} from the related Contextual MDP literature \citep{hallak2015contextual} in order to remove the task identity from the state is a path forward.
Procedurally generating tasks is also one promising path for benchmarks \citep{cobbe2020leveraging, kuttler2020nethack, wang2021alchemy, team2023human}.
When tasks have manually designed discrete variation, it can be difficult to determine whether meta-test tasks even have support within the meta-training distribution.

\paragraph{Generalization to OOD tasks}
In few-shot meta-RL, it is commonly assumed that the meta-training and meta-test tasks are drawn from the same task distribution.
However, in real-world problems, we usually do not know a priori all the situations the agent may face during deployment, and the RL agent will likely encounter test tasks that lie outside the meta-training task distribution.

One key challenge here is that we do not know to what extent the learned inductive bias is still helpful for solving the OOD tasks.
For example, in the navigation task in Figure~\ref{fig:thompson}, the learned inductive bias is an exploration policy that traverses the edge of the semicircle to find the goal first.
If we consider OOD tasks of navigation to goals on the edge of a semicircle with a larger radius, then the learned exploration policy is no longer optimal, but may still help the agent explore more efficiently than from scratch.
However, if the OOD tasks are navigation to goals on a semicircle in the opposite direction, then the learned inductive bias may be actively harmful and slow down learning.

Consequently, simply using the learned inductive bias, which is commonly adopted by existing few-shot meta-RL methods, is not sufficient for OOD generalization.
Instead, the agent needs to adaptively choose how to utilize or adjust the learned inductive bias according to what kinds of OOD tasks it is solving.
On one hand, we want to ensure that the agent can generalize to any OOD task given enough adaptation data, even if the learned inductive bias is misspecified.
In principle, PPG methods can satisfy this requirement while black-box methods cannot, echoing our discussion on the trade-off between generalization and specialization of few-shot meta-RL methods in Section~\ref{sec:ppg}.
However, in practice, if the meta-testing task distribution is sufficiently dissimilar to the meta-training task distribution, then any sort of meta-learning can be catastrophic \citep{xiong2021on}.
On the other hand, we want to utilize as much useful information from the learned inductive bias as possible to improve learning efficiency on OOD tasks.
Although recent works investigate how to improve generalization \citep{lan2019meta,lin2020model,grewal2021variance,xiong2021on,ajay2022distributionally,he2022learning,imagawa2022off,greenberg2023train,wang2023simple} or adaptation efficiency \citep{fakoor2020meta,mendonca2020meta,lee2021improving} on OOD tasks with small distribution shifts, more work remains to be done on how to adaptively handle larger distribution shifts between training and test tasks.
Ideally, OOD methods make use of their inductive bias where possible, and fail gracefully (e.g., defaulting to an engineered learning algorithm), where it is not possible.

\let\mysectionmark\sectionmark
\renewcommand\sectionmark[1]{}
\section{Many-shot Meta-RL: Optimization Issues and Standard Benchmarks}
\label{subsec:long-horizon open problems}
\let\sectionmark\mysectionmark
\sectionmark{Many-shot Meta-RL}

For many-shot meta-RL (see Section~\ref{sec:long_task_horizon_meta_rl}), outer-loop optimization poses significant problems, some of which remain open.
Moreover, there is a lack of standard benchmarks to compare different many-shot meta-RL methods, an important gap to fill for future work.

\enlargethispage{-\baselineskip}
\paragraph{Optimization issues in many-shot meta-RL}
In many-shot meta-RL, the inner-loop updates the policy many times, which leads to a challenging optimization problem in the outer-loop due to not smooth
objective surfaces~\citep{metz2021gradients} and high computation cost.
To deal with these challenges in practice, most methods use the performance after relatively few
inner-loop updates compared to the full inner-loop optimization trajectory as a surrogate objective.
Updating the inner-loop after only a fraction of the lifetime leads to bias in the gradient estimation, which can be detrimental to meta-learning performance~\citep{wu2018understanding}.
How to tackle this optimization issue remains an open problem.
One approach is to use gradient-free
optimization methods such as evolution strategies~\citep{rechenberg1971evolutionsstrategie} as was done by \citet{kirsch2021introducing}, but its sample complexity is much worse than gradient-based optimization in settings where those are applicable.

\paragraph{Truncated optimization in many-shot single-task meta-RL}
Even if optimizing over long lifetimes was possible in the multi-task setting, in the single-task setting we still need to update the inner-loop before learning has finished in order for it to be useful.
One approach to improve truncated optimization on a single task is the bootstrapped surrogate objective \citep{flennerhag2021bootstrapped}, which approximates an update for a longer truncation length with a bootstrapping objective on a shorter truncation length.
However, this also introduces a biased meta-gradient estimation.
Another solution computes meta-gradients with different numbers of updates in the inner-loop and then computes a weighted average, similar to TD($\lambda$) \citep{bonnet2021one}.
Additionally, there are still more sources of bias in this setting, such as bias from the reuse of critics between the inner- and outer-loops \citep{bonnet2022debiasing}.
More research is required to choose the optimal bias-variance trade-off for meta-gradient estimators under the single-task setting.

\paragraph{Non-stationary optimization in many-shot single-task meta-RL}
Another central challenge in many-shot single-task meta-RL is the non-stationarity
 of the inner-loop.
In multi-task meta-RL, the inner-loop revisits the same tasks multiple times, allowing the meta-learner to fit to the stationary training task distribution.
In the single-task case, however, the agent parameters keep changing, making the meta-learning problem non-stationary.
Learning in a non-stationary problem is an open area of research extending beyond online meta-RL.

\paragraph{Benchmarks for many-shot meta-RL}
Many-shot meta-RL methods are mainly evaluated on a wide range of commonly used RL tasks, such as Atari \citep{bellemare2013arcade}, classic control \citep{brockman2016openai} and continuous control \citep{duan2016benchmarking},
to show that the learned RL algorithms have good generalization.
However, some papers test generalization across different domains while others evaluate within a single domain, and there are no unified criteria on how to split the training and test tasks on the chosen domain(s).
In the single-task setting, a benchmark for hyperparameters tuning in RL has been proposed, but focuses on a fixed and small number of discrete meta-parameters \citep{shala2022autorlbench}.
In the multi-task setting, to better evaluate generalization of the learned algorithms, it could be helpful to design and adopt benchmarks that choose the meta-train and meta-test tasks based on some unified standard.
In this direction, a useful benchmark may even provide multiple groups of train and test tasks in order to gradually increase the difficulty and degree of transfer required, where ideally the degree of transfer is quantifiable by some measure of similarity across the MDPs \citep{ammar2014automated}.
Moreover, the ultimate goal of many-shot meta-RL is to design general-purpose RL algorithms that can work well on any reasonable MDP.
However, the exact task structure in such a distribution over all ``reasonable'' MDPs is still unclear and needs clarification, as some structure must be shared across these MDPs to allow for meta-learning in the first place.

\section{Utilizing Offline Data in Meta-RL}
\label{subsec:offline data future work}

So far the majority of research in meta-RL focuses on the setting of using online data for both the outer-loop and the inner-loop.
However, when offline data is available, utilizing it properly is an effective way to reduce the need for expensive online data collection during both meta-training (the outer-loop) and adaptation (the inner-loop).
Depending on which kind of data is available for the outer-loop and inner-loop respectively, we have four different settings, depicted in Figure~\ref{fig:offline}.
Apart from fully online meta-RL, the remaining three settings are still underexplored, and we discuss them below.

\begin{figure}[ht!]
    \centering
    \includegraphics[width=\textwidth,alt={The four settings of online and offline data use in meta-RL: (1) online outer-loop, online inner-loop, (2) offline outer-loop, online inner-loop, (3) online outer-loop, offline inner-loop, and (4) offline outer-loop, offline inner-loop.}]{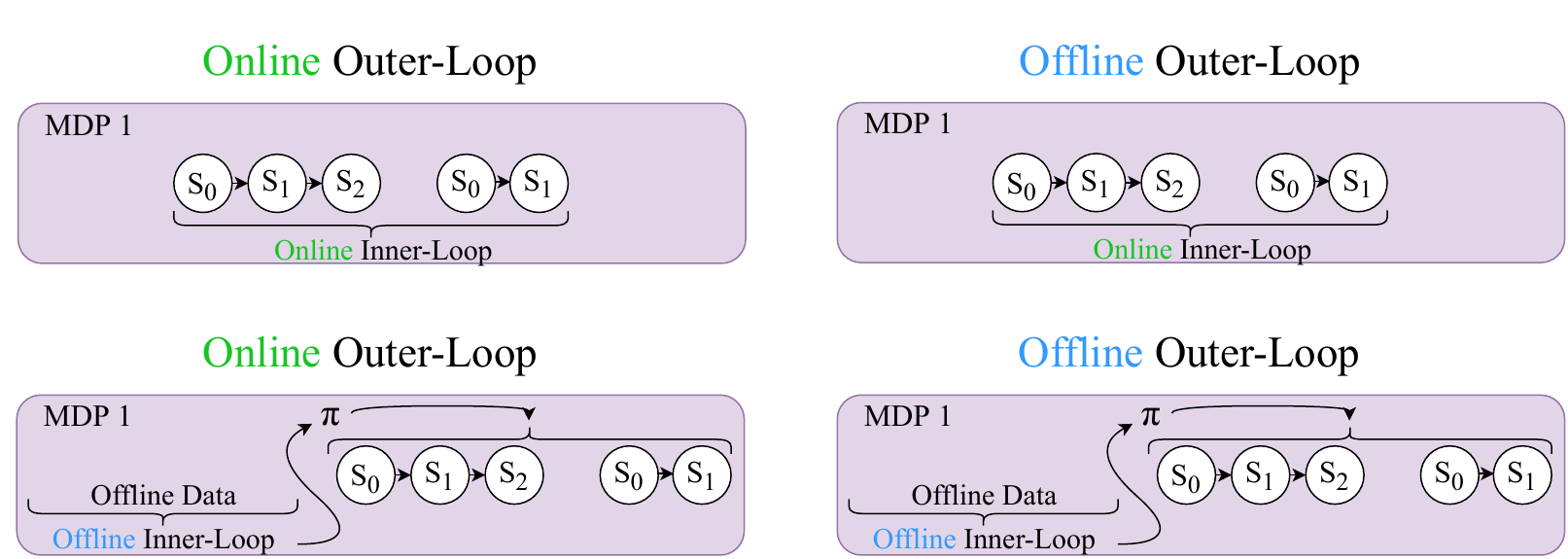}
    \caption{The four different settings using online or offline data. Using an online outer-loop and online inner-loop is the standard setting. Using an offline outer-loop with offline inner-loop allows for safe meta-learning and safe-adaptation by forgoing all exploration, but places additional demands on the data provided. Using an offline outer-loop with an online inner-loop allows for safe meta-learning of exploration behavior, but introduces difficulties such as extensive distribution shift. Using an online outer-loop with offline inner-loop allows for meta-learning offline RL algorithms without distribution shift, but still requires online samples during meta-learning.}
    \label{fig:offline}
\end{figure}

\paragraph{Offline outer-loop and offline inner-loop}
Under this setting, the agent can only adapt with offline data, and learn to optimize its adaptation strategy also with offline data in the outer-loop \citep{mitchell2020offline,lee2020batch,li2020efficient,mitchell2021offline,lin2022modelbased,yuan2022robust}.
This is particularly suitable for scenarios where online exploration and data collection is costly and even dangerous, while offline logs of historical behaviors are abundant, such as in robotics and industrial control \citep{levine2020offline}.
However, the offline setting also introduces new challenges.
First, as in standard offline RL, returns must be estimated solely from offline data, creating issues such as the overestimation of returns \citep{Fujimoto2019offpolicy,levine2020offline}.
Second, and unique to meta-RL, instead of doing online exploration to collect the data, $\mathcal{D}$, for adaptation, we can only adapt with whatever offline data is provided for each task.
Consequently, the adaptation performance critically depends on how informative the offline data is about the task.
Existing works mainly investigate this offline setting on simple task distributions where task identity can be easily inferred from a few randomly sampled transitions in the offline data.
However, how to adapt with offline data on more complicated task distributions (such as those discussed in Section~\ref{para:broader task distribution}), and how different offline data collection schemes may influence the performance of offline adaptation, remain open questions.

\paragraph{Offline outer-loop and online inner-loop}
In this setting, the agent learns to adapt with \emph{online} data, by meta-training on purely \emph{offline} data \citep{dorfman2021offline,pong2022offline,ghosh2022offline}.
Compared to the fully offline setting, this setting is more suitable for the scenarios where few-shot online adaptation is allowed during deployment.
Online adaptation tackles the aforementioned problem of limited exploration when adapting with only offline data, but it also introduces a new challenge: how can we learn a systematic exploration policy from offline data collected by some unknown policies?
Usually, the desired behaviors of the exploration policy are not covered in the offline data, which is generally collected in a task-specific way.
This creates a distribution shift between the exploration policy we want to learn and the offline data we can learn from.
This shift can be especially problematic if the offline data was collected only with an expert that has access to the ground-truth task \citep{rafailov2021reflective}.
In particular, it can introduce a problem of ambiguity in the identity of the MDP \citep{dorfman2021offline}.
To tackle these challenges, existing works make additional assumptions on the data collection scheme, which strictly speaking, violate the offline meta-training setting and thus limit their application.
For example, \citet{pong2022offline} allow for some online meta-training, but only with unsupervised interaction, and they learn a reward function to generate supervision for this unsupervised online interaction.
Additionally, \citet{rafailov2021reflective} make assumptions about the reward that enable easy reward relabeling when swapping offline trajectories between tasks.
How to relax offline assumptions to make this setting more applicable remains an open problem.

\paragraph{Online outer-loop and offline inner-loop}
Under this setting, the agent can only adapt with offline data.
However, online data is available in the outer-loop to help the agent learn what is a good offline adaptation strategy, which may be easier to learn than in an offline outer-loop setting.
In other words, the agent is learning to do offline RL via online RL.
This setting is appealing for two reasons:
(1) It maintains the benefits of using solely offline data for adaptation,
while being easier to meta-train than the fully offline setting.
(2) It is difficult to design good offline RL algorithms \citep{levine2020offline}, and meta-RL provides a promising approach to automating this process.
A couple of existing methods do combine an offline inner-loop with an online outer-loop \citep{prat2021peril,dance2021demonstration}.
However, these approaches both allow for additional online adaptation after the offline adaptation in the inner-loop, and the offline data either contains only observations \citep{dance2021demonstration}, or conditions on additional expert actions \citep{prat2021peril}.
One method uses permutation-invariant memory to enable an off-policy inner-loop, but only evaluates with data collected from prior policies \citep{imagawa2022off}.
Consequently, offline RL via online RL remains an interesting setting for future work with the potential for designing more effective offline RL algorithms by meta-learning in both the few-shot and many-shot settings.

\section{Limitations}
So far we have presented a positive case for the use and development of meta-RL.
While meta-RL can confer many advantages, as a tool for solving problems, it also presents several trade-offs.
At this point, we take a step back to discuss four limitations of meta-RL from a more critical perspective.

While meta-RL can enable sample-efficient learning at deployment, it does so at the expense of increased sample complexity during meta-training.
For this reason, meta-RL is only applicable when upfront data collection is relatively cheap, or adaption during deployment is prohibitively expensive.
For example, the trade-off presented by meta-learning makes sense when a simulator is available for meta-training, or when adaptation needs to be efficient and frequent after meta-training.
While this limitation is restrictive, it is not significantly more restrictive than the standard use case for reinforcement learning.

Meta-RL additionally presents a trade-off between transferability and interpretability on one hand and sample efficiency and engineering burden on the other.
While meta-RL methods may enable adaptation that is more sample-efficient than a manually engineered alternative, the insights produced by such systems may be less interpretable and transferable to novel problems.
For example, in the many-shot setting, it may be possible to meta-learn a general-purpose RL algorithm that is better than state-of-the-art algorithms, but even so, the exact mechanisms improving performance may be unclear.
Consider meta-learning an objective function that is parameterized as a black-box.
The meta-learned objective function may have a simple and interpretable form, but extricating this form from the weights and biases of a neural network is difficult.
Meta-RL replaces engineering effort and domain expertise with computation, and in doing so, trades-off interpretability for performance.

In the few-shot setting (Section~\ref{sec:fast_adaptation}), it is common to adapt to tasks using gradient updates in the inner-loop.
However, whether meta-learning to account for this procedure is worthwhile can depend on the particular task distribution.
For example, when there is a limited number of tasks or limited amount of data in the inner-loop, it may even be preferable to have MAML perform no task adaptation during meta-training and only fine-tune at meta-test time~\citep{gao2020modeling}.
(In the supervised setting, it has likewise been shown that fine-tuning, only after meta-training, outperforms learning to fine-tune, when meta-training a black-box method; \cite{beck2025metalic}.)
Additionally, if the task distribution never requires different actions for the same state in different tasks, e.g., because the state-space does not overlap, then adaptation may not be needed and multi-task training may be sufficient.
In such a setting, even if more adaptation is useful for generalization, meta-RL may confer little to no advantage compared to fine-tuning~\citep{mandi2022effectiveness}.
Moreover, even if adaptation to each task is needed, and sufficient data is provided, as is typically assumed, meta-RL can actually confer a disadvantage.
Specifically, test-time adaptation can decrease performance on some tasks in practice~\citep{deleu2018effects}, and training from scratch can yield higher returns~\citep{xiong2021on}.
(Similar observations have been made in the meta-supervised setting as well; \cite{triantafillou2019meta}.)
In fact, if the meta-testing task distribution is sufficiently dissimilar to the meta-training task distribution, then any sort of meta-learning can be catastrophic, since the agent learns incorrect inductive biases~\citep{xiong2021on}.

Finally, while the meta-RL literature develops many specialized methods, the problem setting may not require such methods.
Consider that the meta-RL problem can be considered a particular type of POMDP, as discussed in Section~\ref{sec:optimal_exploration}.
Given this, it is reasonable to question whether specialized methods are needed for meta-RL at all.
In the zero-shot setting, recent work has suggested that many complex and specialized methods may be unnecessary.
For example, some task-inference methods can be nearly matched by well-tuned end-to-end recurrent networks~\citep{ni2022recurrent}; complicated task-inference parameterizations and supervision can be outperformed by hypernetworks trained end-to-end~\citep{beck2022hyper}; some task inference methods~\citep{humplik2019meta, fu2021towards} can be seen as applications of more general POMDP methods for inferring a hidden state~\citep{moreno2018neural, guo2018neural}; and complicated task-inference reward bonuses for exploration may be just as effectively replaced by more general exploration methods for POMDPs~\citep{yin2021sequential} applied to the meta-RL problem setting~\citep{zintgraf2021exploration}.
Even for complicated task distributions, methods designed for MDPs and more general POMDPs, such as curriculum learning, distillation, and transformer architectures, can be sufficient to enable meta-learning~\citep{team2023human}.
Still, even if specialized methods are not strictly necessary, such methods can enable more efficient meta-learning.
Moreover, whether meta-learning methods are developed explicitly or not, meta-learning will at the very least be an emergent phenomenon of any capable and general agent.
For this reason, insights from meta-RL should assist practitioners in developing and reasoning about such systems.

\chapter{Conclusion} \label{sec:conclusion}
In this survey, we presented a survey of meta-RL research focused on two major categories of algorithms as well as applications.
We found the majority of research focused on the few-shot multi-task setting, where the objective is to learn an RL algorithm that adapts to new tasks from a known task distribution rapidly using as few samples as possible.
We discussed the strengths and weaknesses of the few-shot algorithms, which generally fall in the categories of parameterized policy gradient, black box, and task inference methods.
A central topic in using these methods is how to explore the environment to collect that data.
We identified the different exploration strategies in the literature and discussed when each of them are applicable.
Besides meta-RL in the few-shot setting, a rising topic in meta-RL looks at algorithms in the many-shot setting, where two distinct problems are considered: the generalization to broader task distributions and faster learning on a single task.
We found the methods for these two seemingly opposite problems to be surprisingly similar, as they are often based on augmenting standard RL algorithms with learned components.
We presented promising applications of meta-RL, especially those in robotics, where meta-RL is starting to enable significant sample efficiency gains in e.g., sim-to-real transfer.
The sample efficiency of RL algorithms is a major blocker in learning controllers for real-world applications.
Therefore, if meta-RL delivers on the promise of sample efficient adaptation, it would enable a wide variety of applications.
In order to push meta-RL further and enable new applications, we found that broader and more diverse task distributions need to be developed for training and testing the meta-RL algorithms.
With promising applications in sight and a range of open problems awaiting solutions, we expect meta-RL research to continue to actively grow.

\appendix
\addtocontents{toc}{\protect\setcounter{tocdepth}{-1}}

\chapter{List of Venues Surveyed}
\label{app:conferences}
This survey is primarily based on the meta-RL research presented in the following conferences and workshops for the years from 2017 to 2022:
\begin{itemize}
    \item International Conference on Learning Representations (ICLR)
    \item Conference on Neural Information Processing Systems (NeurIPS)
    \item International Conference on Machine Learning (ICML)
    \item Autonomous Agents and Multiagent Systems (AAMAS)
    \item Annual AAAI Conference on Artificial Intelligence (AAAI)
    \item Conference on Robot Learning (CoRL)
    \item Robotics: Science and Systems (RSS)
    \item \begin{sloppypar}International Conference on Intelligent Robots and Systems (IROS)\end{sloppypar}
    \item NeurIPS Workshop on Meta-Learning
    \item ICLR Workshop on Meta-Learning
\end{itemize}

\addtocontents{toc}{\protect\setcounter{tocdepth}{0}}
\backmatter
\sloppy
\printbibliography

\end{document}